\documentclass[final]{clv2025}

\jvol{vv}
\jnum{nn}
\jyear{2025}

\dochead{Survey} 

\pageonefooter{Action editor: \{action editor name\}. Submission received: DD Month YYYY; revised version received: DD Month YYYY; accepted for publication: DD Month YYYY.}

\usepackage{amsmath}
\usepackage{booktabs}

\usepackage{verbatim}
\usepackage{booktabs}
\usepackage{graphicx}
\usepackage[table]{xcolor}
\usepackage{threeparttable}
\usepackage{multirow}
\usepackage{graphicx}
\usepackage{subfig}
\usepackage{tikz}
\usepackage{url}
\usepackage{hyperref}
\usepackage{forest}
\usepackage[most]{tcolorbox}
\usetikzlibrary{trees,positioning,shapes,shadows,arrows.meta}
\usepackage[parfill]{parskip}
\usepackage{fontawesome}
\usepackage{pifont}
\definecolor{brandeisblue}{rgb}{0.0, 0.44, 1.0}
\definecolor{mygray}{RGB}{220,220,220}
\definecolor{deepgreen}{rgb}{0,0.6,0}

\usepackage{fancyhdr}
\newif\ifblue
\bluefalse   

\newcommand{\colortext}[1]{%
    \ifblue
        \textcolor{blue}{#1}%
    \else
        #1%
    \fi
}

\runningtitle{What is the Role of Small Models in the LLM Era: A Survey}
\runningauthor{Chen and Varoquaux}

\begin{document}

\title{What is the Role of Small Models in the LLM Era: A Survey}

\author{Lihu Chen\thanks{Corresponding authors}$^{1}$,  
Gaël Varoquaux$^{2}$}

\affilblock{
    \affil{Imperial College London, UK\\\quad \email{lihu.chen@imperial.ac.uk}}
    \affil{Soda, Inria Saclay, France\\\quad \email{gael.varoquaux@inria.fr}}
}

\maketitle

\begin{abstract}
Large Language Models (LLMs) have demonstrated remarkable capabilities in various reasoning tasks, which leads to the development of increasingly large models. However, scaling up model sizes results in significantly higher computational costs and energy consumption, which makes these models impractical for academic researchers and businesses with limited resources. At the same time, Small Models (SMs) are frequently used in practical settings, although their significance is currently underestimated. This raises important questions about the role of small models in the era of LLMs, a topic that has received limited attention in prior surveys.
In this work, we systematically examine the relationship between LLMs and SMs from two key perspectives: \emph{Collaboration} and \emph{Competition (\emph{or Complementarity})}.
We hope this survey provides valuable insights for practitioners, fostering a deeper understanding of the contribution of small models and promoting more efficient use of computational resources. Code and additional resources for this survey are available at: \url{https://github.com/tigerchen52/awesome_role_of_small_models} 
\end{abstract}

\section{Introduction}

The fast progress of Large Language Models (LLMs) has transformed the field of natural language processing (NLP).  
LLMs have demonstrated exceptional performance across a range of tasks, including language generation~\cite{dong2022survey}, language understanding~\cite{wang2018glue}, and domain-specific applications such as coding~\cite{jiang2024survey}, medicine~\cite{he2023survey}, and law~\cite{sun2023short}. Notably, certain capabilities are enhanced by increasing the size of models, with some abilities only visible in larger models~\cite{wei2022emergent}.
This has led to a surge of research and development focused on building ever-larger models, such as GPT-4~\cite{openai2023gpt4} and DeepSeek-R1~\cite{deepseek2025r1}.

However, these gains come at a substantial cost. Scaling up model sizes leads to significant increases in computational costs and energy consumption~\cite{wanefficient2023,varoquaux2024hype}. Additionally, training and deploying LLMs is often impractical in resource-constrained settings, such as for academic researchers, businesses without a strong revenue stream, or deployment on edge devices. As a result, these limitations have motivated a growing interest in the development of small models (SMs)~\cite{lu2025demystifying,wang2024comprehensive}, which can achieve competitive performance with reduced data requirements and fewer parameters.

\begin{figure*}
    \centering
    \includegraphics[width=0.80\textwidth]{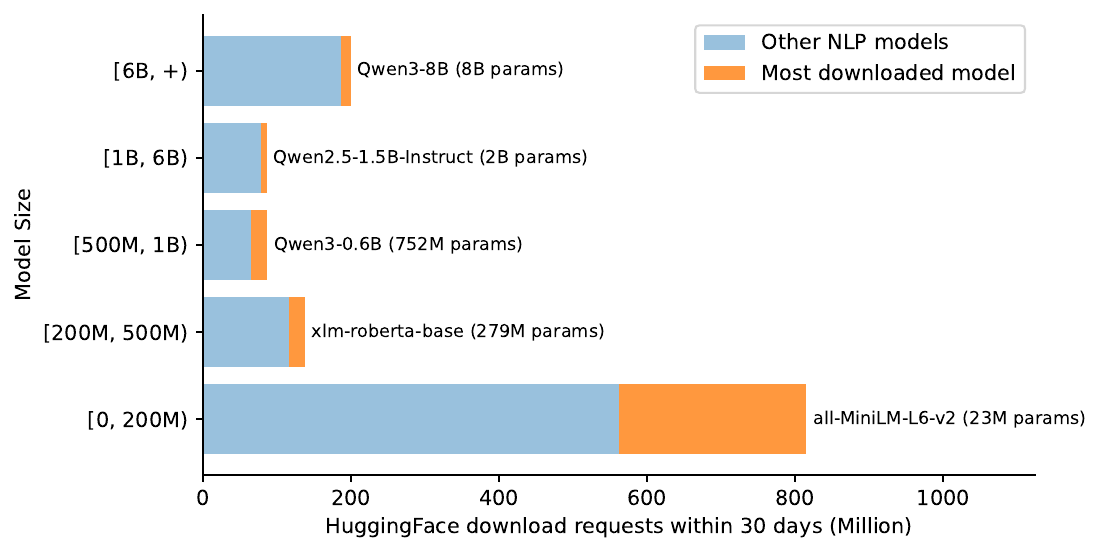}
    \caption{The relationship between model size and monthly downloads. This analysis considers open-source NLP models hosted on HuggingFace and categorizes them into five size groups based on the number of parameters: \texttt{[200M, 500M, 1B, 6B]}. 
    For each group, the full bar represents the total monthly download requests of all models, while the highlighted segment corresponds to the most downloaded model.
    The data was collected on 6 September 2026. }
    \label{fig:model_download}
\end{figure*}

Recognizing the rise of SMs, we now ask \textbf{what qualifies as ``small'' and what model scope we consider in this survey?}  
In this work, we use the term \emph{small models} in a broader sense, to include a wider variety of architectures and use-cases.

\textbf{\emph{Definition of Small Models.}} As counterparts to LLMs, \emph{Small Models} (SMs) generally refer to models with a relatively lower number of parameters. \colortext{Rather than focusing on a particular model architecture, this survey emphasizes the relationship between small and large models, where the notion of \emph{small} is inherently relative to the reference model. Consequently, SMs may include both encoder-only and decoder-only Transformer language models, as well as, more broadly, other neural architectures or even statistical models. Although the examples discussed in this survey are mainly Transformer-based language models, the underlying principles are not tied to a specific architecture.
At the same time, there is no universally accepted definition or parameter threshold separating \emph{large} from \emph{small}. Therefore, we adopt a \emph{relative definition} of model size. For instance, BERT (110M parameters)~\cite{devlin2019bert} is considered small relative to LLaMA-8B~\cite{dubey2024llama}, while LLaMA-8B is small relative to GPT-4~\cite{openai2023gpt4}. This relative perspective ensures that the concepts discussed in this survey remain applicable as model architectures and scales continue to evolve\footnote{Although we adopt a relative definition of model size, most small models discussed in this survey contain fewer than 1B parameters.}.
}

One may assume that the wide adoption of LLMs makes small models no longer prominent. However, our findings suggest that the usage of small models is significantly underestimated in practical settings. Figure~\ref{fig:model_download}, analyzing the downloads of open-source models of various sizes from HuggingFace, shows that smaller models remain highly popular. This raises fundamental questions about the role of small models in the era of LLMs and their \emph{ecological niche} within the broader AI ecosystems, which is a topic seldom discussed in prior surveys.

To assess the role of SMs, it is essential to compare their strengths and weaknesses relative to LLMs. Table~\ref{tab:comparison_llms_sms} highlights four key dimensions to consider:

 \paragraph{\emph{Accuracy}} 
 LLMs have demonstrated superior performance across a wide range of tasks due to their large number of parameters and extensive training on diverse datasets~\cite{raffel2020exploring, kaplan2020scaling}. Although SMs generally lag behind in overall performance, they can achieve competitive results when enhanced by techniques such as knowledge distillation~\cite{xu2024survey}.

\paragraph{\emph{Generality}} 
LLMs are highly generalizable and capable of handling a broad spectrum of tasks with minimal training examples~\cite{dong2022survey, liu2023pre}. In contrast, SMs are often more specialized, and fine-tuning SMs on domain-specific datasets can outperform general LLMs for specific tasks~\cite{hernandez2023we, juan2024fine, zhang2023sentiment}.

\paragraph{\emph{Efficiency}} 
LLMs require substantial computational resources for both training and inference~\cite{wanefficient2023}, leading to high costs and latency, making them less practical for real-time applications (e.g. information retrieval~\cite{reimers2019sentence}) or resource-constrained environments (e.g. edge devices~\cite{dhar2024empirical}). In contrast, SMs require less training data and computational power, offering competitive performance while significantly reducing resource demands. 

\paragraph{\emph{Interpretability}} 
Smaller, shallower models tend to be more transparent and interpretable than their larger, deeper counterparts~\cite{gilpin2018explaining, barcelo2020model}. In fields such as healthcare~\cite{caruana2015intelligible}, finance~\cite{kurshan2021current}, and law~\cite{eliot2021need}, smaller models are often preferred because their decisions must be easily understandable by non-experts (e.g. doctors, financial analysts).

\begin{table}
	\centering
    \large
	\setlength{\tabcolsep}{1.8mm}{
		\begin{threeparttable} 
			\begin{tabular}{c|c|c}
                \toprule
                    \rowcolor{gray!15}\textbf{Dimension}&\textbf{LLMs}&\textbf{SMs}\cr
				\midrule
                    Accuracy & state-of-the-art \color{deepgreen}{\ding{51}}  &  decent \color{red!80}{\ding{55}} \cr
                    \midrule
                    Generality & general-purpose \color{deepgreen}{\ding{51}} & task-specific \color{red!80}{\ding{55}} \cr
                    \midrule
                    Efficiency & resource-intensive \color{red!80}{\ding{55}}&  resource-efficient \color{deepgreen}{\ding{51}}\cr
                    \midrule
                    Interpretability & low interpretable \color{red!80}{\ding{55}}& high interpretable \color{deepgreen}{\ding{51}}\cr
				\bottomrule
			\end{tabular}
			\caption{Comparisons of different dimensions between LLMs and SMs. } 	\label{tab:comparison_llms_sms}
		\end{threeparttable}
	}
\end{table}

In this work, we systematically examine the role of small models in the era of LLMs from two key perspectives:
(1) \textbf{\emph{Collaboration}} (\textsection 
~\ref{sec:collaboration}). 
LLMs offer superior accuracy and can handle a wide range of tasks, while SMs are more specialized and cost-effective. In practice, the collaboration between LLMs and SMs can strike a balance between performance and efficiency.
(2) \textbf{\emph{Competition and Complementarity}} (\textsection 
~\ref{sec:competition}).
SMs possess distinct advantages, such as simplicity, lower cost, and greater interpretability, and their capabilities can be further enhanced with the help of LLM distillations.
It is crucial to carefully assess the trade-offs between LLMs and SMs based on the specific requirements of the task or application.
We aim for this study to provide actionable insights for practitioners, especially for researchers and businesses with constrained resources, who are interested in leveraging SMs alongside LLMs. Furthermore, we hope our analysis clarifies the ecological niche of small models in society, which offers design principles and ideas that remain relevant even as model architectures and sizes continue to evolve over time.

\section{Related Work}

Recently, there has been a noticeable shift in the research community from large language models to small and efficient language models, as small models can achieve competitive performance while effectively reducing computational and monetary overhead. This trend is now supported by several systematic surveys, which we situate and contrast with our work.
For example, ~\citet{lu2025demystifying} provide an overview of 70 state-of-the-art open-source SMs, with a focus on transformer-based, decoder-only language models (100M-5B parameters).
~\citet{wang2024comprehensive} conduct a comprehensive review of SMs, covering their technical approaches, application scenarios, enhancement methods, collaboration strategies with LLMs, and challenges related to trustworthiness.
~\citet{van2024survey} analyze the architectures, training techniques, and 
compression techniques of SMs. 
Similarly, ~\citet{subramanian2025small} examine \textasciitilde 160 papers about SMs in the 1–8 B parameter range, and discuss how SMs can be used to balance performance, efficiency, and cost.

While existing surveys systematically investigate SMs, these works examine these models in isolation (i.e. their architectures, training, and deployment), rather than emphasizing their relationship with LLMs.
In contrast to these works, our survey emphasizes the relationship between large and small models, which extends beyond mere differences in parameter size. Specifically, we examine how small models complement, compete with, and collaborate with large models across different tasks and deployment scenarios. Through this perspective, we aim to discuss the broader role and ecological niche of small models in our society.

\begin{figure*}[!t]
    \scriptsize
    \centering
\tikzset{
    basic/.style  = {draw, text width=.6cm, align=center, font=\sffamily, rectangle},
    root/.style   = {basic, rounded corners=2pt, thin, align=center, fill=green!30},
    wnode/.style = {basic, rounded corners, thin, align=center, draw=blue!50, line width=1pt, fill=pink!10!blue!80!red!10, text width=6.5em},
    w1node/.style = {basic, rounded corners, thin, align=center, fill=pink!10!blue!70!red!10, text width=27.5em, draw=pink!10!blue!70!red!10, line width=1pt},
    znode/.style = {basic, rounded corners, thin, align=center, fill=orange!15, text width=6.5em, draw=orange!50, line width=1pt},
    z1node/.style = {basic, rounded corners,thin, align=center, fill=orange!15, text width=27.5em, draw=orange!15, line width=1pt},
    edge from parent/.style={draw=black, edge from parent fork right},
}

\begin{forest} for tree={
    grow=east,
    growth parent anchor=west,
    parent anchor=east,
    child anchor=west,
    edge={gray, thick},
    l sep=20pt, 
    s sep=12pt, 
    edge path={%
        \noexpand\path [\forestoption{edge}] 
          (!u.parent anchor) -- (.child anchor)
          \forestoption{edge label};
    },
}
[\rotatebox{90}{Collaboration}\\ (\textsection \ref{sec:collaboration}), basic,  l sep=5mm,
    [LLMs Enhance SMs \\ (\textsection \ref{sec:LLM_enhance_SM}), wnode,  l sep=10mm,
    [Knowledge Distillation, wnode
        [\textbf{Representation Distillation} (\textsection \ref{sec:repre_distill}), w1node]
        [\textbf{Data Augmentation} (\textsection \ref{sec:data_aug}), w1node]
        [\textbf{Rationale Distillation} (\textsection \ref{sec:ration_distill}), w1node]
        [\textbf{Full Data Synthesis} (\textsection \ref{sec:train_data_syn}), w1node]
    ]] 
    [SMs Enhance LLMs \\ (\textsection \ref{sec:sms_enhance_llms}), znode, l sep=10mm,
        [Evaluating LLMs\\ (\textsection \ref{sec:eval_llms}), znode
            [\textbf{Using SMs to evaluate LLM's generations}, z1node]
        ]
        [Efficient Inference\\ (\textsection \ref{sec:eficient_inference}), znode 
            [\textbf{Speculative Decoding},  z1node]
            [\textbf{Model Routing}, z1node]
            [\textbf{Model Cascading},  z1node]
        ]
        [Augmented Reasoning\\ (\textsection \ref{sec:augmented_reasoning}), znode
            [\textbf{Deficiency Repair}, z1node]
            [\textbf{Prompt Engineering}, z1node]
            [\textbf{Domain Adaptation}, z1node]
            [\textbf{Retrieval Augmented Generation}, z1node]
        ]
        [Data Curation\\ (\textsection \ref{sec:data_curation}), znode
            [\textbf{Weak-to-Strong Paradigm}, z1node],
            [\textbf{Curating Instruction-Tuning Data}, z1node],
            [\textbf{Curating pre-training data}, z1node]
        ]]
    ]
]
\end{forest}
    \caption{Collaborations between LLMs and SMs}
    \label{fig:collaboration}
\end{figure*}

\section{Collaboration}~\label{sec:collaboration}

In the following, we describe a dual collaboration paradigm in which SMs and LLMs mutually enhance one another to optimize resource usage and enhance reasoning capabilities. In Section \ref{sec:sms_enhance_llms}, we explore how SMs can strengthen LLMs by guiding training, inference, evaluation, and robustness. Conversely, in Section \ref{sec:LLM_enhance_SM}, we examine how LLMs can, in turn, help SMs by providing richer supervision. The complete interaction framework is illustrated in Figure \ref{fig:collaboration}, showing how these two types of models interact and collaborate toward more efficient, powerful systems.

\subsection{Small Models Enhance LLMs}~\label{sec:sms_enhance_llms}

To better understand how SMs enhance LLMs, we organize this section according to the life cycle of LLMs: data preparation (pre-training and fine-tuning), inference, and evaluation. 
First, at the data preparation stage (see Section~\ref{sec:data_curation}), SMs contribute to curating pre-training corpora and instruction-tuning datasets,
which can filter low-quality or undesired content, and enable weak-to-strong training paradigms that improve supervision quality. 
Next, at the inference stage, SMs improve reasoning (See Section~\ref{sec:augmented_reasoning}) and efficiency (See Section~\ref{sec:eficient_inference}) in practical deployment through techniques such as retrieval augmentation, domain adaptation,  model routing, and cascading, which help reduce computational cost while maintaining or even enhancing reasoning performance. 
Finally, at the evaluation stage (see Section~\ref{sec:eval_llms}), SMs serve as lightweight verifiers to assess, diagnose, and repair LLM outputs, increasing reliability and robustness. 
From this life-cycle perspective, SMs are not simply smaller alternatives to LLMs, but useful tools that help improve LLMs at every stage of the pipeline.

\subsubsection{Data Curation}~\label{sec:data_curation}

\begin{figure*}
    \scriptsize
    \centering
\tikzset{
    basic/.style  = {draw, text width=.6cm, align=center, font=\sffamily, rectangle},
    root/.style   = {basic, rounded corners=2pt, thin, align=center, fill=green!30},
    wnode/.style = {basic, rounded corners, thin, align=center, draw=gray!35, fill=gray!5, line width=1pt, text width=7.5em},
    w1node/.style = {basic, rounded corners, thin, align=center, fill=gray!5, text width=7.5em, draw=gray!35, line width=1pt},
    w2node/.style = {basic, rounded corners, thin, align=center, fill=gray!5, text width=15.5em, draw=gray!35, line width=1pt},
    edge from parent/.style={draw=black, edge from parent fork right},
}

\begin{forest} for tree={
    grow=east,
    growth parent anchor=west,
    parent anchor=east,
    child anchor=west,
    edge={gray, thick},
    l sep=20pt, 
    s sep=12pt, 
    edge path={%
        \noexpand\path [\forestoption{edge}] 
          (!u.parent anchor) -- (.child anchor)
          \forestoption{edge label};
    },
}
    [Data Curation \\ (\textsection~\ref{sec:data_curation}), wnode,  l sep=10mm,
        [Weak-to-Strong Paradigm, wnode,  l sep=10mm,
        [\tiny Test-Time Alignment, w2node]
        [\tiny Data Annotation, w2node],
        ] 
        [Fine-Tuning Data \\ , wnode
            [\tiny Alignment Data, w2node]
            [\tiny Task-Specific Fine-tuning Data, w2node],
        ]
        [Pre-training Data \\ , wnode
            [Data Reweighting, 
            w1node
            [\tiny Instance-Level Methods, w2node],
            [\tiny Domain-Level Methods, w2node]
            ]
            [Data Selection, 
            w1node
            [\tiny Privacy, w2node]
            [\tiny Toxicity, w2node]
            [\tiny Duplication, w2node],
            [\tiny Low-Quality Content, w2node]
            ]
        ]] 
\end{forest}
    \caption{Taxonomy of data curation}
    \label{fig:data_curation}
\end{figure*}

In the era of LLMs, a prevailing ideology is ``more is better'' --- more data,  more parameters, more GPUs. Indeed, the scaling laws reveal that model performance improves with both more parameters and tokens~\cite{kaplan2020scaling}. However, as the data availability reaches its limits, a new paradigm is emerging: rather than simply using ever-larger datasets, the key lies in curating those texts with refined strategies. From filtering noisy web text during pre-training to selecting a small, high-quality set for fine-tuning, recent work demonstrates that less but more refined data enables more efficient and powerful training, and better alignment with human purpose~\cite{albalak2024survey}.
This shift follows a weak-to-strong learning paradigm, in which rich but noisy signals give way to smaller, higher-quality supervision~\cite{burnsweak}.

In the following, we present how to use small models to curate data from several aspects: pre-training data, instruction-tuning data, and the weak-to-strong paradigm, as shown in Figure~\ref{fig:data_curation}.

\paragraph{\textbf{Curating Pre-training Data}}
The reasoning capabilities of LLMs are largely attributed to their pre-training on extensive and diverse datasets, typically sourced from web scrapes, books, and scientific literature. 
Since expanding the quantity and diversity of these training datasets enhances the generalization ability of LLMs, significant efforts have been made to compile large-scale and diverse pre-training corpora, such as C4~\cite{raffel2020exploring} and Pile~\cite{gao2020pile}.
However, the idea of creating ever-larger datasets faces a significant challenge: data availability is finite, and there is a looming possibility that public human text data could soon be exhausted~\cite{villalobosposition}.
Moreover, not all data contributes equally to model performance; web-scraped content often includes noise and low-quality text. 
This has led to a paradigm shift from focusing purely on the quantity of data to prioritizing the quality of data. Recent research supports the notion that ``less is more''~\cite{marion2023less}, advocating for data selection or pruning techniques to curate high-quality subsets from large datasets, thereby enhancing model performance. We can divide these approaches into two categories, as shown in Figure~\ref{fig:data_curation}: (1) Data Selection, and (2) Data Reweighting.

\textbf{(1) Data Selection} uses a small model to identify and curate a high-quality subset of the raw pre-training corpus. 
Earlier approaches largely relied on manual, rule-based heuristics such as blacklist filtering and MinHash deduplication to remove undesirable or duplicated text~\cite{raffel2020exploring, tirumala2024d4, penedo2023refinedweb, wenzek2020ccnet}.
Recognizing the limitations of hand-crafted rules, more recent methods use small proxy models to assess content quality, which enables the selection of high-quality subsets. For instance, a simple classifier can be trained to assess content quality, focusing on the removal of \textit{noisy}~\cite{brown2020language, du2022glam, chowdhery2023palm}, \textit{toxic}~\cite{Arnett2024ToxicityOfTheCommons,Kamphuis2024TinyToxicDetector}, and \textit{private}~\cite{subramani-etal-2023-detecting,yu2023selective} data.
For example, FineWeb-Edu~\cite{penedo2024fineweb} trains a linear regression model on text embeddings to select high-quality data dedicated to the educational field.
Moreover, a small language model may compute perplexity scores as a proxy for textual coherence and quality~\cite{wenzek2020ccnet, marion2023less}. 
Other recent methods introduce the \textit{Importance Resampling} frameworks that select a subset of a large raw unlabeled dataset to match a desired target distribution~\cite{xie2023data}.
Data selection reduces the dataset size but aims to raise the average usefulness of training tokens.

Another important role of small models in pre-training data curation is language identification (LID).
When constructing multilingual corpora, accurately identifying the main language of each document is a critical step. 
Because this process must be applied to billions of documents, LID systems need to be both computationally efficient and inexpensive to run.
Consequently, lightweight classifiers are commonly employed to filter and select texts for specific languages, which is particularly important for ensuring adequate coverage of low-resource languages~\cite{kargaran2023glotlid,burchell2023open,nllb2024scaling}.
Recent large-scale datasets such as FineWeb~\cite{penedo2024fineweb,penedo2025fineweb2} and HPLT~\cite{oepen2025hplt} rely on such LID models to curate high-quality multilingual resources for training LLMs.

\textbf{(2) Data Reweighting} goes beyond simple filtering by adjusting sampling probabilities. The key idea is the use of a \emph{small proxy model} whose job is to estimate how much weight should be assigned to each data partition. We can divide reweighting into two key categories: (1) \textit{Domain-Level Methods.} 
In this approach, a small proxy model is trained over different domains or sources, e.g., Wikipedia, books, web texts, to evaluate the performance or loss across each domain. 
For example, the framework DoReMi~\cite{xie2024doremi} uses a 280 million parameter proxy model to compute domain weights for training an 8B parameter model, improving downstream accuracy by 6.5 points and reaching baseline accuracy 2.6× faster. 
Meanwhile, AutoScale~\cite{Kang2024AutoScale} computes optimal domain compositions at small scale and then uses a predictor to extrapolate to large scale.
Finally, small agents are introduced to learn how to reweight each domain via reinforcement learning~\cite{Yang2025DataMixingAgent}.
(2) \textit{Instance-Level Methods.}
Here the focus is on how to assign weights to individual examples (instances), and again a small model often plays a key role in estimating how informative or representative these instances are. For example, the method \texttt{PRESENCE}~\cite{thakkar2023self} uses a proxy model to compute self-influence scores, \colortext{which approximate how much giving a training example more weight during training would affect its own loss. Since these scores serve as a proxy for the degree to which an example is memorized by the model, they are subsequently used to guide instance-level data reweighting during pre-training.}
Small proxy models here serve as a cheap but predictive surrogate of how a large model would perform, and their assessments guide the weight assignments, which helps make data usage more efficient.

\paragraph{\textbf{Curating Fine-Tuning Data}}
LLMs acquire substantial knowledge through pre-training, and fine-tuning is used to adapt these capabilities either to specific tasks or to align them with human preferences~\cite{ouyang2022training,bai2022training}. In this light, it is useful to distinguish two types of data curation. 
(1) \textit{Task-Specific Fine-Tuning Data.}
This category of data is used for downstream tasks or specific domains such as classification and domain-specific QA. The curation focuses on selecting samples that are representative of the target distribution and avoid redundancy. A small language model can be leveraged as a data-selector. For example, we might use a smaller model to compute embedding similarity for each candidate data point, and then rank and select the most representative examples for fine-tuning while discarding the less informative ones~\cite{liu2025rethinking}. Another approach is to use the internal signals of a small model such as attention patterns to score data points, and then select the most informative subset for a specific task~\cite{wang2025datawhisperer}.
(2) \textit{Alignment Data.}
The goal is to steer the model toward desired instructions and align with human preferences such as honesty and safety. Here, data selection focuses more on using small models to efficiently collect high-quality instruction-response pairs rather than simply increasing quantity.
Specifically, the study, \textit{Less is More for Alignment}, demonstrates that fine-tuning on just 1,000 carefully curated instruction examples can yield a well-aligned model~\cite{zhou2024lima}.
This highlights the importance of selecting high-quality data for efficient instruction tuning~\cite{longpre2023flan,chen2023alpagasus}.
Model-oriented data selection (MoDS)~\cite{du2023mods} is one approach that employs a small language model, DeBERTa~\cite{hedeberta2021}, to evaluate instruction data based on quality, coverage, and necessity. Additionally, the LESS framework~\cite{xialess2024} demonstrates that smaller models can be used to select influential data not only for larger models but also for models from different families. This underscores the potential of using targeted data selection techniques to optimize instruction tuning processes.

\paragraph{\textbf{Weak-to-Strong Paradigm}}
LLMs are typically aligned with human values through reinforcement learning with human feedback (RLHF), where behaviors favored by humans are rewarded, and those rated poorly are penalized~\cite{shen2023large}.
However, as LLMs continue to evolve and surpass human capabilities in various tasks, they are becoming \emph{superhuman models}, capable of performing complex and creative tasks that may exceed human understanding. 
For instance, these models can generate thousands of lines of specialized code, engage in intricate mathematical reasoning, and produce lengthy, creative novels. Evaluating the correctness and safety of such outputs poses significant challenges for human evaluators.
This scenario introduces a new paradigm for aligning superhuman models, termed \emph{weak-to-strong generalization}, which involves using weaker (smaller) models as supervisors for stronger (larger) models~\cite{burnsweak}.
The goal is for strong models to learn from limited or imperfect signals and generalize beyond their weaker teachers' capabilities.
We categorize recent advances in weak-to-strong generalization into two primary classes: (1) Data Annotation, and (2) Test-Time Alignment.

\textbf{(1) Data Annotation.} This foundational approach fine-tunes a large model using labels or preferences generated by weaker models.
~\citet{burnsweak} formalize the paradigm and demonstrate that strong models can extrapolate from weak supervision.
~\citet{guo2024improving} introduce an approach that enhances weak-to-strong generalization by incorporating reliability estimation across multiple answers provided by weak models. This method improves the alignment process by filtering out uncertain data or adjusting the weight of reliable data.
Beyond data labeling, weak models can also collaborate with large models during the inference phase to further enhance alignment.
Instead of relying on a single weak teacher, recent methods employ ensembles of weak or specialized models to improve label quality and diversity.
For example, ~\citet{liu2024co} suggest using a diverse set of specialized weak teachers, rather than relying on a single generalist model, to collectively supervise the strong student model.
~\citet{lang2025debate} introduce debate mechanisms between weaker models to refine signals before training a strong model.
This weak-to-strong paradigm is not limited to language models but has also been extended to vision foundation models~\cite{guo2024vision}.

\textbf{(2) Test-Time Alignment.}
Beyond training-time labeling, weak models can help strong models during inference to refine responses dynamically.
Weak-to-Strong Search~\cite{zhou2024weak} approaches the alignment of a large model as a test-time greedy search, aiming to maximize the log-likelihood difference between small tuned and untuned models, which function as a dense reward signal and a critic, respectively. 
Aligner~\cite{ji2024aligner} employs a small model to learn the correctional residuals between preferred and dispreferred responses, enabling direct application to various upstream LLMs for aligning with human preferences.

\medskip
\noindent\textbf{Summary}

Given the limits of high-quality human-generated data, scaling language models can no longer rely solely on collecting more text. Instead, progress increasingly depends on curating existing data more effectively and embracing the principle that ``less is more''. 
In this context, SMs play a crucial role throughout both pre-training and fine-tuning. During pre-training, SMs support data selection by filtering low-quality, duplicated, toxic, or privacy-sensitive content, and by reweighting domains or instances to better shape the learning distribution.
During fine-tuning, SMs can help construct task-specific and alignment datasets, improving annotation quality and reducing reliance on expensive human supervision.

As LLM capabilities continue to improve, human annotation becomes increasingly difficult. 
Stronger models may exceed human ability in certain reasoning tasks, which makes direct supervision less reliable. The Weak-to-Strong paradigm addresses this challenge by showing that weaker supervisors (including small models) can still guide stronger models effectively. By carefully designing supervision signals and alignment objectives, SMs can help extract knowledge from powerful models and contribute to building reliable reward models.

In practice, the benefits of using SMs for data curation are threefold. 
(1) Less is More. 
Small models can reduce dataset size while retaining or even improving quality for LLM training.
(2) Cost Saving. When human annotation is costly and time-consuming, smaller models provide a scalable and economical alternative.
(3) More specialized Datasets.  Small models can ensure high-relevance data, particularly for task- or domain-specific domains, such as low-resource multilingual, legal, and medical tasks.

\medskip
\noindent\textbf{Future Directions}

(1) Humans can learn from relatively few examples, which contrasts with the massive data requirements of current LLM pre-training paradigms~\cite{warstadt2023findings}. It is therefore crucial to investigate methods that use less data to obtain better reasoning models, and to explore how small models in particular can play a key role in achieving this goal.
\\[2ex]
(2) While data curation offers clear advantages, LLMs still have a tendency to produce hallucinated and toxic content. Moreover, removing low-quality or toxic text could potentially degrade certain capabilities, such as generality~\cite{longpre2024pretrainer}. Therefore, it is essential to define more nuanced criteria for evaluating data quality, including dimensions like factuality, safety, and diversity~\cite{wettigqurating,liumakes2024}. Investigating the use of small models to develop effective and efficient data selection methods is a valuable area of study.\\[2ex]
(3) Synthetic data serves as a valuable supplement to the limited amount of human-generated data~\cite{long2024llms}, yet the potential of small models in curating synthetic data remains largely unexplored.\\[2ex]
(4) We tend to choose data that is clear and representative for alignment tuning, but including a subset of overly difficult examples can harm alignment performance~\cite{gao2025principled}. It is therefore crucial to dynamically estimate a model’s current capacity and then tailor the selection so it contains an appropriate mix of easy and hard samples.\\[2ex]
(5) While the weak-to-strong framework is effective in eliciting knowledge from stronger models, it is still far from recovering the full performance gap between weak and strong models. It is crucial to ensure that the strong model has a deep, intuitive understanding of the task at hand, is capable of correcting the weak model's errors, and naturally aligns with the objectives of the task~\cite{burnsweak}. Future work should focus on identifying properties and methods that help achieve this goal. 
\\[2ex]
(6) The current understanding of weak-to-strong generalization is limited. Researchers should develop a deep understanding of the underlying mechanisms that govern the success or failure of alignment methods, e.g. theoretical analysis~\cite{lang2024theoretical}, errors in weak supervision~\cite{guo2024improving}, and extrapolating generalization errors using scaling laws~\cite{kaplan2020scaling}.

\subsubsection{Augmented Reasoning}~\label{sec:augmented_reasoning}

\begin{figure*}
    \scriptsize
    \centering
\tikzset{
    basic/.style  = {draw, text width=.6cm, align=center, font=\sffamily, rectangle},
    root/.style   = {basic, rounded corners=2pt, thin, align=center, fill=green!30},
    wnode/.style = {basic, rounded corners, thin, align=center, draw=gray!35, fill=gray!5, line width=1pt, text width=7.5em},
    w1node/.style = {basic, rounded corners, thin, align=center, fill=gray!5, text width=7.5em, draw=gray!35, line width=1pt},
    w2node/.style = {basic, rounded corners, thin, align=center, fill=gray!5, text width=15.5em, draw=gray!35, line width=1pt},
    edge from parent/.style={draw=black, edge from parent fork right},
}

\begin{forest} for tree={
    grow=east,
    growth parent anchor=west,
    parent anchor=east,
    child anchor=west,
    edge={gray, thick},
    l sep=20pt, 
    s sep=12pt, 
    edge path={%
        \noexpand\path [\forestoption{edge}] 
          (!u.parent anchor) -- (.child anchor)
          \forestoption{edge label};
    },
}
    [Augmented Reasoning \\ (\textsection~\ref{sec:augmented_reasoning}), wnode,  l sep=10mm,
        [Deficiency Repair \\ , wnode
            [\tiny Small Model Plugins, w2node],
            [\tiny Contrastive Decoding, w2node],
        ], 
        [Prompt Engineering \\ , wnode
            [\tiny Prompt Retrieval; Design;  Verification, w2node]
        ], 
        [Domain Adaptation \\ , wnode
            [\tiny Black-Box Adaptation, w2node],
            [\tiny White-Box Adaptation, w2node]
        ], 
        [Retrieval Augmented Generation \\ , wnode
            [\tiny Other sources, w2node]
            [\tiny Structured Knowledge, w2node],
            [\tiny Textual Document, w2node]
        ]] 
\end{forest}
    \caption{Taxonomy of augmented reasoning}
    \label{fig:augmented_reasoning}
\end{figure*}

Despite their impressive scale, LLMs rely on the internal knowledge encoded in their parameters during the pre-training stage. 
While this internalized knowledge enables strong generalization across many tasks, it is inherently limited. The model may lack up-to-date information~\cite{Kasai2022RealTimeQA}, domain-specific expertise~\cite{Feng2023Trends}, or robust multi-step reasoning ability~\cite{Mondorf2024BeyondAccuracy}. 
As a result, when a task exceeds the model’s internal knowledge capacity~\cite{li-etal-2025-knowledge-boundary}, LLMs can generate responses that appear plausible but are factually incorrect or hallucinated~\cite{huang2025survey}.

From this perspective, many recent advances can be understood as efforts to bridge the gap between internal knowledge (what the model has memorized) and external knowledge (information and guidance introduced at run time). When the task falls beyond the model's internal capacity, it is necessary to augment reasoning by using external signals. 
In this section, we organize existing approaches into four categories: (1) Retrieval-Augmented Generation, which enriches internal knowledge with retrieved documents. 
(2) Domain Adaptation, which reshapes internal knowledge toward specialized distributions. 
(3) Prompt Engineering, which guides the model to better use its existing knowledge. 
(4) Deficiency Repair, which identifies and corrects systematic weaknesses. 
Together, these strategies illustrate different ways of balancing internal and external knowledge to improve reasoning reliability.

\paragraph{\textbf{Retrieval Augmented Generation}}
LLMs exhibit impressive reasoning capabilities, yet their ability to memorize specific knowledge is somewhat limited. Consequently, LLMs may struggle with tasks that require domain-specific expertise or up-to-date information. To address these limitations, Retrieval-Augmented Generation (RAG) enhances LLMs by employing a lightweight retriever to find relevant document fragments from external knowledge bases, document collections, or other tools~\cite{gao2023retrieval,lewis2020retrieval}. By incorporating external knowledge, RAG effectively mitigates the issue of generating factually inaccurate content, often referred to as hallucinations~\cite{shuster2021retrieval}. RAG methods can be broadly categorized into three types based on the nature of the retrieval source.

\textbf{(1) Textual Document} is the most commonly used retrieval sources in RAG methods, encompassing resources such as Wikipedia~\cite{trivedi2023interleaving,asaiself2023}, cross-lingual translations~\cite{nie2023cross,wang2024retrieval} and domain-specific corpus (e.g. medical~\cite{xiong2024benchmarking} and legal~\cite{yue2023disc} domains).
These approaches generally employ lightweight retrieval models, such as sparse BM25~\cite{robertson2009probabilistic} and dense BERT-based~\cite{izacard2021unsupervised} retrievers, to extract relevant text from these sources.

\textbf{(2) Structured Knowledge} encompasses sources such as knowledge bases and databases, which are typically verified and can provide more precise information. For example, KnowledgeGPT~\cite{wang2023knowledgpt} enables LLMs to retrieve information from knowledge bases, while T-RAG~\cite{pan2022end} enhances answers by concatenating retrieved tables with the query. StructGPT~\cite{jiang2023structgpt} further augments generation by retrieving from hybrid sources, including knowledge bases, tables, and databases. The retriever in these methods can be a lightweight entity linker, query executor, or API.

\textbf{(3) Other Sources} include code, tools, and even images, which enable LLMs to leverage external information for enhanced reasoning.
For instance, DocPrompting~\cite{zhoudocprompting} employs a BM25 retriever to obtain relevant code documentation before code generation. Similarly, Toolformer~\cite{schick2024toolformer} demonstrates that LMs can self-learn to use external tools, such as translators, calculators, and calendars, through simple APIs, leading to significant performance improvements.

\paragraph{\textbf{Domain Adaptation}}
General-purpose LLMs still require further customization to achieve optimal performance in specific use cases (\textit{e.g.} coding) and domains (\textit{e.g.} medical tasks)~\cite{Feng2023Trends}. While fine-tuning on specialized data is one approach to adapting LLMs, this process has become increasingly resource-intensive, and in some cases, it is not feasible—especially when access to internal model parameters is restricted, as with models like ChatGPT~\cite{Ling2023DomainSpecializationLLM}. Recent research has explored adapting LLMs using smaller models, which can be categorized into two approaches: White-Box Adaptation and Black-Box Adaptation, depending on whether access to the model's internal states is available.

\textbf{(1) White-Box Adaptation} typically involves fine-tuning a small model to adjust the token distributions of frozen LLMs for a specific target domain.
For instance, CombLM~\cite{ormazabal2023comblm} learns a linear function to combine the probability distributions from the large black-box model with those from a smaller domain-specific expert model. IPA~\cite{lu2023inference} introduces a lightweight adapter that tailors a large model toward desired objectives during decoding without requiring fine-tuning. IPA achieves this by optimizing the combined distribution using reinforcement learning. Proxy-tuning~\cite{liu2024tuning} fine-tunes a smaller language model, contrasting the probabilities between the tuned model (the expert) and its untuned version (the anti-expert) to guide the larger base model.
These approaches only modify the parameters of small domain-specific experts, allowing LLMs to be adapted to specific domain tasks. However, white-box adaptation is not applicable to API-only modeling services, where access to internal model parameters is restricted.

\textbf{(2) Black-Box Adaptation} involves using a small domain-specific model to guide LLMs toward a target domain by providing textual relevant knowledge.
Retrieval Augmented Generation (RAG) can extract query-relevant knowledge from an external document collection or knowledge base, and thus enhance general LLMs
by leveraging their in-context learning ability. 
It involves first
using a lightweight retriever to find relevant content from the domain corpus, which is then incorporated into the LLM’s input to improve its understanding of domain-specific knowledge~\cite{siriwardhana2023improving,shi2023replug,gao2023retrieval}. 
Another approach employs small expert models to generate background knowledge for the base LLM. For example, BLADE~\cite{li2024blade} and Knowledge Card~\cite{fengknowledge} first pre-train a small expert model on domain-specific data, which then generates expertise knowledge in response to a query, thereby enhancing the base LLM's performance. MedAdapter~\cite{Shi2024MedAdapter} fine-tunes a small BERT-sized adapter to rank candidate solutions generated by LLMs.

\paragraph{\textbf{Prompt Engineering}}
Prompt-based learning is a prevalent paradigm in LLMs where prompts are crafted to facilitate few-shot or even zero-shot learning, enabling adaptation to new scenarios with minimal or no labeled data~\cite{liu2023pre}. 
This approach leverages the power of In-Context Learning (ICL)~\cite{dong2022survey}, which operates without performing parameter updates. Instead, it relies on a prompt context that includes a few demonstration examples structured within natural language templates.

In this learning process, small models can be employed to enhance prompts, thereby improving the performance of larger models. 
For instance, Uprise~\cite{cheng2023uprise} optimizes a lightweight retriever that autonomously retrieves prompts for zero-shot tasks, thereby minimizing the manual effort required for prompt engineering. Similarly, DaSLaM~\cite{juneja2023small} uses a small model to break down complex problems into subproblems that necessitate fewer reasoning steps, leading to performance improvements in larger models across multiple reasoning datasets.

\colortext{Beyond prompt retrieval and decomposition, small models can also enhance in-context learning (ICL), a widely used prompt engineering technique, by improving the quality of demonstrations included in prompts. For example, some methods fine-tune small models to generate pseudo labels for input examples~\cite{xu2023small,lee2024can}, which produce higher-quality demonstrations that lead to better ICL performance than the original prompts. Furthermore, in multi-stage prompting pipelines, small models can verify~\cite{hsu2024calm} or rewrite~\cite{vernikos2024small} the outputs of LLMs, further improving response quality without requiring LLM fine-tuning.
}

\paragraph{\textbf{Deficiency Repair}}
Powerful LLMs may generate repeated, untruthful, and toxic content, and small models can be used to repair these defects. We introduce two ways to achieve this goal: contrastive decoding and small model plug-ins. 

\textbf{(1) Contrastive Decoding} exploits the contrasts between a larger model (expert) and a smaller model (amateur) by choosing tokens that maximize their log-likelihood difference.  Existing work has explored the synergistic use of logits from both LLMs and SMs to reduce repeated text~\cite{li2023contrastive}, mitigate hallucinations~\cite{sennrich2024mitigating}, augment reasoning
capabilities~\cite{o2023contrastive} and safeguard
user privacy~\cite{zhang2024cogenesis}. 
Since fine-tuning LLMs is computing-intensive, proxy tuning proposes fine-tuning a small model and contrasting the difference between the original LLMs and small models to adapt to the target task~\cite{liu2024tuning}.

\textbf{(2) Small Model Plugins} fine-tune a specialized small model to address some of the shortcomings of the larger model. 
For example, the performance of LLMs may degrade when encountering out-of-vocabulary (OOV) words. 
To address this issue, we can train a small model to mimic the behavior of the large model and impute representations for unseen words~\cite{pinter2017mimicking,chen2022imputing}. In this way, we can make large models robust with little cost. 
Additionally, LLMs may generate hallucinated texts and we can train a small model to detect hallucinations~\cite{cheng2024small} or calibrate confidence scores~\cite{chen2024reconfidencing}.

\medskip
\noindent\textbf{Summary}

LLMs store parametric knowledge, i.e., facts and patterns learned in their weights during pre-training, but this kind of knowledge is static and limited in depth for domain-specific tasks, leading to plausible yet hallucinated outputs.
When a task exceeds what an LLM can reliably know or reason about on its own, it becomes important to integrate external knowledge and auxiliary reasoning mechanisms. 
In practice, four strategies using smaller models to augment LLM reasoning have been introduced in this section: Retrieval-Augmented Generation (RAG), domain adaptation, prompt engineering, and deficiency repair.

In real-world settings, practitioners can choose among them based on task requirements and the nature of the external knowledge involved. 
\begin{itemize}
    \item When the task demands up-to-date or domain-specific facts, RAG is often the primary and low-cost strategy.
    Retrieving relevant information at inference time and prioritizing grounded sources rather than static parametric knowledge. This approach reduces hallucinations and allows real-time integration of new information without retraining the LLM.
    \item When a task requires adapting an LLM to a specialized domain but full fine-tuning is expensive or impractical, domain adaptation via small models offers a cost-effective alternative. 
    Domain-expert models can capture targeted, high-precision knowledge and provide structured guidance to the general-purpose LLM. Rather than modifying the large model's parameters, these small experts inject domain-specific signals at the inference stage.
    \item When the input pattern is not aligned with the distribution of parametric knowledge, prompt engineering can be introduced to guide the LLM's reasoning process. Optimized prompts can surface relevant external knowledge more effectively and help steer the model toward correct inference patterns without high computational cost.
    \item When outputs require specific properties, such as safety, interpretability, or uncertainty expression, deficiency repair mechanisms can be employed to address these limitations. 
    Here, small auxiliary models evaluate or adjust the LLM's outputs, effectively correcting or flagging problematic content.
\end{itemize}


\medskip
\noindent\textbf{Future Directions}

(1) The retrieval augmented text generation performance is very sensitive to the retrieval quality~\cite{yoranmaking}. 
However, the widely-used dense-retrieval systems, i.e., a small embedding model with a similarity calculation, have theoretical limitations. 
In particular, their ability to represent different combinations of relevant documents can be constrained by the embedding dimension~\cite{Weller2025Theoretical}. Therefore, more expressive and efficient retrieval approaches may be needed for advanced retrieval settings.
\\[2ex]
(2) LLMs are still good at memorizing patterns rather than true logical reasoning. Recent findings indicate that LLMs struggle with classic reasoning problems and fail to generalize beyond the patterns seen in their training data, such as deductive reasoning~\cite{SealsShalin2024Deductive} and compositional reasoning~\cite{Sinha2024Compositional}.
It is important to explore how to use small models to enhance the logical reasoning capability of LLMs.
\\[2ex]
(3) Current domain adaptation methods often necessitate pre-training or fine-tuning a domain-specific expert, which is impractical for resource-constrained tasks. 
Exploring efficient methods to transfer and adapt models across many domains in low-resource settings is a valuable area of research~\cite{sun2024bbox}.


\subsubsection{Efficient Inference}~\label{sec:eficient_inference}

LLMs have demonstrated remarkable capabilities, but their gains in performance come with substantial costs. 
Increasing model size typically entails slower inference, higher API costs, and significant environmental and energy burdens~\cite{wu2022sustainable}. 
As model deployment scales, these costs become not only financial concerns but also sustainability challenges.

In contrast, smaller models offer clear efficiency advantages: they are faster, cheaper to run, and more energy-efficient. 
While they may not match the performance of their larger counterparts, many real-world queries do not require the full capacity of a large model.
User requests span a broad spectrum of difficulties, which suggests that uniform reliance on large models is often unnecessary.

This observation motivates a core idea in efficient inference. Allocating computational resources adaptively rather than uniformly. 
By orchestrating a list of models with different capacities, systems can use expensive large models for genuinely hard queries while allowing smaller models to handle simpler ones. 
Such adaptive allocation preserves performance where needed while substantially reducing overall cost.
Broadly, existing approaches to this paradigm fall into three categories: model cascading, model routing, and speculative decoding, as illustrated in Figure~\ref{fig:efficient_inference}.


\begin{figure*}
    \scriptsize
    \centering
\tikzset{
    basic/.style  = {draw, text width=.6cm, align=center, font=\sffamily, rectangle},
    root/.style   = {basic, rounded corners=2pt, thin, align=center, fill=green!30},
    wnode/.style = {basic, rounded corners, thin, align=center, draw=gray!35, fill=gray!5, line width=1pt, text width=7.5em},
    w1node/.style = {basic, rounded corners, thin, align=center, fill=gray!5, text width=7.5em, draw=gray!35, line width=1pt},
    w2node/.style = {basic, rounded corners, thin, align=center, fill=gray!5, text width=15.5em, draw=gray!35, line width=1pt},
    edge from parent/.style={draw=black, edge from parent fork right},
}

\begin{forest} for tree={
    grow=east,
    growth parent anchor=west,
    parent anchor=east,
    child anchor=west,
    edge={gray, thick},
    l sep=20pt, 
    s sep=12pt, 
    edge path={%
        \noexpand\path [\forestoption{edge}] 
          (!u.parent anchor) -- (.child anchor)
          \forestoption{edge label};
    },
}
    [Efficient Inference \\ (\textsection~\ref{sec:eficient_inference}), wnode,  l sep=10mm,
        [Speculative Decoding \\ , wnode
            [\tiny Draft-then-Verify Pipeline, w2node],
        ], 
        [Model Routing \\ , wnode
            [\tiny Training-required  Methods, w2node],
            [\tiny Training-free Methods, w2node]
        ], 
        [Model Cascading \\ , wnode
            [\tiny Agreement-based Methods, w2node]
            [\tiny Cost-aware Methods, w2node]
            [\tiny Quality-based Methods, w2node],
            [\tiny Confidence-based Methods, w2node]
        ]] 
\end{forest}
    \caption{Taxonomy of efficient inference}
    \label{fig:efficient_inference}
\end{figure*}

\paragraph{\textbf{Model Cascading}} This process involves the sequential use of multiple models to make predictions or decisions, where each model in the cascade has a different level of complexity. 
The output of one model may trigger the activation of the next model in the sequence. This approach allows for the collaboration of models of varying sizes, enabling smaller models to handle simpler input queries while transferring more complex tasks to larger models. 
The critical step in this process is determining whether a given model is capable of addressing the input question. This method effectively optimizes inference speed and reduces financial costs.

\textbf{(1) Confidence-based Methods.} The smaller model produces a confidence score or probability and, if that score falls below a threshold, the query is forwarded. For instance, 
this work, query-level uncertainty~\cite{Chen2025QueryLevelUncertainty}, proposes a training-free method to estimate whether a model is capable of addressing a given query without generating any answers, which can achieve effective model cascading while reducing inference costs.  Confidence-based methods aim to assess the knowledge gap, and then allocate a task to the model with an appropriate size~\cite{chen2024data,EnomotoEda2021LearningToCascade}.

\textbf{(2) Quality-based Methods.} The system evaluates the quality of the output by using a verifier model and triggers escalation if the output quality falls short. For example, some existing techniques train a small evaluator to assess the output correctness~\cite{kag2023efficient,chen2020frugalml,chen2023frugalgpt}, thereby deciding whether to escalate the query to a more complex model.

\textbf{(3) Cost-aware Methods.} Beyond simple correctness, the trigger mechanism may incorporate cost-aware objectives such as latency or monetary costs~\cite{wang2017idk}. For example, we can study various models with different sizes, comparing their accuracy, financial cost, and latency,  and then select which model to use based on budget or latency limitation~\cite{Wang2024CascadeAwareTraining,Zhang2024EfficientContextualLLMCascades}.

\textbf{(4) Agreement-based Methods.} Given that LLMs can perform self-verification~\cite{dhuliawala2023chain} and provide confidence levels in their responses~\cite{tian2023just}, we can use the consistency between multiple answers as a trigger.  If the outputs disagree or diverge, the task is deferred to a larger model. 
AutoMix~\cite{madaan2023automix} employs verification prompts to query the model multiple times, using the consistency of these responses as an estimated confidence score.    
The framework then determines whether the current model's output should be accepted or if the query should be forwarded to other models for enhanced performance.

\paragraph{\textbf{Model Routing}} This process optimizes the deployment of multiple models of varying sizes by dynamically directing input data to the most appropriate models, thereby enhancing both efficiency and effectiveness in practical applications. The core component of this approach is the development of a router that assigns input to one or more suitable models within the pool. Existing approaches can be divided into two main categories: (1) Training-free Router and (2) Training-required Router.

\textbf{(1) Training-free Router.} These methods do not require a specific supervised training step for a routing model. Instead, they rely on heuristics such as similarity ranking, uncertainty or other lightweight signals. For example, the router can be guided by estimated uncertainty of predictions, which allows the system to dynamically select between larger and smaller models~\cite{Su2025CPRouter}. Or we can use the agreement (or similarity) among outputs from different models to select the best model for each input. The intuition is that if a model’s output is very different compared to the other models' outputs for the same input, then that model's output is more likely to be of lower quality and unreliable~\cite{guha2024smoothie}. 
\colortext{We can also use semantic tags, i.e., high-level semantic descriptions of a query (e.g., task type, domain, or intent), to route requests to appropriate models~\cite{Chen2025TagRouter}. These tags are predicted by a lightweight tagger and enable effective routing for open-domain text generation tasks.}
\colortext{In this setting, the router avoids the cost of data annotation and supervised router training, which makes it easy to adapt to new domains. However, the inference efficiency depends on the routing strategy. While routers based on lightweight signals such as uncertainty incur only a small overhead, methods requiring multiple model outputs may increase inference cost.}

\textbf{(2) Training-required Router.} These methods train a dedicated routing model, such as a classifier or ranking model, to learn how to allocate user tasks to the appropriate size models. 
A straightforward approach is to consider the input-output pairs from all models and select the best-performing one~\cite{jiang2023llm,ding2024hybrid}. However, this comprehensive ensemble strategy does not significantly reduce inference costs. To address this, some methods train efficient, reward-based routers that select optimal models without needing to access the models' outputs~\cite{lu2023routing,narayanan2023tryage}.
OrchestraLLM~\cite{lee2024orchestrallm} introduces a retrieval-based dynamic router that assumes instances with similar semantic embeddings share the same difficulty level. This allows for the selection of an appropriate expert based on the embedding distances between the testing instance and those in expert pools. Similarly, 
\colortext{RouteLLM~\cite{ong2024routellm} formulates routing as a preference prediction problem by training a lightweight router on human preference data and augmented examples to predict whether a query requires a stronger LLM or can be handled by a weaker one. Based on the predicted user preference, the router dynamically selects the appropriate model, which can reduce inference cost while maintaining response quality and generalizing well to out-of-domain queries}.
FORC~\cite{vsakota2024fly} proposes a meta-model (a regression model) to assign queries to the most suitable model without requiring the execution of any large models during the process. The meta-model is trained on existing pairs of queries and model performance scores. 
In this category, the router has to be maintained, but can often yield stronger selection performance.

Furthermore, recent benchmarks for model routing have been established~\cite{hu2024routerbench,shnitzer2023large}, facilitating the collaboration between larger and smaller models.

\paragraph{\textbf{Speculative Decoding}} This technique aims to speed up the decoding process of a generative model, which often involves using a smaller, faster auxiliary model alongside the main, larger model, which consists of a two-phase ``draft-then-verify'' pipeline. 
\colortext{A smaller, faster draft model first generates multiple candidate tokens, which are then verified by the larger target model in a single parallel forward pass rather than being generated one by one~\cite{leviathan2023fast,chen2023accelerating,xia2024unlocking}. Consequently, when most drafted tokens are accepted, the number of sequential decoding steps required by the target model is substantially reduced, which leads to lower inference latency while preserving the original output distribution.}

\medskip
\noindent\textbf{Summary}

From a practitioner's perspective, the choice among model cascading, routing, and speculative decoding depends on deployment goals. 
\textbf{Model cascading} is most suitable when query difficulty varies widely, and cost reduction is the primary objective. 
A small model handles easy cases, while harder queries are deferred to larger models. 
This step can ensure robustness with minimal engineering efforts. 
\textbf{Model routing} is preferable when the task involves multiple types (e.g., coding and summarization) or domains (e.g., law and healthcare), as a routing mechanism can directly assign each query to the most appropriate model in a single pass. 
\colortext{In contrast, speculative decoding is particularly suitable when a large model is required for high-quality generation. A smaller model drafts multiple candidate tokens that are verified by the large model in parallel, which is able to reduce sequential decoding steps and inference latency without sacrificing output quality.}


\medskip
\noindent\textbf{Future Directions}

(1) Existing model ensembling methods typically rely on a limited, pre-defined list of models, yet the real world encompasses open-domain and constantly evolving LLMs, such as those available on HuggingFace. Exploring ways to leverage these extensive model libraries to create intelligent and efficient systems holds significant promise~\cite{shen2024hugginggpt}.
\\[2ex]
(2) In some existing approaches to speculative decoding, the draft model is required to share the vocabulary with the main model, which typically limits speculative decoding to models from the same family (for example, different sizes of GPT). However, it would be beneficial to explore collaborations between models from diverse sources and vocabularies, thereby expanding the set of viable auxiliary models and enabling more flexible, cost-efficient inference~\cite{Remy2024TransTokenization}.


\begin{figure*}
    \scriptsize
    \centering
\tikzset{
    basic/.style  = {draw, text width=.6cm, align=center, font=\sffamily, rectangle},
    root/.style   = {basic, rounded corners=2pt, thin, align=center, fill=green!30},
    wnode/.style = {basic, rounded corners, thin, align=center, draw=gray!35, fill=gray!5, line width=1pt, text width=7.5em},
    w1node/.style = {basic, rounded corners, thin, align=center, fill=gray!5, text width=7.5em, draw=gray!35, line width=1pt},
    w2node/.style = {basic, rounded corners, thin, align=center, fill=gray!5, text width=15.5em, draw=gray!35, line width=1pt},
    edge from parent/.style={draw=black, edge from parent fork right},
}

\begin{forest} for tree={
    grow=east,
    growth parent anchor=west,
    parent anchor=east,
    child anchor=west,
    edge={gray, thick},
    l sep=20pt, 
    s sep=12pt, 
    edge path={%
        \noexpand\path [\forestoption{edge}] 
          (!u.parent anchor) -- (.child anchor)
          \forestoption{edge label};
    },
}
    [Evaluating LLMs \\ (\textsection~\ref{sec:eval_llms}), wnode,  l sep=10mm,
        [Reference-free Evaluation \\ , wnode
            [\tiny Small Models as Judges, w2node],
            [\tiny Small Proxy Evaluators, w2node]
        ], 
        [Reference-based Evaluation \\ , wnode
            [\tiny Computing semantic similarity between candidate and reference texts, w2node]
        ]] 
\end{forest}
    \caption{Taxonomy of evaluating LLMs}
    \label{fig:eval_llms}
\end{figure*}

\subsubsection{Evaluating LLMs}~\label{sec:eval_llms}
Evaluating open-ended text generation remains one of the significant challenges in deploying LLMs~\cite{chang2024survey}. Unlike classification tasks with clear ground-truth labels, generative outputs can be valid in many different forms, making quality assessment inherently subjective and challenging. 
Traditional metrics such as BLEU~\cite{papineni2002bleu} and ROUGE~\cite{lin2004rouge} rely on surface-level overlap with reference texts, but lexical similarity alone often fails to capture semantic relatedness and reasoning correctness~\cite{liu2016not}. 
As models become more capable and outputs more diverse, these limitations become increasingly pronounced.
\\
This motivates a shift from string matching to model-based evaluation, where smaller models are used as automated judges to approximate human assessment. Rather than measuring n-gram overlap, these evaluators aim to assess semantic similarity, coherence, factual consistency, or overall quality. Broadly, such approaches fall into two categories: (1) \emph{reference-based methods}, which compare outputs against ground-truth texts using learned semantic metrics, and (2) \emph{reference-free methods}, which directly judge quality without requiring explicit references, as illustrated in Figure~\ref{fig:eval_llms}. These paradigms reflect a broader transition toward scalable evaluation frameworks for open-ended generation.


\textbf{(1) Reference-based Evaluation} approaches rely on comparing outputs to human-written references.
For instance, BERTSCORE~\cite{zhang2019bertscore} uses a BERT encoder to compute semantic similarity between candidate and reference texts, and BARTScore~\cite{yuan2021bartscore} leverages the BART encoder-decoder model to evaluate aspects like informativeness, fluency, and factuality.
Although useful when references are available, these methods may struggle in highly open-ended tasks due to limited reference coverage.

\textbf{(2) Reference-free Evaluation} approaches evaluate model outputs without relying on human-written references.
This includes using a small proxy evaluator to estimate how well the generation would perform, such as fluency, relevance, coherence, and factuality.
For example, we can apply small natural language inference (NLI) models to estimate the uncertainty of LLM responses~\cite{manakul2023selfcheckgpt,kuhnsemantic2023}, or employ proxy models to predict LLM performance~\cite{anugraha2024proxylm}, which substantially reduces the computational costs associated with fine-tuning and inference during model selection.
Recent advances include frameworks that treat an LLM as a judge~\cite{gu2024survey}, which introduce another judge LM to evaluate the quality of generated texts. 
Existing findings indicate that smaller judges can provide reasonable signals for ranking tasks even if their absolute alignment was weaker~\cite{thakur2024judging}.

\medskip
\noindent\textbf{Summary}

From a deployment perspective, the choice between reference-based and reference-free evaluation depends mainly on the availability of high-quality references. 
Reference-based methods are most suitable when reliable human-written references exist, and outputs are relatively constrained (e.g., translation, summarization). In such settings, semantic metrics like BERTScore or BARTScore provide stable and reproducible comparisons at low cost. 
However, when tasks are highly open-ended (e.g., creative writing and reasoning) or reference coverage is sparse, reference-based metrics may not be applicable. 
In these cases, reference-free methods become more appropriate, as they directly assess fluency, coherence, factual consistency, or reasoning quality without relying on gold texts. 
Small proxy models or NLI-based evaluators are particularly useful when scalability and cost are priorities. 
Meanwhile, using an LLM as a judge is preferable when high-fidelity evaluation is required, and computational cost is less restrictive, for example in reasoning-required tasks.


\medskip
\noindent\textbf{Future Directions}

(1) As large models advance, they increasingly generate lengthy and complex texts, such as specialized code and scientific papers, which are challenging for humans to evaluate. Consequently, it is essential to develop efficient evaluators to assess various aspects of the generated content, such as factuality~\cite{min2023factscore}, safety~\cite{zhang2024safetybench}, and uncertainty~\cite{huang2023look}.
\\[2ex]
(2) Small evaluators may not be robust, and may be sensitive to the order of answers in pairwise comparisons~\cite{li2024llms}. This judge can also introduce bias that undermines the reliability and trustworthiness of the evaluation~\cite{ye2024justice}. There remains significant room for using small models as judges.



\begin{figure*}
    \scriptsize
    \centering
\tikzset{
    basic/.style  = {draw, text width=.6cm, align=center, font=\sffamily, rectangle},
    root/.style   = {basic, rounded corners=2pt, thin, align=center, fill=green!30},
    wnode/.style = {basic, rounded corners, thin, align=center, draw=gray!35, fill=gray!5, line width=1pt, text width=7.5em},
    w1node/.style = {basic, rounded corners, thin, align=center, fill=gray!5, text width=7.5em, draw=gray!35, line width=1pt},
    w2node/.style = {basic, rounded corners, thin, align=center, fill=gray!5, text width=15.5em, draw=gray!35, line width=1pt},
    edge from parent/.style={draw=black, edge from parent fork right},
}

\begin{forest} for tree={
    grow=east,
    growth parent anchor=west,
    parent anchor=east,
    child anchor=west,
    edge={gray, thick},
    l sep=20pt, 
    s sep=12pt, 
    edge path={%
        \noexpand\path [\forestoption{edge}] 
          (!u.parent anchor) -- (.child anchor)
          \forestoption{edge label};
    },
}
    [Knowledge Distillation, wnode,  l sep=10mm,
        [Representation Distillation \\ , wnode
            [\tiny Feature-based Methods, w2node],
            [\tiny Logit-based Methods, w2node]
        ], 
        [Data Augmentation \\ , wnode
            [\tiny Increasing Data Diversity, w2node],
            [\tiny Generating Pseudo Labels, w2node]
        ], 
        [Rationale Distillation \\ , wnode
            [\tiny Structured Reasoning Distillation, w2node],
            [\tiny Chain-of-Thought Distillation, w2node]
        ], 
        [Full Data Synthesis, wnode
            [\tiny Alignment Data Generation, w2node],
            [\tiny Training Data Generation, w2node]
        ]] 
\end{forest}
    \caption{Taxonomy of knowledge distillation}
    \label{fig:knowledge_distillation}
\end{figure*}


\subsection{LLMs Enhance Small Models}~\label{sec:LLM_enhance_SM}

LLMs can enhance small models mainly through \emph{Knowledge Distillation}~\cite{hinton2015distilling, gou2021knowledge, zhu2023survey, xu2024survey},
where knowledge learned by a large, powerful teacher model is transferred to a smaller, more efficient student model. 
Rather than directly increasing model size or training data, this paradigm uses LLMs to provide richer supervision and guidance during training, which enables small models to achieve stronger performance under limited capacity and computational budgets. 
In practice, knowledge distillation from LLMs can be performed at multiple levels of granularity.
First, LLMs can synthesize full training data, including both inputs and labels (Section~\ref{sec:train_data_syn}). 
Second, they can provide explicit rationales, i.e., the reasoning steps, that guide small models toward more structured and interpretable reasoning (Section~\ref{sec:ration_distill}). 
Third, LLMs can support data augmentation by rephrasing and diversifying existing datasets (Section~\ref{sec:data_aug}). 
Finally, distillation can operate on the teacher model's internal representations, where internal features or intermediate states are used to shape the learned representations of small models (Section~\ref{sec:repre_distill}). 
The first three approaches correspond to black-box distillation, as they depend only on observable outputs of the teacher model, whereas representation distillation is a white-box method that requires access to internal model states.
Together, these approaches transfer the knowledge from LLMs to small models, improving their reasoning ability without increasing model size.

\subsubsection{Full Data Synthesis}\label{sec:train_data_syn}

Human-created data is expensive and finite, and there is a concern that publicly available human text may soon be depleted~\cite{villalobosposition}.
In response, using LLMs to generate full training data to learn a task-specific small model is both efficient and practical.

\textbf{(1) Training Data Generation.}
The idea of this category emphasizes generating both \emph{inputs} and \emph{labels} from scratch~\cite{naduaș2025synthetic,chung2023increasing}.
For example, ~\citet{ye2022zerogen} propose generating a synthetic dataset from an LLM in an unsupervised zero-shot fashion, and then training a much smaller downstream model on this dataset to achieve efficient inference.
~\citet{meng2022generating} use a larger model to generate class-conditioned synthetic texts for zero-shot language understanding tasks.
Subsequent studies have extended this method to various tasks, including text classification~\cite{li2023synthetic}, clinical text mining~\cite{tang2023does}, information extraction~\cite{josifoski2023exploiting}, and hate speech detection~\cite{hartvigsen2022toxigen}.

\textbf{(2) Alignment Data Generation.}
Beyond task-oriented synthetic generation, another line of recent work focuses on alignment data generation, where LLMs synthesize instruction–response pairs to fine-tune the target model, aiming to align it with human-preferred properties.
Examples include CodecLM, which adaptively generates high-quality synthetic instruction data tailored to specific instruction distributions~\cite{wang2024codeclm}, and Bonito, which converts unannotated text into instruction tuning datasets to improve zero-shot adaptation of LLMs~\cite{nayak2024learning}. 
Rather than merely scaling up instruction datasets, these alignment synthesis methods highlight the importance of data quality and distributional matching.

\subsubsection{Rationale Distillation}\label{sec:ration_distill}
This distillation focuses on transferring reasoning rationales from teacher to student, going beyond simple final-answer supervision. 

\textbf{(1) Chain-of-Thought (CoT) Distillation.}
This approach uses explicit step-by-step reasoning steps generated by large language models to provide richer supervision for smaller models. 
In CoT distillation, the student model is trained not only on task inputs and outputs but also on multi-step textual rationales from the teacher, which has been shown to enhance reasoning capabilities on complex tasks~\cite{li2022explanations, hsieh2023distilling, shridhar2023distilling, magister2023teaching, li2023symbolic, fu2023specializing, tian2024tinyllm}. 
Recent work has also explored alternative textual reasoning transfer. 
For example, adversarial distillation frameworks such as Lion~\cite{jiang2023lion} iteratively generate and focus on hard instructions to improve student performance via feedback from the teacher model. 
Counterfactual distillation methods leverage counterfactual examples to make SMs more robust to out-of-distribution (OOD) data~\cite{feng2024teaching}.

\textbf{(2) Structured Reasoning Distillation.} Apart from natural-language CoT distillation, there is a growing line of work that moves beyond surface textual chains to distill reasoning in more structured forms. We refer to this as \emph{Structured Reasoning Distillation}, where reasoning knowledge is captured in structured signals and discrete logic. For instance, Program-aided Distillation~\cite{zhu2024pad} leverages synthetic reasoning programs generated by LLMs for fine-tuning smaller models, which shows promising performance on arithmetic reasoning, symbolic reasoning, and general ability.
Reasoning Scaffolding~\cite{wen2025reasoning} distills the structured, algorithmic flow of a teacher's reasoning, moving beyond surface-level text imitation, which leads to more accurate and logically robust reasoning than standard CoT distillation.

\subsubsection{Data Augmentation}\label{sec:data_aug}
In this context, data augmentation refers to the use of LLMs to modify existing data points to increase data diversity, which can then be directly used to train smaller models~\cite{ding2024data,chen2023empirical}. The use of LLMs for dataset augmentation has become a powerful strategy in NLP, offering a way to enrich training samples when labeled data is scarce.

\textbf{(1) Generating Pseudo Labels.}
One common form of augmentation uses LLMs to assign labels to existing unlabeled or weakly labeled inputs, which can produce pseudo-labeled data that supplements supervised learning~\cite{wang2021want,gao2022self}. This parallels knowledge-distillation via data labeling, where teacher models bootstrap learning by providing labels for otherwise unlabeled inputs.

\textbf{(2) Increasing Data Diversity.}
Another major category involves transforming or rewriting existing data to create variations that preserve semantic content while improving dataset diversity. 
LLMs can paraphrase or rewrite texts to generate additional training samples, which has been shown to improve model robustness and generalization~\cite{mi2022improving,witteveen2019paraphrasing}. This technique has been applied in tasks such as information retrieval, where LLMs rewrite queries to better match target documents~\cite{ma2023query}. More broadly, LLM-based rewriting has been used to augment datasets for text classification~\cite{ding2024data}. Furthermore, data augmentation can be applied to various tasks such as personality detection~\cite{hu2024llm}, intent classification~\cite{sahu2022data}, and dialogue understanding~\cite{chen2022weakly}. Fine-tuning smaller models with these augmented samples can significantly enhance their efficacy and robustness.

\subsubsection{Representation Distillation}\label{sec:repre_distill}
This approach involves using internal states of the teacher model, which provides transparency in the training process of the student model. This approach leverages the teacher's output distributions and internal features such as hidden representations and attention maps to guide the student's learning process. 

\textbf{(1) Logit-based Methods.} The student imitates the teacher's final output distributions, e.g., soft labels or logits. This is the most direct form of knowledge transfer. For example, \citet{hinton2015distilling} propose using the teacher's softened softmax outputs as targets for the student.
More recently, \citet{gu2024minillm} in MiniLLM adopt the reverse KL divergence between student and teacher distributions as the training objective. Similarly, \citet{yang2023knowledge} propose a unified approach to both teacher-student and self-distillation by using normalized loss functions and customized soft labels, thereby refining how soft targets are used in distillation.

\textbf{(2) Feature-based Methods.} Beyond just outputs, the student learns to imitate the internal representations of the teacher model, such as hidden states, attention maps, or layer activations.
This process provides richer information about the teacher model's internal representations than using its output predictions alone.
For example, DistilBERT~\cite{sanh2019distilbert} and ZeroQuant~\cite{yao2022zeroquant} include a distance loss between the hidden states of teacher and student in the training objective.
~\citet{liang2023less} propose a task-aware layer-wise distillation method, where the student aligns with the teacher's hidden representations through a task-specific filtering to select only the informative knowledge, thereby reducing redundancy.
\colortext{Additionally, ~\citet{liu2023llm} use synthetic generations from the teacher as pseudo-data to perform feature-based distillation for a fully quantized student model. This is achieved by aligning the teacher's intermediate representations, including hidden activations and KV-cache, thereby preserving the student's internal representations after quantization.}

\medskip
\noindent\textbf{Summary}

Knowledge distillation facilitates the transfer of knowledge from a larger model to a smaller one, which enables the development of more cost-effective and efficient models. 
In practice, the choice of distillation strategy depends on whether the goal is data expansion, reasoning transfer, or model compression. 
Full data synthesis is useful when human-labeled data is scarce, and scaling supervision is the priority. 
Rationale distillation emphasizes transferring reasoning capabilities from a large model to a smaller one, such as improving multi-step reasoning or explainable decision-making. 
Data augmentation methods aim to improve robustness and diversity when a baseline dataset already exists but lacks coverage or diversity. 
Finally, representation distillation is ideal when the goal is model compression and the internal state of the teacher model is accessible.

\medskip
\noindent\textbf{Future Directions}

(1) Currently, closed-source LLMs remain more powerful than their open-source counterparts. However, using closed-source models for data synthesis may raise privacy and security concerns, particularly in sensitive contexts such as medical scenarios involving patient data~\cite{ollion2023chatgpt}. Addressing how to protect data privacy during this process is a critical area of concern.
\\[2ex]
(2) Generating training data with large-scale models is costly, making it essential to explore methods for reducing the expense while still producing high-quality data. For instance, recent research suggests that smaller, less powerful models can sometimes generate better training data points~\cite{bansal2024smaller}.
\\[2ex]
\colortext{(3) Although data distillation and synthetic data generation have shown promising results, they may also introduce noisy, hallucinated content, and distribution bias~\cite{liu2024best,chen-etal-2024-unveiling-flaws}. Developing effective data quality assessment, filtering, and verification methods to identify reliable synthetic data before model training remains an important research direction~\cite{long2024llms}.}


\begin{figure*}
    \scriptsize
    \centering
\tikzset{
    basic/.style  = {draw, text width=2.6cm, align=center, font=\sffamily, rectangle},
    root/.style   = {basic, rounded corners=2pt, thin, align=center, fill=green!30},
    wnode/.style = {basic, rounded corners, thin, align=center, draw=yellow!50, line width=1pt, fill=yellow!10, text width=10.5em},
    w1node/.style = {basic, rounded corners, thin, align=center, fill=yellow!10, text width=25.5em, draw=yellow!50, line width=1pt},
    znode/.style = {basic, rounded corners, thin, align=center, fill=green!15, text width=10.5em, draw=green!50, line width=1pt},
    z1node/.style = {basic, rounded corners,thin, align=center, fill=green!10, text width=25.5em, draw=green!50, line width=1pt},
    edge from parent/.style={draw=black, edge from parent fork right},
    xnode/.style = {basic, rounded corners, thin, align=center, fill=red!15, text width=10.5em, draw=red!50, line width=1pt},
    x1node/.style = {basic, rounded corners,thin, align=center, fill=red!15, text width=25.5em, draw=red!15, line width=1pt},
}

\begin{forest} for tree={
    grow=east,
    growth parent anchor=west,
    parent anchor=east,
    child anchor=west,
    edge={gray, thick},
    l sep=20pt, 
    s sep=12pt, 
    edge path={%
        \noexpand\path [\forestoption{edge}] 
          (!u.parent anchor) -- (.child anchor)
          \forestoption{edge label};
    },
}
[{Competition and Complementarity}\\ (\textsection~\ref{sec:competition}), basic,  l sep=5mm,
    [Interpretability-required Environment \\ (\textsection~\ref{sec:intepretability_env}), xnode,  l sep=10mm,
        [healthcare; finance;
        law, x1node
        ]] 
    [Task-specific Environment \\ (\textsection~\ref{sec:task_env}), wnode,  l sep=10mm,
        [Other Specialized Tasks: machine-generated text detection; spreadsheet representation; information extraction, w1node
        ]
        [Short Text Tasks, w1node
        ],
        [Tabular Reasoning, w1node
        ],
        [Low-Resource Languages, w1node
        ],
        [Domain-Specific Tasks: biomedical texts and legal texts, w1node
        ]] 
    [Computation-constrained Environment \\ (\textsection~\ref{sec:computation_env}), znode, l sep=10mm,
        [Low-Latency Computing\\ , z1node
        ]
        [Edge Computing , z1node 
        ]
    ]
]
\end{forest}
    \caption{Competition and complementarity between SMs and LLMs}
    \label{fig:competition}
\end{figure*}

\section{Competition and Complementarity}~\label{sec:competition}

In this section, we examine the relationship between small models and large models by emphasizing the competition and complementarity. Small models offer distinct advantages in cost-efficiency and interpretability (small models or shallow networks tend to be more interpretable than their large counterparts)~\cite{gilpin2018explaining, barcelo2020model}.
At the same time, large models can boost small ones through knowledge distillation and data-generation~\cite{gou2021knowledge,naduaș2025synthetic}.
We argue that rather than replacing one with the other, the optimal ecosystem is hybrid.
Small models serving specialized or cost-effective roles, with large models built to support, guide, and augment them. 
Below, we outline three scenarios—computation-constrained environments (Section \ref{sec:computation_env}), task-specific environments (Section \ref{sec:task_env}), and interpretability-critical environments (Section \ref{sec:intepretability_env}), as shown in Figure~\ref{fig:competition}. 
Small models are not in opposition to large models but form an indispensable and complementary part of the particular deployment settings.

It is important to note that when we refer to small models in this section, we are not limiting the concept to Transformer-only architectures. 
The notion of small models may naturally extend to other architectures, such as shallow neural networks or even statistical models. 
This inclusive definition enables our discussion to remain relevant even as architectures evolve and model sizes continue to rise in the future. 
If tomorrow a new class of model emerges, what we classify as ``small'' and ``large'' will simply shift in relative terms, but the principles we adopt in this work remain applicable.

\subsection{Computation-constrained Environment}~\label{sec:computation_env}

As LLMs become more capable, their deployments come with substantial computational demands, which is impractical in many operational settings, e.g., edge computing~\cite{wanefficient2023,dhar2024empirical} and low-latency computing. By contrast, smaller models offer an alternative in environments where latency or hardware are constrained. In this section, we focus on two key categories: (1) Edge Computing and (2) Low-Latency Computing. 

\paragraph{\textbf{Edge Computing}}
In edge devices such as mobile phones and IoT devices, memory, power and connectivity are severely limited~\cite{dhar2024empirical}. Large models are often infeasible for these resource-constrained conditions. 
Existing surveys have evaluated small language models, such as Phi-3-mini (3.8B)~\cite{abdin2024phi}, MiniCPM~\cite{hu2024minicpm}, and Gemma-2B~\cite{team2024gemma}, on performance and running time for edge settings~\cite{lu2025demystifying,jang2025edge}, which offers a comprehensive view of SMs across hardware and systems.
These studies advocate that smaller models are viable approaches for resource-constrained deployment scenarios like smartphones and Web-of-Things devices.

\paragraph{\textbf{Low-Latency Computing}}
In low-latency inference scenarios, such as search engines, document retrieval, or recommendation systems, the ability to process many requests quickly and return responses with minimal delay is very important. 
Many tasks, however,  are not knowledge-intensive and do not demand complex reasoning that can be effectively handled by smaller models.

\colortext{To provide an empirical comparison between model scale and task performance, we evaluate a collection of representative text embedding models, including both encoder-only and decoder-only architectures, with different parameter sizes on the Massive Text Embedding Benchmark (MTEB)~\cite{muennighoff2023mteb}.  We select four representative tasks, namely text similarity, text classification, information retrieval, and text clustering, and report the average performance over five datasets for each task. MTEB was chosen because it is a widely adopted benchmark covering diverse text understanding and retrieval tasks, which makes it suitable for analyzing how model scaling affects performance under practical deployment scenarios.
}

\begin{figure*}[t]%
\centering
\subfloat[\centering Cosine Similarity on \texttt{Text Similarity}]{{\includegraphics[width=0.24\textwidth]{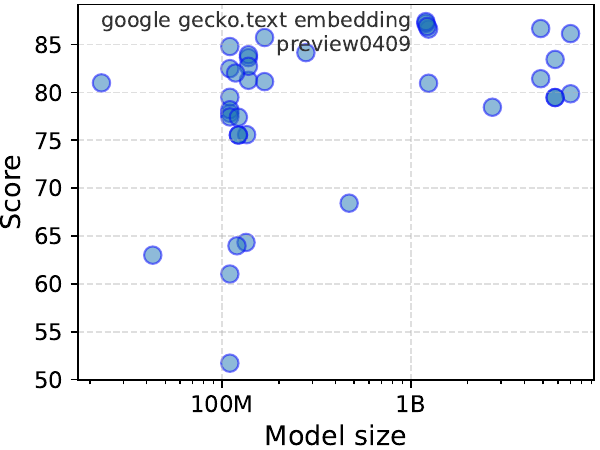}}}
\subfloat[\centering  Accuracy on \texttt{Text Classification}]{{\includegraphics[width=0.24\textwidth]{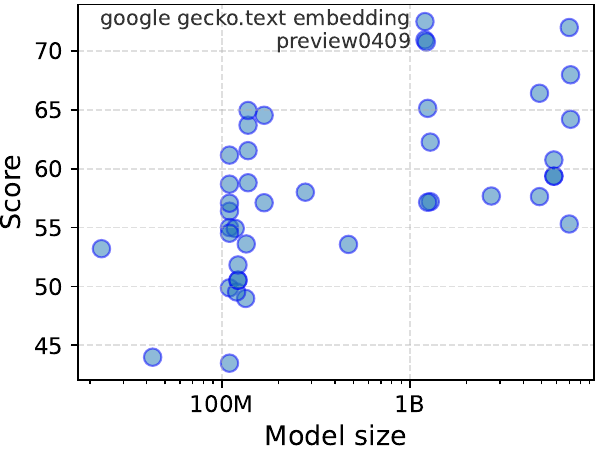} }}%
\subfloat[\centering NDCG@10 on \texttt{Information Retrieval}]{{\includegraphics[width=0.24\textwidth]{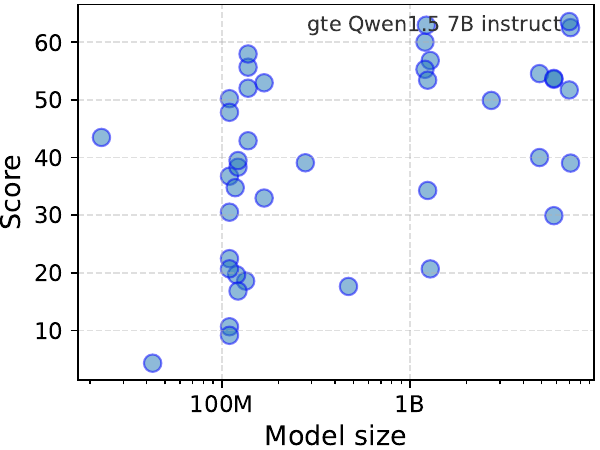}\label{fig:ir_results} }}
\subfloat[\centering v\_measure on \texttt{Text Clustering}]{{\includegraphics[width=0.24\textwidth]{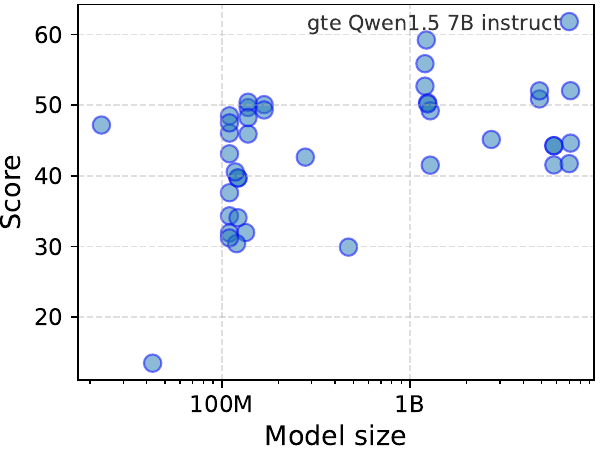} }}%

\caption{
\colortext{Performance of representative text embedding models, including both encoder-only and decoder-only architectures, on four MTEB task categories. Results are averaged over five datasets for each task.}}
	\label{fig:mteb_results}%
\end{figure*}

For instance, Figure~\ref{fig:mteb_results} illustrates the relationship between performance and model size across four tasks in MTEB~\cite{muennighoff2023mteb}, where we observe diminishing returns from increasing model sizes, particularly in tasks like text similarity and classification.
In the case of information retrieval (Figure~\ref{fig:ir_results}), which involves computing similarity between a query and a document collection, faster inference speed is critical. Under these conditions, the lightweight encoder-based embedding models
remain widely used in the IR task~\cite{reimers2019sentence,samarinas2025distillation,yu2025integrating}.

Small models are increasingly valuable in scenarios where computational resources are limited. Techniques such as knowledge distillation~\cite{xu2024survey} allow the transfer of knowledge from LLMs to smaller models, enabling these smaller models to achieve similar performance while significantly reducing model size. 
Consequently, smaller models are strategic components for edge deployment and low-latency computing.

\subsection{Task-specific Environment}~\label{sec:task_env}

Training LLMs requires trillions of tokens~\cite{raffel2020exploring, kaplan2020scaling, gao2020pile}, but sufficient data is often unavailable for certain specialized domains (e.g., biomedical text) or tasks (e.g., tabular reasoning). In such cases, pre-training a large foundational model is not feasible, and small models can offer promising returns in this case.
We outline several task-specific scenarios where small models can deliver comparable results.

\paragraph{\textbf{Domain-Specific Tasks}}
Domains such as biomedical or legal fields often have fewer training tokens available. Recent studies have shown that fine-tuning small models on domain-specific datasets can outperform general LLMs on various biomedical~\cite{hernandez2023we, juan2024fine} and legal~\cite{chalkidis2023chatgpt} tasks.

\paragraph{\textbf{Low-Resource Languages}}

Low-resource languages often lack sufficient annotated data to effectively train large, powerful models. In such settings, smaller multilingual models, such as mBERT~\cite{devlin2019bert}, offer a more practical and promising alternative~\cite{gurgurov2025small}.
In low-resource language translation, specialized small models are commonly used to synthesize training data by generating large-scale parallel corpora from monolingual text~\cite{sennrich2016improving}. This approach enables the creation of sufficient pretraining data for low-resource languages,  while small encoder–decoder models can outperform general-purpose LLMs in these settings~\cite{doshi2024pretraining,wang2025multilingual}.

\paragraph{\textbf{Tabular Reasoning}}
Tabular datasets are typically smaller than benchmarks in other domains, such as text or image data, and are highly structured, consisting of heterogeneous data types (e.g., numerical, categorical, ordinal). Due to these characteristics, small tree-based models can achieve competitive performance compared to large deep-learning models for tabular data~\cite{grinsztajn2022tree}.

\paragraph{\textbf{Short Text Tasks}}
Short text representation and reasoning do not generally require extensive background knowledge. As a result, small models are particularly effective for tasks such as text classification~\cite{zhang2023sentiment}, phrase representation~\cite{chen2024learning}, and entity retrieval~\cite{chen2021lightweight}.

\paragraph{\textbf{Other Specialized Tasks}}
In certain niche areas, smaller models can surpass larger ones. Examples include machine-generated text detection~\cite{mireshghallah2023smaller}, spreadsheet representation~\cite{joshi2024flame}, and information extraction~\cite{ma2023large}.

Across these settings, the advantages of small models arise from limited data availability and well-defined task structures. In domain-specific tasks and low-resource languages, training data is limited, which makes smaller models more data-efficient, easier to adapt, and often more practical to deploy. For tasks with distinctive and localized patterns, such as short-text understanding, the narrow semantic scope reduces the need for large contextual understanding. As a result, small models tend to show more stable behavior in task-constrained settings, 
and these advantages arise not from model size alone, but from a better alignment between task complexity and available data.

\subsection{Interpretability-required Environment}~\label{sec:intepretability_env}

The goal of interpretability is to provide a human-understandable explanation of a model’s internal reasoning process~\cite{lipton2018mythos, gilpin2018explaining}, i.e., how the model works (\emph{transparency}). Generally, smaller (e.g. shallow) and simpler (e.g. tree-based) models offer better interpretability compared to larger (e.g. deep), more complex models (e.g. neural)~\cite{barcelo2020model, gosiewska2021simpler}.

In practice, industries such as healthcare~\cite{caruana2015intelligible}, finance~\cite{kurshan2021current}, and law~\cite{eliot2021need} often favor smaller, more interpretable models because the decisions produced by these models must be understandable to non-experts (e.g., doctors, financial analysts). In high-stakes decision-making contexts, models that can be easily audited and explained are typically preferred.

When selecting LLMs or SMs to use, it is important to make trade-offs for balancing model complexity with the need for human understanding.

\medskip
\noindent\textbf{Summary}

Smaller models emerge as a viable alternative, which can achieve competitive performances in resource-limited scenarios. 
In domains such as healthcare and law, where transparency and interpretability are essential, practitioners often favour compact models because they are more auditable and easier to understand.
At the same time, techniques such as knowledge distillation further enable small models to obtain capability from large models.
In sum, small models represent not only a cost-sensitive substitute but also a strategic complement to large models.

\medskip
\noindent\textbf{Future Directions}

(1) According to recent industry research, by 2028 roughly 68\% of customer-service interactions are expected to be processed by \emph{agentic systems}~\cite{cisco2025agentic}.
Smaller models are particularly well-suited for agentic systems that perform many small, specialized, and repetitive tasks. It is therefore important to explore how these compact models can serve as the economical components of agentic AI~\cite{belcak2025small}.
\\[2ex]
(2) For some specific tasks such as domain-specific tasks and tabular reasoning, the performance of dedicated small models can even outperform general-purpose LLMs.   
How to organically combine this level of specialization with generality, and enable AI systems to expand their knowledge boundaries and handle a broader range of tasks, is a research topic worth studying.


\section{\colortext{Open Challenges for Small Models}}

\colortext{\paragraph{Reliable Small Models}
Despite their advantages in efficiency, small models continue to face several fundamental limitations. A major limitation is the capability gap compared with LLMs, particularly on knowledge reasoning tasks. Scaling laws suggest that performance generally improves with increasing model size~\cite{kaplan2020scaling}, while several advanced capabilities emerge only beyond certain scales~\cite{wei2022emergent}. Moreover, small models tend to be less robust in real-world deployment. They are more susceptible to distribution shifts, perturbations, and catastrophic forgetting during continual adaptation, which leads to degraded performance on unseen domains and evolving tasks~\cite{chen2022imputing,ramasesh2021effect}. Addressing both the capability gap and robustness remains a key challenge for the next generation of small models.
}

\colortext{
\paragraph{Open-world Model Collaboration}
Current inference collaboration paradigms between small and large models, e.g., model routing and speculative decoding, are mainly designed for a fixed pool of homogeneous models, often from the same model family. 
As AI systems evolve toward open-world deployment, future collaboration is expected to involve a much larger ecosystem of heterogeneous models, including open-weight and API-only models with diverse capabilities. 
Designing collaboration mechanisms that scale to such heterogeneous and open environments remains largely unexplored. 
Although RouteLLM~\cite{ong2024routellm} demonstrates strong transferability, existing routing approaches generally assume a predefined set of candidate models, which cannot fully address open-world settings in which the candidate model pool may evolve dynamically.
Future research should investigate how to accurately estimate the knowledge boundary and capability of each model~\cite{Chen2025QueryLevelUncertainty}, so that queries can be dynamically allocated to the most suitable model. Furthermore, existing approaches should explicitly consider collaborations in different architectures and modalities~\cite{shen2024hugginggpt, Remy2024TransTokenization}.
}

\colortext{\paragraph{Trustworthy Small Evaluators}
As AI-generated content becomes increasingly widespread across society, efficiently evaluating its quality and reliability will become crucial. Small models are promising candidates for evaluating factuality, safety, and fairness of AI-generated content. 
However, ensuring these small evaluators are trustworthy remains a major challenge. 
Existing studies show that small judges can be sensitive to answer ordering~\cite{li2024llms} and may exhibit biases~\cite{ye2024justice}, which leads to inconsistent and unreliable evaluation results. 
Future research should therefore focus on developing more robust and fair small evaluators, together with benchmarks and evaluation protocols for measuring their reliability.
}

\colortext{\paragraph{Agentic Small Models}
With the rapid usage of agentic AI across real-world applications~\cite{cisco2025agentic}, small models are expected to become economical building blocks of future intelligent systems~\cite{belcak2025small}. 
A key open challenge is how to design small models as specialized agents that efficiently perform dedicated functions, such as planning, retrieval, tool selection, code generation, document verification, or memory management, while collaborating with other agents to solve complex tasks.
Despite their promising role, agentic small models remain largely underexplored. Developing specialized small agents that can efficiently collaborate with tools and other models is an important direction for future research.}


\section{Conclusion}\label{sec:conclusion}

In this work, we systematically analyze the relationship between LLMs and SMs from both collaborative and competitive perspectives. 
However, the notion of ``small'' is relative. As model sizes continue to grow, potentially toward ever-larger multimodal foundation models, today's large models may become tomorrow's small ones. 
Regarding the future relationship between small models and large models, we expect to observe an evolutionary dynamic. 
Although LLMs outperform smaller models in most tasks, small models persist by occupying ecological niches where large models are maladapted, such as low-latency inference, on-device deployment, and low-resource environments.

More importantly, the dominant trend might shift from competition to cooperation. Small models and LLMs increasingly work together through knowledge distillation, efficient inference techniques, data curation, and evaluation frameworks.
This division and cooperation create a diversified and balanced AI ecosystem in which models of different scales specialize and complement each other.
Combining SMs and LLMs makes systems more scalable across different tasks, more energy-efficient, and better able to adapt to changing needs.

In summary, small models and large models coexist by specializing in different resource niches and increasingly cooperating to build a more efficient and robust AI ecosystem. Our relative definition of model size ensures that this core idea remains relevant as model scales continue to evolve.


\appendix



\bibliographystyle{compling}
\bibliography{COLI_template}

@article{kaplan2020scaling,
    author = {Kaplan, Jared and McCandlish, Sam and Henighan, Tom and Brown, Tom B and Chess, Benjamin and Child, Rewon and Gray, Scott and Radford, Alec and Wu, Jeffrey and Amodei, Dario},
    journal = {arXiv preprint arXiv:2001.08361},
    title = {Scaling laws for neural language models},
    url = {https://arxiv.org/abs/2001.08361},
year = {2020}
}

@article{wei2022emergent,
    author = {Wei, Jason and Tay, Yi and Bommasani, Rishi and Raffel, Colin and Zoph, Barret and Borgeaud, Sebastian and Yogatama, Dani and Bosma, Maarten and Zhou, Denny and Metzler, Donald and others},
    journal = {Transactions on Machine Learning Research},
    title = {Emergent Abilities of Large Language Models},
    year = {2022}
}

@article{
chen2023frugalgpt,
title={Frugal{GPT}: How to Use Large Language Models While Reducing Cost and Improving Performance},
author={Lingjiao Chen and Matei Zaharia and James Zou},
journal={Transactions on Machine Learning Research},
issn={2835-8856},
year={2024},
url={https://openreview.net/forum?id=cSimKw5p6R}
}

@inproceedings{leviathan2023fast,
    author = {Leviathan, Yaniv and Kalman, Matan and Matias, Yossi},
    booktitle = {International Conference on Machine Learning},
    organization = {PMLR},
    title = {Fast inference from transformers via speculative decoding},
    year = {2023}
}

@inproceedings{ma2023large,
    author = {Ma, Yubo and Cao, Yixin and Hong, Yong and Sun, Aixin},
    booktitle = {Findings of the Association for Computational Linguistics: EMNLP 2023},
    title = {Large Language Model Is Not a Good Few-shot Information Extractor, but a Good Reranker for Hard Samples!},
    year = {2023}
}

@inproceedings{juneja2023small,
    author = {Juneja, Gurusha and Dutta, Subhabrata and Chakrabarti, Soumen and Manchanda, Sunny and Chakraborty, Tanmoy},
    booktitle = {Proc. of EMNLP},
    title = {Small Language Models Fine-tuned to Coordinate Larger Language Models improve Complex Reasoning},
    year = {2023}
}

@inproceedings{xu2023small,
  title={Small models are valuable plug-ins for large language models},
  author={Xu, Canwen and Xu, Yichong and Wang, Shuohang and Liu, Yang and Zhu, Chenguang and McAuley, Julian},
  booktitle={Findings of the Association for Computational Linguistics: ACL 2024},
  pages={283--294},
  year={2024}
}

@inproceedings{li2024blade,
  title={Blade: Enhancing black-box large language models with small domain-specific models},
  author={Li, Haitao and Ai, Qingyao and Chen, Jia and Dong, Qian and Wu, Zhijing and Liu, Yiqun},
  booktitle={Proceedings of the AAAI Conference on Artificial Intelligence},
  volume={39},
  number={23},
  pages={24422--24430},
  year={2025}
}

@inproceedings{li2023contrastive,
    author = {Li, Xiang Lisa and Holtzman, Ari and Fried, Daniel and Liang, Percy and Eisner, Jason and Hashimoto, Tatsunori B and Zettlemoyer, Luke and Lewis, Mike},
    booktitle = {Proc. of ACL},
    title = {Contrastive Decoding: Open-ended Text Generation as Optimization},
    year = {2023}
}

@article{gao2023retrieval,
  title={Retrieval-augmented generation for large language models: A survey},
  author={Gao, Yunfan and Xiong, Yun and Gao, Xinyu and Jia, Kangxiang and Pan, Jinliu and Bi, Yuxi and Dai, Yi and Sun, Jiawei and Wang, Meng and Wang, Haofen},
  journal = {arXiv preprint arXiv:2312.10997},
  year={2023},
}

@inproceedings{chen2022imputing,
  title={Imputing Out-of-Vocabulary Embeddings with LOVE Makes LanguageModels Robust with Little Cost},
  author={Chen, Lihu and Varoquaux, Gael and Suchanek, Fabian},
  booktitle={Proceedings of the 60th Annual Meeting of the Association for Computational Linguistics (Volume 1: Long Papers)},
  pages={3488--3504},
  year={2022}
}

@inproceedings{hsieh2023distilling,
    author = {Hsieh, Cheng-Yu and Li, Chun-Liang and YEH, CHIH-KUAN and Nakhost, Hootan and Fujii, Yasuhisa and Ratner, Alex Jason and Krishna, Ranjay and Lee, Chen-Yu and Pfister, Tomas},
    booktitle = {The 61st Annual Meeting Of The Association For Computational Linguistics},
    title = {Distilling Step-by-Step! Outperforming Larger Language Models with Less Training Data and Smaller Model Sizes},
    year = {2023}
}

@inproceedings{shridhar2023distilling,
    author = {Shridhar, Kumar and Stolfo, Alessandro and Sachan, Mrinmaya},
    booktitle = {Findings of the Association for Computational Linguistics: ACL 2023},
    title = {Distilling Reasoning Capabilities into Smaller Language Models},
    year = {2023}
}

@inproceedings{hernandez2023we,
  title={Do we still need clinical language models?},
  author={Lehman, Eric and Hernandez, Evan and Mahajan, Diwakar and Wulff, Jonas and Smith, Micah J and Ziegler, Zachary and Nadler, Daniel and Szolovits, Peter and Johnson, Alistair and Alsentzer, Emily},
  booktitle={Conference on health, inference, and learning},
  pages={578--597},
  year={2023},
  organization={PMLR}
}

@article{chalkidis2023chatgpt,
  author       = {Ilias Chalkidis},
  title        = {ChatGPT may Pass the Bar Exam Soon, but Has a Long Way to Go for the LexGLUE Benchmark},
  journal = {arXiv preprint arXiv:2304.12202},
  year         = {2023},
archivePrefix= {arXiv},
  primaryClass = {cs.CL},
  url          = {https://arxiv.org/abs/2304.12202},
}

@inproceedings{zhang2023sentiment,
  title={Sentiment analysis in the era of large language models: A reality check},
  author={Zhang, Wenxuan and Deng, Yue and Liu, Bing and Pan, Sinno and Bing, Lidong},
  booktitle={Findings of the Association for Computational Linguistics: NAACL 2024},
  pages={3881--3906},
  year={2024}
}

@inproceedings{joshi2024flame,
    author = {Joshi, Harshit and Ebenezer, Abishai and Sanchez, Jos{\'e} Cambronero and Gulwani, Sumit and Kanade, Aditya and Le, Vu and Radi{\v{c}}ek, Ivan and Verbruggen, Gust},
    booktitle = {Proc. of AAAI},
    number = {12},
    title = {Flame: A small language model for spreadsheet formulas},
    year = {2024}
}

@article{mireshghallah2023smaller,
    author = {Mireshghallah, Fatemehsadat and Mattern, Justus and Gao, Sicun and Shokri, Reza and Berg-Kirkpatrick, Taylor},
    journal = {arXiv preprint arXiv:2305.09859},
    title = {Smaller language models are better black-box machine-generated text detectors},
    url = {https://arxiv.org/abs/2305.09859},
year = {2023}
}

@inproceedings{zhoudocprompting,
    author = {Zhou, Shuyan and Alon, Uri and Xu, Frank F and Jiang, Zhengbao and Neubig, Graham},
    booktitle = {The Eleventh International Conference on Learning Representations},
    title = {DocPrompting: Generating Code by Retrieving the Docs},
    year = {2023}
}

@inproceedings{brown2020language,
    author = {Tom B. Brown and
Benjamin Mann and
Nick Ryder and
Melanie Subbiah and
Jared Kaplan and
Prafulla Dhariwal and
Arvind Neelakantan and
Pranav Shyam and
Girish Sastry and
Amanda Askell and
Sandhini Agarwal and
Ariel Herbert{-}Voss and
Gretchen Krueger and
Tom Henighan and
Rewon Child and
Aditya Ramesh and
Daniel M. Ziegler and
Jeffrey Wu and
Clemens Winter and
Christopher Hesse and
Mark Chen and
Eric Sigler and
Mateusz Litwin and
Scott Gray and
Benjamin Chess and
Jack Clark and
Christopher Berner and
Sam McCandlish and
Alec Radford and
Ilya Sutskever and
Dario Amodei},
    booktitle = {Advances in Neural Information Processing Systems 33: Annual Conference
on Neural Information Processing Systems 2020, NeurIPS 2020, December
6-12, 2020, virtual},
    editor = {Hugo Larochelle and
Marc'Aurelio Ranzato and
Raia Hadsell and
Maria{-}Florina Balcan and
Hsuan{-}Tien Lin},
    title = {Language Models are Few-Shot Learners},
    url = {https://proceedings.neurips.cc/paper/2020/hash/1457c0d6bfcb4967418bfb8ac142f64a-Abstract.html},
    year = {2020}
}

@inproceedings{burnsweak,
    author = {Burns, Collin and Izmailov, Pavel and Kirchner, Jan Hendrik and Baker, Bowen and Gao, Leo and Aschenbrenner, Leopold and Chen, Yining and Ecoffet, Adrien and Joglekar, Manas and Leike, Jan and others},
    booktitle = {Forty-first International Conference on Machine Learning},
    title = {Weak-to-Strong Generalization: Eliciting Strong Capabilities With Weak Supervision},
    year = {2024}
}

@article{zhou2024weak,
  title={Weak-to-strong search: Align large language models via searching over small language models},
  author={Zhou, Zhanhui and Liu, Zhixuan and Liu, Jie and Dong, Zhichen and Yang, Chao and Qiao, Yu},
  journal={Advances in Neural Information Processing Systems},
  volume={37},
  pages={4819--4851},
  year={2024}
}

@article{liu2024co,
    author = {Liu, Yuejiang and Alahi, Alexandre},
    journal = {arXiv preprint arXiv:2402.15505},
    title = {Co-supervised learning: Improving weak-to-strong generalization with hierarchical mixture of experts},
    url = {https://arxiv.org/abs/2402.15505},
year = {2024}
}

@article{guo2024vision,
    author = {Guo, Jianyuan and Chen, Hanting and Wang, Chengcheng and Han, Kai and Xu, Chang and Wang, Yunhe},
    journal = {arXiv preprint arXiv:2402.03749},
    title = {Vision superalignment: Weak-to-strong generalization for vision foundation models},
    url = {https://arxiv.org/abs/2402.03749},
year = {2024}
}

@article{chen2023accelerating,
    author = {Chen, Charlie and Borgeaud, Sebastian and Irving, Geoffrey and Lespiau, Jean-Baptiste and Sifre, Laurent and Jumper, John},
    journal = {arXiv preprint arXiv:2302.01318},
    title = {Accelerating large language model decoding with speculative sampling},
    url = {https://arxiv.org/abs/2302.01318},
year = {2023}
}

@inproceedings{chen2024data,
    author = {Chen, Dong and Zhuang, Yueting and Zhang, Shuo and Liu, Jinfeng and Dong, Su and Tang, Siliang},
    booktitle = {Proc. of AAAI},
    number = {10},
    title = {Data Shunt: Collaboration of Small and Large Models for Lower Costs and Better Performance},
    year = {2024}
}

@inproceedings{anugraha2024proxylm,
  title={Proxylm: Predicting language model performance on multilingual tasks via proxy models},
  author={Anugraha, David and Winata, Genta Indra and Li, Chenyue and Irawan, Patrick Amadeus and Lee, En-Shiun Annie},
  booktitle={Findings of the Association for Computational Linguistics: NAACL 2025},
  pages={1981--2011},
  year={2025}
}

@inproceedings{fengknowledge,
    author = {Feng, Shangbin and Shi, Weijia and Bai, Yuyang and Balachandran, Vidhisha and He, Tianxing and Tsvetkov, Yulia},
    booktitle = {The Twelfth International Conference on Learning Representations},
    title = {Knowledge Card: Filling LLMs' Knowledge Gaps with Plug-in Specialized Language Models},
    year = {2024}
}

@inproceedings{ormazabal2023comblm,
  title={CombLM: Adapting black-box language models through small fine-tuned models},
  author={Ormazabal, Aitor and Artetxe, Mikel and Agirre, Eneko},
  booktitle={Proceedings of the 2023 Conference on Empirical Methods in Natural Language Processing},
  pages={2961--2974},
  year={2023}
}

@inproceedings{lee2024can,
    author = {Lee, Jooyoung and Yang, Fan and Tran, Thanh and Hu, Qian and Barut, Emre and Chang, Kai-Wei},
    booktitle = {Proceedings of the 2024 Joint International Conference on Computational Linguistics, Language Resources and Evaluation (LREC-COLING 2024)},
    title = {Can Small Language Models Help Large Language Models Reason Better?: LM-Guided Chain-of-Thought},
    year = {2024}
}

@inproceedings{cheng2023uprise,
    author = {Cheng, Daixuan and Huang, Shaohan and Bi, Junyu and Zhan, Yuefeng and Liu, Jianfeng and Wang, Yujing and Sun, Hao and Wei, Furu and Deng, Weiwei and Zhang, Qi},
    booktitle = {Proc. of EMNLP},
    title = {UPRISE: Universal Prompt Retrieval for Improving Zero-Shot Evaluation},
    year = {2023}
}

@inproceedings{cheng2024small,
  title={Small agent can also rock! empowering small language models as hallucination detector},
  author={Cheng, Xiaoxue and Li, Junyi and Zhao, Wayne Xin and Zhang, Hongzhi and Zhang, Fuzheng and Zhang, Di and Gai, Kun and Wen, Ji-Rong},
  booktitle={Proceedings of the 2024 Conference on Empirical Methods in Natural Language Processing},
  pages={14600--14615},
  year={2024}
}

@inproceedings{magister2023teaching,
    author = {Magister, Lucie Charlotte and Mallinson, Jonathan and Adamek, Jakub Dominik and Malmi, Eric and Severyn, Aliaksei},
    booktitle = {The 61st Annual Meeting Of The Association For Computational Linguistics},
    title = {Teaching Small Language Models to Reason},
    year = {2023}
}

@inproceedings{hu2024llm,
    author = {Hu, Linmei and He, Hongyu and Wang, Duokang and Zhao, Ziwang and Shao, Yingxia and Nie, Liqiang},
    booktitle = {Proc. of AAAI},
    number = {16},
    title = {LLM vs Small Model? Large Language Model Based Text Augmentation Enhanced Personality Detection Model},
    year = {2024}
}

@inproceedings{long2024llms,
  title={On LLMs-driven synthetic data generation, curation, and evaluation: A survey},
  author={Long, Lin and Wang, Rui and Xiao, Ruixuan and Zhao, Junbo and Ding, Xiao and Chen, Gang and Wang, Haobo},
  booktitle={Findings of the Association for Computational Linguistics: ACL 2024},
  pages={11065--11082},
  year={2024}
}

@inproceedings{
ong2024routellm,
title={Route{LLM}: Learning to Route {LLM}s from Preference Data},
author={Isaac Ong and Amjad Almahairi and Vincent Wu and Wei-Lin Chiang and Tianhao Wu and Joseph E. Gonzalez and M Waleed Kadous and Ion Stoica},
booktitle={The Thirteenth International Conference on Learning Representations},
year={2025},
url={https://openreview.net/forum?id=8sSqNntaMr}
}

@inproceedings{lu2023inference,
    author = {Lu, Ximing and Brahman, Faeze and West, Peter and Jung, Jaehun and Chandu, Khyathi and Ravichander, Abhilasha and Ammanabrolu, Prithviraj and Jiang, Liwei and Ramnath, Sahana and Dziri, Nouha and others},
    booktitle = {Proc. of EMNLP},
    title = {Inference-time policy adapters (ipa): Tailoring extreme-scale lms without fine-tuning},
    year = {2023}
}

@inproceedings{sun2024bbox,
  title={BBox-Adapter: Lightweight Adapting for Black-Box Large Language Models},
  author={Sun, Haotian and Zhuang, Yuchen and Wei, Wei and Zhang, Chao and Dai, Bo},
  booktitle={International Conference on Machine Learning},
  pages={47280--47304},
  year={2024},
  organization={PMLR}
}

@inproceedings{
liu2024tuning,
title={Tuning Language Models by Proxy},
author={Alisa Liu and Xiaochuang Han and Yizhong Wang and Yulia Tsvetkov and Yejin Choi and Noah A. Smith},
booktitle={First Conference on Language Modeling},
year={2024},
url={https://openreview.net/forum?id=dribhnhm1i}
}

@article{madaan2023automix,
  title={Automix: Automatically mixing language models},
  author={Aggarwal, Pranjal and Madaan, Aman and Anand, Ankit and Potharaju, Srividya Pranavi and Mishra, Swaroop and Zhou, Pei and Gupta, Aditya and Rajagopal, Dheeraj and Kappaganthu, Karthik and Yang, Yiming and others},
  journal={Advances in Neural Information Processing Systems},
  volume={37},
  pages={131000--131034},
  year={2024}
}

@inproceedings{chen2020frugalml,
    author = {Lingjiao Chen and
Matei Zaharia and
James Y. Zou},
    booktitle = {Advances in Neural Information Processing Systems 33: Annual Conference
on Neural Information Processing Systems 2020, NeurIPS 2020, December
6-12, 2020, virtual},
    editor = {Hugo Larochelle and
Marc'Aurelio Ranzato and
Raia Hadsell and
Maria{-}Florina Balcan and
Hsuan{-}Tien Lin},
    title = {FrugalML: How to use {ML} Prediction APIs more accurately and cheaply},
    url = {https://proceedings.neurips.cc/paper/2020/hash/789ba2ae4d335e8a2ad283a3f7effced-Abstract.html},
    year = {2020}
}

@inproceedings{zhang2019bertscore,
    author = {Tianyi Zhang and
Varsha Kishore and
Felix Wu and
Kilian Q. Weinberger and
Yoav Artzi},
    booktitle = {Proc. of ICLR},
    title = {BERTScore: Evaluating Text Generation with {BERT}},
    url = {https://openreview.net/forum?id=SkeHuCVFDr},
    year = {2020}
}

@inproceedings{hsu2024calm,
  title={Calm: Contrasting large and small language models to verify grounded generation},
  author={Hsu, I-Hung and Wang, Zifeng and Le, Long and Werlen, Lesly Miculicich and Peng, Nanyun and Lee, Chen-Yu and Pfister, Tomas},
  booktitle={Findings of the Association for Computational Linguistics: ACL 2024},
  pages={12782--12803},
  year={2024}
}

@inproceedings{kuhnsemantic2023,
    author = {Kuhn, Lorenz and Gal, Yarin and Farquhar, Sebastian},
    booktitle = {The Eleventh International Conference on Learning Representations},
    title = {Semantic Uncertainty: Linguistic Invariances for Uncertainty Estimation in Natural Language Generation},
    year = {2023}
}

@inproceedings{manakul2023selfcheckgpt,
    author = {Manakul, Potsawee and Liusie, Adian and Gales, Mark},
    booktitle = {Proc. of EMNLP},
    title = {SelfCheckGPT: Zero-Resource Black-Box Hallucination Detection for Generative Large Language Models},
    year = {2023}
}

@inproceedings{ma2023query,
    author = {Ma, Xinbei and Gong, Yeyun and He, Pengcheng and Zhao, Hai and Duan, Nan},
    booktitle = {Proc. of EMNLP},
    title = {Query Rewriting in Retrieval-Augmented Large Language Models},
    year = {2023}
}

@inproceedings{witteveen2019paraphrasing,
    author = {Witteveen, Sam  and
Andrews, Martin},
    booktitle = {Proceedings of the 3rd Workshop on Neural Generation and Translation},
    title = {Paraphrasing with Large Language Models},
    url = {https://aclanthology.org/D19-5623},
    year = {2019}
}

@inproceedings{wang2021want,
    author = {Wang, Shuohang  and
Liu, Yang  and
Xu, Yichong  and
Zhu, Chenguang  and
Zeng, Michael},
    booktitle = {Findings of the Association for Computational Linguistics: EMNLP 2021},
    title = {Want To Reduce Labeling Cost? {GPT}-3 Can Help},
    url = {https://aclanthology.org/2021.findings-emnlp.354},
    year = {2021}
}

@inproceedings{lewis2020retrieval,
    author = {Patrick S. H. Lewis and
Ethan Perez and
Aleksandra Piktus and
Fabio Petroni and
Vladimir Karpukhin and
Naman Goyal and
Heinrich K{\"{u}}ttler and
Mike Lewis and
Wen{-}tau Yih and
Tim Rockt{\"{a}}schel and
Sebastian Riedel and
Douwe Kiela},
    booktitle = {Advances in Neural Information Processing Systems 33: Annual Conference
on Neural Information Processing Systems 2020, NeurIPS 2020, December
6-12, 2020, virtual},
    editor = {Hugo Larochelle and
Marc'Aurelio Ranzato and
Raia Hadsell and
Maria{-}Florina Balcan and
Hsuan{-}Tien Lin},
    title = {Retrieval-Augmented Generation for Knowledge-Intensive {NLP} Tasks},
    url = {https://proceedings.neurips.cc/paper/2020/hash/6b493230205f780e1bc26945df7481e5-Abstract.html},
    year = {2020}
}

@inproceedings{jiang2023structgpt,
    author = {Jiang, Jinhao and Zhou, Kun and Dong, Zican and Ye, Keming and Zhao, Wayne Xin and Wen, Ji-Rong},
    booktitle = {Proc. of EMNLP},
    title = {StructGPT: A General Framework for Large Language Model to Reason over Structured Data},
    year = {2023}
}

@inproceedings{
schick2024toolformer,
title={Toolformer: Language Models Can Teach Themselves to Use Tools},
author={Timo Schick and Jane Dwivedi-Yu and Roberto Dessi and Roberta Raileanu and Maria Lomeli and Eric Hambro and Luke Zettlemoyer and Nicola Cancedda and Thomas Scialom},
booktitle={Thirty-seventh Conference on Neural Information Processing Systems},
year={2023},
url={https://openreview.net/forum?id=Yacmpz84TH}
}

@article{raffel2020exploring,
    author = {Colin Raffel and
Noam Shazeer and
Adam Roberts and
Katherine Lee and
Sharan Narang and
Michael Matena and
Yanqi Zhou and
Wei Li and
Peter J. Liu},
    journal = {J. Mach. Learn. Res.},
    title = {Exploring the Limits of Transfer Learning with a Unified Text-to-Text
Transformer},
    url = {http://jmlr.org/papers/v21/20-074.html},
    year = {2020}
}

@article{gao2020pile,
    author = {Gao, Leo and Biderman, Stella and Black, Sid and Golding, Laurence and Hoppe, Travis and Foster, Charles and Phang, Jason and He, Horace and Thite, Anish and Nabeshima, Noa and others},
    journal = {arXiv preprint arXiv:2101.00027},
    title = {The pile: An 800gb dataset of diverse text for language modeling},
    url = {https://arxiv.org/abs/2101.00027},
year = {2021}
}

@inproceedings{villalobosposition,
    author = {Villalobos, Pablo and Ho, Anson and Sevilla, Jaime and Besiroglu, Tamay and Heim, Lennart and Hobbhahn, Marius},
    booktitle = {Forty-first International Conference on Machine Learning},
    title = {Position: Will we run out of data? Limits of LLM scaling based on human-generated data},
    year = {2024}
}

@article{marion2023less,
    author = {Marion, Max and {\"U}st{\"u}n, Ahmet and Pozzobon, Luiza and Wang, Alex and Fadaee, Marzieh and Hooker, Sara},
    journal = {arXiv preprint arXiv:2309.04564},
    title = {When less is more: Investigating data pruning for pretraining llms at scale},
    url = {https://arxiv.org/abs/2309.04564},
year = {2023}
}

@article{tirumala2024d4,
    author = {Tirumala, Kushal and Simig, Daniel and Aghajanyan, Armen and Morcos, Ari},
    journal = {Advances in Neural Information Processing Systems},
    title = {D4: Improving llm pretraining via document de-duplication and diversification},
    year = {2023}
}

@inproceedings{
penedo2023refinedweb,
title={The RefinedWeb Dataset for Falcon {LLM}: Outperforming Curated Corpora with Web Data Only},
author={Guilherme Penedo and Quentin Malartic and Daniel Hesslow and Ruxandra Cojocaru and Hamza Alobeidli and Alessandro Cappelli and Baptiste Pannier and Ebtesam Almazrouei and Julien Launay},
booktitle={Thirty-seventh Conference on Neural Information Processing Systems Datasets and Benchmarks Track},
year={2023},
url={https://openreview.net/forum?id=kM5eGcdCzq}
}

@inproceedings{wenzek2020ccnet,
    author = {Wenzek, Guillaume  and
Lachaux, Marie-Anne  and
Conneau, Alexis  and
Chaudhary, Vishrav  and
Guzm{\'a}n, Francisco  and
Joulin, Armand  and
Grave, Edouard},
    booktitle = {Proceedings of the Twelfth Language Resources and Evaluation Conference},
    isbn = {979-10-95546-34-4},
    language = {English},
    title = {{CCN}et: Extracting High Quality Monolingual Datasets from Web Crawl Data},
    url = {https://aclanthology.org/2020.lrec-1.494},
    year = {2020}
}

@inproceedings{du2022glam,
    author = {Nan Du and
Yanping Huang and
Andrew M. Dai and
Simon Tong and
Dmitry Lepikhin and
Yuanzhong Xu and
Maxim Krikun and
Yanqi Zhou and
Adams Wei Yu and
Orhan Firat and
Barret Zoph and
Liam Fedus and
Maarten P. Bosma and
Zongwei Zhou and
Tao Wang and
Yu Emma Wang and
Kellie Webster and
Marie Pellat and
Kevin Robinson and
Kathleen S. Meier{-}Hellstern and
Toju Duke and
Lucas Dixon and
Kun Zhang and
Quoc V. Le and
Yonghui Wu and
Zhifeng Chen and
Claire Cui},
    booktitle = {International Conference on Machine Learning, {ICML} 2022, 17-23 July
2022, Baltimore, Maryland, {USA}},
    editor = {Kamalika Chaudhuri and
Stefanie Jegelka and
Le Song and
Csaba Szepesv{\'{a}}ri and
Gang Niu and
Sivan Sabato},
    series = {Proceedings of Machine Learning Research},
    title = {GLaM: Efficient Scaling of Language Models with Mixture-of-Experts},
    url = {https://proceedings.mlr.press/v162/du22c.html},
    year = {2022}
}

@article{chowdhery2023palm,
    author = {Chowdhery, Aakanksha and Narang, Sharan and Devlin, Jacob and Bosma, Maarten and Mishra, Gaurav and Roberts, Adam and Barham, Paul and Chung, Hyung Won and Sutton, Charles and Gehrmann, Sebastian and others},
    journal = {Journal of Machine Learning Research},
    number = {240},
    title = {Palm: Scaling language modeling with pathways},
    year = {2023}
}

@article{xie2023data,
    author = {Xie, Sang Michael and Santurkar, Shibani and Ma, Tengyu and Liang, Percy S},
    journal = {Advances in Neural Information Processing Systems},
    title = {Data selection for language models via importance resampling},
    year = {2023}
}

@inproceedings{wettigqurating,
    author = {Wettig, Alexander and Gupta, Aatmik and Malik, Saumya and Chen, Danqi},
    booktitle = {Forty-first International Conference on Machine Learning},
    title = {QuRating: Selecting High-Quality Data for Training Language Models},
    year = {2024}
}

@article{bai2022training,
    author = {Bai, Yuntao and Jones, Andy and Ndousse, Kamal and Askell, Amanda and Chen, Anna and DasSarma, Nova and Drain, Dawn and Fort, Stanislav and Ganguli, Deep and Henighan, Tom and others},
    journal = {arXiv preprint arXiv:2204.05862},
    title = {Training a helpful and harmless assistant with reinforcement learning from human feedback},
    url = {https://arxiv.org/abs/2204.05862},
year = {2022}
}

@article{ouyang2022training,
    author = {Ouyang, Long and Wu, Jeffrey and Jiang, Xu and Almeida, Diogo and Wainwright, Carroll and Mishkin, Pamela and Zhang, Chong and Agarwal, Sandhini and Slama, Katarina and Ray, Alex and others},
    journal = {Advances in neural information processing systems},
    title = {Training language models to follow instructions with human feedback},
    year = {2022}
}

@inproceedings{
zhou2024lima,
title={{LIMA}: Less Is More for Alignment},
author={Chunting Zhou and Pengfei Liu and Puxin Xu and Srini Iyer and Jiao Sun and Yuning Mao and Xuezhe Ma and Avia Efrat and Ping Yu and LILI YU and Susan Zhang and Gargi Ghosh and Mike Lewis and Luke Zettlemoyer and Omer Levy},
booktitle={Thirty-seventh Conference on Neural Information Processing Systems},
year={2023},
url={https://openreview.net/forum?id=KBMOKmX2he}
}

@inproceedings{longpre2023flan,
    author = {Longpre, Shayne and Hou, Le and Vu, Tu and Webson, Albert and Chung, Hyung Won and Tay, Yi and Zhou, Denny and Le, Quoc V and Zoph, Barret and Wei, Jason and others},
    booktitle = {International Conference on Machine Learning},
    organization = {PMLR},
    title = {The flan collection: Designing data and methods for effective instruction tuning},
    year = {2023}
}

@inproceedings{
chen2023alpagasus,
title={{DELIFT}: Data Efficient Language model Instruction Fine-Tuning},
author={Ishika Agarwal and Krishnateja Killamsetty and Lucian Popa and Marina Danilevsky},
booktitle={The Thirteenth International Conference on Learning Representations},
year={2025},
url={https://openreview.net/forum?id=Fty0wTcemV}
}

@article{du2023mods,
    author = {Du, Qianlong and Zong, Chengqing and Zhang, Jiajun},
    journal = {arXiv preprint arXiv:2311.15653},
    title = {Mods: Model-oriented data selection for instruction tuning},
    url = {https://arxiv.org/abs/2311.15653},
year = {2023}
}

@inproceedings{hedeberta2021,
    author = {Pengcheng He and
Xiaodong Liu and
Jianfeng Gao and
Weizhu Chen},
    booktitle = {Proc. of ICLR},
    title = {Deberta: decoding-Enhanced Bert with Disentangled Attention},
    url = {https://openreview.net/forum?id=XPZIaotutsD},
    year = {2021}
}

@inproceedings{xialess2024,
    author = {Xia, Mengzhou and Malladi, Sadhika and Gururangan, Suchin and Arora, Sanjeev and Chen, Danqi},
    booktitle = {Forty-first International Conference on Machine Learning},
    title = {LESS: Selecting Influential Data for Targeted Instruction Tuning},
    year = {2024}
}

@inproceedings{longpre2024pretrainer,
    author = {Longpre, Shayne and Yauney, Gregory and Reif, Emily and Lee, Katherine and Roberts, Adam and Zoph, Barret and Zhou, Denny and Wei, Jason and Robinson, Kevin and Mimno, David and others},
    booktitle = {Proceedings of the 2024 Conference of the North American Chapter of the Association for Computational Linguistics: Human Language Technologies (Volume 1: Long Papers)},
    title = {A Pretrainer’s Guide to Training Data: Measuring the Effects of Data Age, Domain Coverage, Quality, \& Toxicity},
    year = {2024}
}

@inproceedings{liumakes2024,
    author = {Liu, Wei and Zeng, Weihao and He, Keqing and Jiang, Yong and He, Junxian},
    booktitle = {The Twelfth International Conference on Learning Representations},
    title = {What Makes Good Data for Alignment? A Comprehensive Study of Automatic Data Selection in Instruction Tuning},
    year = {2024}
}

@article{
albalak2024survey,
title={A Survey on Data Selection for Language Models},
author={Alon Albalak and Yanai Elazar and Sang Michael Xie and Shayne Longpre and Nathan Lambert and Xinyi Wang and Niklas Muennighoff and Bairu Hou and Liangming Pan and Haewon Jeong and Colin Raffel and Shiyu Chang and Tatsunori Hashimoto and William Yang Wang},
journal={Transactions on Machine Learning Research},
issn={2835-8856},
year={2024},
url={https://openreview.net/forum?id=XfHWcNTSHp}
}

@article{shen2023large,
    author = {Shen, Tianhao and Jin, Renren and Huang, Yufei and Liu, Chuang and Dong, Weilong and Guo, Zishan and Wu, Xinwei and Liu, Yan and Xiong, Deyi},
    journal = {arXiv preprint arXiv:2309.15025},
    title = {Large language model alignment: A survey},
    url = {https://arxiv.org/abs/2309.15025},
year = {2023}
}

@article{guo2024improving,
    author = {Guo, Yue and Yang, Yi},
    journal = {arXiv preprint arXiv:2406.19032},
    title = {Improving Weak-to-Strong Generalization with Reliability-Aware Alignment},
    url = {https://arxiv.org/abs/2406.19032},
year = {2024}
}

@inproceedings{
ji2024aligner,
title={Aligner: Efficient Alignment by Learning to Correct},
author={Jiaming Ji and Boyuan Chen and Hantao Lou and Donghai Hong and Borong Zhang and Xuehai Pan and Tianyi Qiu and Juntao Dai and Yaodong Yang},
booktitle={The Thirty-eighth Annual Conference on Neural Information Processing Systems},
year={2024},
url={https://openreview.net/forum?id=kq166jACVP}
}

@article{lang2024theoretical,
  title={Theoretical analysis of weak-to-strong generalization},
  author={Lang, Hunter and Sontag, David and Vijayaraghavan, Aravindan},
  journal={Advances in neural information processing systems},
  volume={37},
  pages={46837--46880},
  year={2024}
}

@article{wu2022sustainable,
    author = {Wu, Carole-Jean and Raghavendra, Ramya and Gupta, Udit and Acun, Bilge and Ardalani, Newsha and Maeng, Kiwan and Chang, Gloria and Aga, Fiona and Huang, Jinshi and Bai, Charles and others},
    journal = {Proceedings of Machine Learning and Systems},
    title = {Sustainable ai: Environmental implications, challenges and opportunities},
    year = {2022}
}

@inproceedings{
kag2023efficient,
title={Efficient Edge Inference by Selective Query},
author={Anil Kag and Igor Fedorov and Aditya Gangrade and Paul Whatmough and Venkatesh Saligrama},
booktitle={The Eleventh International Conference on Learning Representations },
year={2023},
url={https://openreview.net/forum?id=jpR98ZdIm2q}
}

@inproceedings{dhuliawala2023chain,
  title={Chain-of-verification reduces hallucination in large language models},
  author={Dhuliawala, Shehzaad and Komeili, Mojtaba and Xu, Jing and Raileanu, Roberta and Li, Xian and Celikyilmaz, Asli and Weston, Jason},
  booktitle={Findings of the association for computational linguistics: ACL 2024},
  pages={3563--3578},
  year={2024}
}

@inproceedings{tian2023just,
  title={Just ask for calibration: Strategies for eliciting calibrated confidence scores from language models fine-tuned with human feedback},
  author={Tian, Katherine and Mitchell, Eric and Zhou, Allan and Sharma, Archit and Rafailov, Rafael and Yao, Huaxiu and Finn, Chelsea and Manning, Christopher D},
  booktitle={Proceedings of the 2023 Conference on Empirical Methods in Natural Language Processing},
  pages={5433--5442},
  year={2023}
}

@inproceedings{ding2024hybrid,
  title={Hybrid llm: Cost-efficient and quality-aware query routing},
  author={Ding, Dujian and Mallick, Ankur and Wang, Chi and Sim, Robert and Mukherjee, Subhabrata and R{\"u}hle, Victor and Lakshmanan, Laks VS and Awadallah, Ahmed Hassan},
  booktitle={The Twelfth International Conference on Learning Representations},
  year={2024}
}

@inproceedings{jiang2023llm,
    author = {Jiang, Dongfu and Ren, Xiang and Lin, Bill Yuchen},
    booktitle = {The 61st Annual Meeting Of The Association For Computational Linguistics},
    title = {LLM-Blender: Ensembling Large Language Models with Pairwise Ranking and Generative Fusion},
    year = {2023}
}

@article{narayanan2023tryage,
    author = {Narayanan Hari, Surya and Thomson, Matt},
    journal = {arXiv preprint arXiv:2308.11601},
    title = {Tryage: Real-time, intelligent Routing of User Prompts to Large Language Models},
    year = {2023},
}

@inproceedings{lu2023routing,
  title={Routing to the expert: Efficient reward-guided ensemble of large language models},
  author={Lu, Keming and Yuan, Hongyi and Lin, Runji and Lin, Junyang and Yuan, Zheng and Zhou, Chang and Zhou, Jingren},
  booktitle={Proceedings of the 2024 Conference of the North American Chapter of the Association for Computational Linguistics: Human Language Technologies (Volume 1: Long Papers)},
  pages={1964--1974},
  year={2024}
}

@inproceedings{lee2024orchestrallm,
    author = {Lee, Chia-Hsuan and Cheng, Hao and Ostendorf, Mari},
    booktitle = {Proceedings of the 2024 Conference of the North American Chapter of the Association for Computational Linguistics: Human Language Technologies (Volume 1: Long Papers)},
    title = {OrchestraLLM: Efficient Orchestration of Language Models for Dialogue State Tracking},
    year = {2024}
}

@inproceedings{
hu2024routerbench,
title={RouterBench: A Benchmark for Multi-{LLM} Routing System},
author={Qitian Jason Hu and Jacob Bieker and Xiuyu Li and Nan Jiang and Benjamin Keigwin and Gaurav Ranganath and Kurt Keutzer and Shriyash Kaustubh Upadhyay},
booktitle={Agentic Markets Workshop at ICML 2024},
year={2024},
url={https://openreview.net/forum?id=IVXmV8Uxwh}
}

@inproceedings{
shnitzer2023large,
title={Large Language Model Routing with Benchmark Datasets},
author={Tal Shnitzer and Anthony Ou and M{\'\i}rian Silva and Kate Soule and Yuekai Sun and Justin Solomon and Neil Thompson and Mikhail Yurochkin},
booktitle={First Conference on Language Modeling},
year={2024},
url={https://openreview.net/forum?id=Zb0ajZ7vAt}
}

@inproceedings{vsakota2024fly,
    author = {{\v{S}}akota, Marija and Peyrard, Maxime and West, Robert},
    booktitle = {Proceedings of the 17th ACM International Conference on Web Search and Data Mining},
    title = {Fly-swat or cannon? cost-effective language model choice via meta-modeling},
    year = {2024}
}

@article{xia2024unlocking,
  title={Unlocking efficiency in large language model inference: A comprehensive survey of speculative decoding},
  author={Xia, Heming and Yang, Zhe and Dong, Qingxiu and Wang, Peiyi and Li, Yongqi and Ge, Tao and Liu, Tianyu and Li, Wenjie and Sui, Zhifang},
  journal={Findings of the Association for Computational Linguistics: ACL 2024},
  pages={7655--7671},
  year={2024}
}

@inproceedings{
shen2024hugginggpt,
title={Hugging{GPT}: Solving {AI} Tasks with Chat{GPT} and its Friends in Hugging Face},
author={Yongliang Shen and Kaitao Song and Xu Tan and Dongsheng Li and Weiming Lu and Yueting Zhuang},
booktitle={Thirty-seventh Conference on Neural Information Processing Systems},
year={2023},
url={https://openreview.net/forum?id=yHdTscY6Ci}
}

@article{chang2024survey,
    author = {Chang, Yupeng and Wang, Xu and Wang, Jindong and Wu, Yuan and Yang, Linyi and Zhu, Kaijie and Chen, Hao and Yi, Xiaoyuan and Wang, Cunxiang and Wang, Yidong and others},
    journal = {ACM Transactions on Intelligent Systems and Technology},
    number = {3},
    title = {A survey on evaluation of large language models},
    year = {2024}
}

@inproceedings{papineni2002bleu,
    author = {Papineni, Kishore  and
Roukos, Salim  and
Ward, Todd  and
Zhu, Wei-Jing},
    booktitle = {Proc. of ACL},
    title = {{B}leu: a Method for Automatic Evaluation of Machine Translation},
    url = {https://aclanthology.org/P02-1040},
    year = {2002}
}

@inproceedings{lin2004rouge,
    author = {Lin, Chin-Yew},
    booktitle = {Text Summarization Branches Out},
    title = {{ROUGE}: A Package for Automatic Evaluation of Summaries},
    url = {https://aclanthology.org/W04-1013},
    year = {2004}
}

@inproceedings{liu2016not,
    author = {Liu, Chia-Wei  and
Lowe, Ryan  and
Serban, Iulian  and
Noseworthy, Mike  and
Charlin, Laurent  and
Pineau, Joelle},
    booktitle = {Proc. of EMNLP},
    title = {How {NOT} To Evaluate Your Dialogue System: An Empirical Study of Unsupervised Evaluation Metrics for Dialogue Response Generation},
    url = {https://aclanthology.org/D16-1230},
    year = {2016}
}

@inproceedings{yuan2021bartscore,
    author = {Weizhe Yuan and
Graham Neubig and
Pengfei Liu},
    booktitle = {Advances in Neural Information Processing Systems 34: Annual Conference
on Neural Information Processing Systems 2021, NeurIPS 2021, December
6-14, 2021, virtual},
    editor = {Marc'Aurelio Ranzato and
Alina Beygelzimer and
Yann N. Dauphin and
Percy Liang and
Jennifer Wortman Vaughan},
    title = {BARTScore: Evaluating Generated Text as Text Generation},
    url = {https://proceedings.neurips.cc/paper/2021/hash/e4d2b6e6fdeca3e60e0f1a62fee3d9dd-Abstract.html},
    year = {2021}
}

@inproceedings{min2023factscore,
    author = {Min, Sewon and Krishna, Kalpesh and Lyu, Xinxi and Lewis, Mike and Yih, Wen-tau and Koh, Pang and Iyyer, Mohit and Zettlemoyer, Luke and Hajishirzi, Hannaneh},
    booktitle = {Proc. of EMNLP},
    title = {FActScore: Fine-grained Atomic Evaluation of Factual Precision in Long Form Text Generation},
    year = {2023}
}

@inproceedings{zhang2024safetybench,
    author = {Zhang, Zhexin and Lei, Leqi and Wu, Lindong and Sun, Rui and Huang, Yongkang and Long, Chong and Liu, Xiao and Lei, Xuanyu and Tang, Jie and Huang, Minlie},
    booktitle = {Proc. of ACL},
    title = {SafetyBench: Evaluating the Safety of Large Language Models},
    year = {2024}
}

@article{huang2023look,
  title={Look before you leap: An exploratory study of uncertainty analysis for large language models},
  author={Huang, Yuheng and Song, Jiayang and Wang, Zhijie and Zhao, Shengming and Chen, Huaming and Juefei-Xu, Felix and Ma, Lei},
  journal={IEEE Transactions on Software Engineering},
  volume={51},
  number={2},
  pages={413--429},
  year={2025},
  publisher={IEEE}
}

@article{siriwardhana2023improving,
    author = {Siriwardhana, Shamane and Weerasekera, Rivindu and Wen, Elliott and Kaluarachchi, Tharindu and Rana, Rajib and Nanayakkara, Suranga},
    journal = {Transactions of the Association for Computational Linguistics},
    title = {Improving the domain adaptation of retrieval augmented generation (RAG) models for open domain question answering},
    year = {2023}
}

@inproceedings{shi2023replug,
  title={Replug: Retrieval-augmented black-box language models},
  author={Shi, Weijia and Min, Sewon and Yasunaga, Michihiro and Seo, Minjoon and James, Richard and Lewis, Mike and Zettlemoyer, Luke and Yih, Wen-tau},
  booktitle={Proceedings of the 2024 Conference of the North American Chapter of the Association for Computational Linguistics: Human Language Technologies (Volume 1: Long Papers)},
  pages={8371--8384},
  year={2024}
}

@inproceedings{shuster2021retrieval,
    author = {Shuster, Kurt  and
Poff, Spencer  and
Chen, Moya  and
Kiela, Douwe  and
Weston, Jason},
    booktitle = {Findings of the Association for Computational Linguistics: EMNLP 2021},
    title = {Retrieval Augmentation Reduces Hallucination in Conversation},
    url = {https://aclanthology.org/2021.findings-emnlp.320},
    year = {2021}
}

@inproceedings{trivedi2023interleaving,
    author = {Trivedi, Harsh and Balasubramanian, Niranjan and Khot, Tushar and Sabharwal, Ashish},
    booktitle = {Proc. of ACL},
    title = {Interleaving Retrieval with Chain-of-Thought Reasoning for Knowledge-Intensive Multi-Step Questions},
    year = {2023}
}

@inproceedings{
asaiself2023,
title={Self-{RAG}: Learning to Retrieve, Generate, and Critique through Self-Reflection},
author={Akari Asai and Zeqiu Wu and Yizhong Wang and Avirup Sil and Hannaneh Hajishirzi},
booktitle={The Twelfth International Conference on Learning Representations},
year={2024},
url={https://openreview.net/forum?id=hSyW5go0v8}
}

@inproceedings{nie2023cross,
    author = {Nie, Ercong and Liang, Sheng and Schmid, Helmut and Sch{\"u}tze, Hinrich},
    booktitle = {Findings of the Association for Computational Linguistics: ACL 2023},
    title = {Cross-Lingual Retrieval Augmented Prompt for Low-Resource Languages},
    year = {2023}
}

@inproceedings{xiong2024benchmarking,
  title={Benchmarking retrieval-augmented generation for medicine},
  author={Xiong, Guangzhi and Jin, Qiao and Lu, Zhiyong and Zhang, Aidong},
  booktitle={Findings of the Association for Computational Linguistics: ACL 2024},
  pages={6233--6251},
  year={2024}
}

@article{yue2023disc,
    author = {Yue, Shengbin and Chen, Wei and Wang, Siyuan and Li, Bingxuan and Shen, Chenchen and Liu, Shujun and Zhou, Yuxuan and Xiao, Yao and Yun, Song and Huang, Xuanjing and others},
    journal = {arXiv preprint arXiv:2309.11325},
    title = {DISC-LawLLM: Fine-tuning Large Language Models for Intelligent Legal Services},
    year = {2023},
}

@article{izacard2021unsupervised,
  author       = {Gautier Izacard and
                  Mathilde Caron and
                  Lucas Hosseini and
                  Sebastian Riedel and
                  Piotr Bojanowski and
                  Armand Joulin and
                  Edouard Grave},
  title        = {Unsupervised Dense Information Retrieval with Contrastive Learning},
  journal      = {Trans. Mach. Learn. Res.},
  volume       = {2022},
  year         = {2022},
  url          = {https://openreview.net/forum?id=jKN1pXi7b0},
  bibsource    = {dblp computer science bibliography, https://dblp.org}
}

@article{robertson2009probabilistic,
    author = {Robertson, Stephen and Zaragoza, Hugo and others},
    journal = {Foundations and Trends{\textregistered} in Information Retrieval},
    number = {4},
    title = {The probabilistic relevance framework: BM25 and beyond},
    year = {2009}
}

@article{wang2023knowledgpt,
    author = {Wang, Xintao and Yang, Qianwen and Qiu, Yongting and Liang, Jiaqing and He, Qianyu and Gu, Zhouhong and Xiao, Yanghua and Wang, Wei},
    journal = {arXiv preprint arXiv:2308.11761},
    title = {Knowledgpt: Enhancing large language models with retrieval and storage access on knowledge bases},
    url = {https://arxiv.org/abs/2308.11761},
year = {2023}
}

@article{pan2022end,
    author = {Pan, Feifei and Canim, Mustafa and Glass, Michael and Gliozzo, Alfio and Hendler, James},
    journal = {arXiv preprint arXiv:2203.16714},
    title = {End-to-end table question answering via retrieval-augmented generation},
    url = {https://arxiv.org/abs/2203.16714},
year = {2022}
}

@inproceedings{
yoranmaking,
title={Making Retrieval-Augmented Language Models Robust to Irrelevant Context},
author={Ori Yoran and Tomer Wolfson and Ori Ram and Jonathan Berant},
booktitle={The Twelfth International Conference on Learning Representations},
year={2024},
url={https://openreview.net/forum?id=ZS4m74kZpH}
}

@article{liu2023pre,
    author = {Liu, Pengfei and Yuan, Weizhe and Fu, Jinlan and Jiang, Zhengbao and Hayashi, Hiroaki and Neubig, Graham},
    journal = {ACM Computing Surveys},
    number = {9},
    title = {Pre-train, prompt, and predict: A systematic survey of prompting methods in natural language processing},
    year = {2023}
}

@inproceedings{dong2022survey,
title = "A Survey on In-context Learning",
    author = "Dong, Qingxiu  and
      Li, Lei  and
      Dai, Damai  and
      Zheng, Ce  and
      Ma, Jingyuan  and
      Li, Rui  and
      Xia, Heming  and
      Xu, Jingjing  and
      Wu, Zhiyong  and
      Chang, Baobao  and
      Sun, Xu  and
      Li, Lei  and
      Sui, Zhifang",
    editor = "Al-Onaizan, Yaser  and
      Bansal, Mohit  and
      Chen, Yun-Nung",
    booktitle = "Proceedings of the 2024 Conference on Empirical Methods in Natural Language Processing",
    month = nov,
    year = "2024",
    address = "Miami, Florida, USA",
    publisher = "Association for Computational Linguistics",
    url = "https://aclanthology.org/2024.emnlp-main.64/",
    doi = "10.18653/v1/2024.emnlp-main.64",
    pages = "1107--1128"
}

@inproceedings{vernikos2024small,
    author = {Vernikos, Giorgos and Bra{\v{z}}inskas, Arthur and Adamek, Jakub and Mallinson, Jonathan and Severyn, Aliaksei and Malmi, Eric},
    booktitle = {Proceedings of the 18th Conference of the European Chapter of the Association for Computational Linguistics (Volume 1: Long Papers)},
    title = {Small Language Models Improve Giants by Rewriting Their Outputs},
    year = {2024}
}

@inproceedings{sennrich2024mitigating,
    author = {Sennrich, Rico and Vamvas, Jannis and Mohammadshahi, Alireza},
    booktitle = {Proceedings of the 18th Conference of the European Chapter of the Association for Computational Linguistics (Volume 2: Short Papers)},
    title = {Mitigating Hallucinations and Off-target Machine Translation with Source-Contrastive and Language-Contrastive Decoding},
    year = {2024}
}

@article{o2023contrastive,
    author = {O'Brien, Sean and Lewis, Mike},
    journal = {arXiv preprint arXiv:2309.09117},
    title = {Contrastive decoding improves reasoning in large language models},
    url = {https://arxiv.org/abs/2309.09117},
year = {2023}
}

@inproceedings{zhang2024cogenesis,
  title={Cogenesis: A framework collaborating large and small language models for secure context-aware instruction following},
  author={Zhang, Kaiyan and Wang, Jianyu and Hua, Ermo and Qi, Biqing and Ding, Ning and Zhou, Bowen},
  booktitle={Proceedings of the 62nd Annual Meeting of the Association for Computational Linguistics (Volume 1: Long Papers)},
  pages={4295--4312},
  year={2024}
}

@inproceedings{pinter2017mimicking,
    author = {Pinter, Yuval  and
Guthrie, Robert  and
Eisenstein, Jacob},
    booktitle = {Proc. of EMNLP},
    title = {Mimicking Word Embeddings using Subword {RNN}s},
    url = {https://aclanthology.org/D17-1010},
    year = {2017}
}

@inproceedings{chen2024reconfidencing,
  title={Reconfidencing llms from the grouping loss perspective},
  author={Chen, Lihu and Perez-Lebel, Alexandre and Suchanek, Fabian and Varoquaux, Ga{\"e}l},
  booktitle={Findings of the Association for Computational Linguistics: EMNLP 2024},
  pages={1567--1581},
  year={2024}
}

@article{hinton2015distilling,
  title={Distilling the knowledge in a neural network},
  author={Hinton, Geoffrey and Vinyals, Oriol and Dean, Jeff},
  journal = {arXiv preprint arXiv:1503.02531},
  year={2015},
}

@article{sanh2019distilbert,
  title={DistilBERT, a distilled version of BERT: smaller, faster, cheaper and lighter},
  author={Sanh, Victor and Debut, Lysandre and Chaumond, Julien and Wolf, Thomas},
  journal = {arXiv preprint arXiv:1910.01108},
  year={2019},
}

@article{yao2022zeroquant,
    author = {Yao, Zhewei and Yazdani Aminabadi, Reza and Zhang, Minjia and Wu, Xiaoxia and Li, Conglong and He, Yuxiong},
    journal = {Advances in Neural Information Processing Systems},
    title = {Zeroquant: Efficient and affordable post-training quantization for large-scale transformers},
    year = {2022}
}

@article{gou2021knowledge,
    author = {Gou, Jianping and Yu, Baosheng and Maybank, Stephen J and Tao, Dacheng},
    journal = {International Journal of Computer Vision},
    number = {6},
    title = {Knowledge distillation: A survey},
    year = {2021}
}

@inproceedings{li2023symbolic,
    author = {Li, Liunian Harold and Hessel, Jack and Yu, Youngjae and Ren, Xiang and Chang, Kai-Wei and Choi, Yejin},
    booktitle = {Proc. of ACL},
    title = {Symbolic Chain-of-Thought Distillation: Small Models Can Also “Think” Step-by-Step},
    year = {2023}
}

@article{li2022explanations,
    author = {Li, Shiyang and Chen, Jianshu and Shen, Yelong and Chen, Zhiyu and Zhang, Xinlu and Li, Zekun and Wang, Hong and Qian, Jing and Peng, Baolin and Mao, Yi and others},
    journal = {arXiv preprint arXiv:2210.06726},
    title = {Explanations from large language models make small reasoners better},
    url = {https://arxiv.org/abs/2210.06726},
year = {2022}
}

@inproceedings{fu2023specializing,
    author = {Fu, Yao and Peng, Hao and Ou, Litu and Sabharwal, Ashish and Khot, Tushar},
    booktitle = {International Conference on Machine Learning},
    organization = {PMLR},
    title = {Specializing smaller language models towards multi-step reasoning},
    year = {2023}
}

@article{tian2024tinyllm,
    author = {Tian, Yijun and Han, Yikun and Chen, Xiusi and Wang, Wei and Chawla, Nitesh V},
    journal = {arXiv preprint arXiv:2402.04616},
    title = {Tinyllm: Learning a small student from multiple large language models},
    url = {https://arxiv.org/abs/2402.04616},
year = {2024}
}

@inproceedings{ye2022zerogen,
    author = {Ye, Jiacheng  and
Gao, Jiahui  and
Li, Qintong  and
Xu, Hang  and
Feng, Jiangtao  and
Wu, Zhiyong  and
Yu, Tao  and
Kong, Lingpeng},
    booktitle = {Proc. of EMNLP},
    title = {{Z}ero{G}en: Efficient Zero-shot Learning via Dataset Generation},
    url = {https://aclanthology.org/2022.emnlp-main.801},
    year = {2022}
}

@article{meng2022generating,
    author = {Meng, Yu and Huang, Jiaxin and Zhang, Yu and Han, Jiawei},
    journal = {Advances in Neural Information Processing Systems},
    title = {Generating training data with language models: Towards zero-shot language understanding},
    year = {2022}
}

@inproceedings{chung2023increasing,
    author = {Chung, John and Kamar, Ece and Amershi, Saleema},
    booktitle = {Proc. of ACL},
    title = {Increasing Diversity While Maintaining Accuracy: Text Data Generation with Large Language Models and Human Interventions},
    year = {2023}
}

@article{tang2023does,
    author = {Tang, Ruixiang and Han, Xiaotian and Jiang, Xiaoqian and Hu, Xia},
    journal = {arXiv preprint arXiv:2303.04360},
    title = {Does synthetic data generation of llms help clinical text mining?},
    url = {https://arxiv.org/abs/2303.04360},
year = {2023}
}

@inproceedings{josifoski2023exploiting,
    author = {Josifoski, Martin and Sakota, Marija and Peyrard, Maxime and West, Robert},
    booktitle = {Proc. of EMNLP},
    title = {Exploiting Asymmetry for Synthetic Training Data Generation: SynthIE and the Case of Information Extraction},
    year = {2023}
}

@inproceedings{hartvigsen2022toxigen,
    author = {Hartvigsen, Thomas  and
Gabriel, Saadia  and
Palangi, Hamid  and
Sap, Maarten  and
Ray, Dipankar  and
Kamar, Ece},
    booktitle = {Proc. of ACL},
    title = {{T}oxi{G}en: A Large-Scale Machine-Generated Dataset for Adversarial and Implicit Hate Speech Detection},
    url = {https://aclanthology.org/2022.acl-long.234},
    year = {2022}
}

@inproceedings{gao2022self,
  title={Self-guided noise-free data generation for efficient zero-shot learning},
  author={Gao, Jiahui and Pi, Renjie and Yong, Lin and Xu, Hang and Ye, Jiacheng and Wu, Zhiyong and Zhang, Weizhong and Liang, Xiaodan and Li, Zhenguo and Kong, Lingpeng},
  booktitle={The Eleventh international conference on learning representations},
  year={2023}
}

@inproceedings{li2023synthetic,
    author = {Li, Zhuoyan and Zhu, Hangxiao and Lu, Zhuoran and Yin, Ming},
    booktitle = {Proc. of EMNLP},
    title = {Synthetic Data Generation with Large Language Models for Text Classification: Potential and Limitations},
    year = {2023}
}

@inproceedings{ding2024data,
  title={Data augmentation using llms: Data perspectives, learning paradigms and challenges},
  author={Ding, Bosheng and Qin, Chengwei and Zhao, Ruochen and Luo, Tianze and Li, Xinze and Chen, Guizhen and Xia, Wenhan and Hu, Junjie and Tuan, Luu Anh and Joty, Shafiq},
  booktitle={Findings of the Association for Computational Linguistics: ACL 2024},
  pages={1679--1705},
  year={2024}
}

@article{chen2023empirical,
    author = {Chen, Jiaao and Tam, Derek and Raffel, Colin and Bansal, Mohit and Yang, Diyi},
    journal = {Transactions of the Association for Computational Linguistics},
    title = {An empirical survey of data augmentation for limited data learning in nlp},
    year = {2023}
}

@article{mi2022improving,
    author = {Mi, Chenggang and Xie, Lei and Zhang, Yanning},
    journal = {Neural Networks},
    title = {Improving data augmentation for low resource speech-to-text translation with diverse paraphrasing},
    year = {2022}
}

@inproceedings{sahu2022data,
    author = {Sahu, Gaurav  and
Rodriguez, Pau  and
Laradji, Issam  and
Atighehchian, Parmida  and
Vazquez, David  and
Bahdanau, Dzmitry},
    booktitle = {Proceedings of the 4th Workshop on NLP for Conversational AI},
    title = {Data Augmentation for Intent Classification with Off-the-shelf Large Language Models},
    url = {https://aclanthology.org/2022.nlp4convai-1.5},
    year = {2022}
}

@inproceedings{chen2022weakly,
    author = {Chen, Maximillian and Papangelis, Alexandros and Tao, Chenyang and Rosenbaum, Andy and Kim, Seokhwan and Liu, Yang and Yu, Zhou and Hakkani-Tur, Dilek},
    booktitle = {NeurIPS 2022 Workshop on Synthetic Data for Empowering ML Research},
    title = {Weakly Supervised Data Augmentation Through Prompting for Dialogue Understanding},
    year = {2022}
}

@article{zhu2023survey,
  title={A Survey on Model Compression for Large Language Models},
  author={Zhu, Xunyu and Li, Jian and Liu, Yong and Ma, Can and Wang, Weiping},
  journal={Transactions of the Association for Computational Linguistics},
  volume={12},
  pages={1556--1577},
  year={2024}
}

@inproceedings{jiang2023lion,
    author = {Jiang, Yuxin and Chan, Chunkit and Chen, Mingyang and Wang, Wei},
    booktitle = {Proc. of EMNLP},
    title = {Lion: Adversarial Distillation of Proprietary Large Language Models},
    year = {2023}
}

@article{xu2024survey,
    author = {Xu, Xiaohan and Li, Ming and Tao, Chongyang and Shen, Tao and Cheng, Reynold and Li, Jinyang and Xu, Can and Tao, Dacheng and Zhou, Tianyi},
    journal = {arXiv preprint arXiv:2402.13116},
    title = {A survey on knowledge distillation of large language models},
    url = {https://arxiv.org/abs/2402.13116},
year = {2024}
}

@inproceedings{gu2024minillm,
    author = {Gu, Yuxian and Dong, Li and Wei, Furu and Huang, Minlie},
    booktitle = {The Twelfth International Conference on Learning Representations},
    title = {MiniLLM: Knowledge distillation of large language models},
    year = {2024}
}

@article{ollion2023chatgpt,
    author = {Ollion, Etienne and Shen, Rubing and Macanovic, Ana and Chatelain, Arnault},
    journal = {SocArXiv preprint},
    title = {ChatGPT for Text Annotation? Mind the Hype},
    year = {2023}
}

@inproceedings{
bansal2024smaller,
title={Smaller, Weaker, Yet Better: Training {LLM} Reasoners via Compute-Optimal Sampling},
author={Hritik Bansal and Arian Hosseini and Rishabh Agarwal and Vinh Q. Tran and Mehran Kazemi},
booktitle={The Thirteenth International Conference on Learning Representations},
year={2025},
url={https://openreview.net/forum?id=3OyaXFQuDl}
}

@inproceedings{liang2023less,
    author = {Liang, Chen and Zuo, Simiao and Zhang, Qingru and He, Pengcheng and Chen, Weizhu and Zhao, Tuo},
    booktitle = {International Conference on Machine Learning},
    organization = {PMLR},
    title = {Less is more: Task-aware layer-wise distillation for language model compression},
    year = {2023}
}

@inproceedings{liu2023llm,
  title={Llm-qat: Data-free quantization aware training for large language models},
  author={Liu, Zechun and Oguz, Barlas and Zhao, Changsheng and Chang, Ernie and Stock, Pierre and Mehdad, Yashar and Shi, Yangyang and Krishnamoorthi, Raghuraman and Chandra, Vikas},
  booktitle={Findings of the Association for Computational Linguistics: ACL 2024},
  pages={467--484},
  year={2024}
}

@article{wanefficient2023,
    author = {Wan, Zhongwei and Wang, Xin and Liu, Che and Alam, Samiul and Zheng, Yu and Liu, Jiachen and Qu, Zhongnan and Yan, Shen and Zhu, Yi and Zhang, Quanlu and others},
    journal = {Transactions on Machine Learning Research},
    title = {Efficient Large Language Models: A Survey},
    year = {2023}
}

@inproceedings{dhar2024empirical,
    author = {Dhar, Nobel and Deng, Bobin and Lo, Dan and Wu, Xiaofeng and Zhao, Liang and Suo, Kun},
    booktitle = {Proceedings of the 2024 ACM Southeast Conference},
    title = {An empirical analysis and resource footprint study of deploying large language models on edge devices},
    year = {2024}
}

@inproceedings{muennighoff2023mteb,
    author = {Muennighoff, Niklas  and
Tazi, Nouamane  and
Magne, Loic  and
Reimers, Nils},
    booktitle = {Proceedings of the 17th Conference of the European Chapter of the Association for Computational Linguistics},
    title = {{MTEB}: Massive Text Embedding Benchmark},
    url = {https://aclanthology.org/2023.eacl-main.148},
    year = {2023}
}

@article{abdin2024phi,
    author = {Abdin, Marah and Jacobs, Sam Ade and Awan, Ammar Ahmad and Aneja, Jyoti and Awadallah, Ahmed and Awadalla, Hany and Bach, Nguyen and Bahree, Amit and Bakhtiari, Arash and Behl, Harkirat and others},
    journal = {arXiv preprint arXiv:2404.14219},
    title = {Phi-3 technical report: A highly capable language model locally on your phone},
    url = {https://arxiv.org/abs/2404.14219},
year = {2024}
}

@article{team2024gemma,
    author = {{Gemma Team} and Riviere, Morgane and Pathak, Shreya and Sessa, Pier Giuseppe and Hardin, Cassidy and Bhupatiraju, Surya and Hussenot, L{\'e}onard and Mesnard, Thomas and Shahriari, Bobak and Ram{\'e}, Alexandre and others},
    journal = {arXiv preprint arXiv:2408.00118},
    title = {Gemma 2: Improving open language models at a practical size},
    url = {https://arxiv.org/abs/2408.00118},
year = {2024}
}

@inproceedings{reimers2019sentence,
    author = {Reimers, Nils  and
Gurevych, Iryna},
    booktitle = {Proc. of EMNLP},
    title = {Sentence-{BERT}: Sentence Embeddings using {S}iamese {BERT}-Networks},
    url = {https://aclanthology.org/D19-1410},
    year = {2019}
}

@inproceedings{
hu2024minicpm,
title={Mini{CPM}: Unveiling the Potential of Small Language Models with Scalable Training Strategies},
author={Shengding Hu and Yuge Tu and Xu Han and Ganqu Cui and Chaoqun He and Weilin Zhao and Xiang Long and Zhi Zheng and Yewei Fang and Yuxiang Huang and Xinrong Zhang and Zhen Leng Thai and Chongyi Wang and Yuan Yao and Chenyang Zhao and Jie Zhou and Jie Cai and Zhongwu Zhai and Ning Ding and Chao Jia and Guoyang Zeng and dahai li and Zhiyuan Liu and Maosong Sun},
booktitle={First Conference on Language Modeling},
year={2024},
url={https://openreview.net/forum?id=3X2L2TFr0f}
}

@article{juan2024fine,
    author = {Bucher, Martin Juan José and Martini, Marco},
    journal = {arXiv preprint arXiv:2406.08660},
    title = {Fine-Tuned'Small'LLMs (Still) Significantly Outperform Zero-Shot Generative AI Models in Text Classification},
    year = {2024},
}

@inproceedings{chen2024learning,
  title={Learning High-Quality and General-Purpose Phrase Representations},
  author={Chen, Lihu and Varoquaux, Ga{\"e}l and Suchanek, Fabian M},
  booktitle={EACL 2024-The 18th Conference of the European Chapter of the Association for Computational Linguistics},
  year={2024}
}

@inproceedings{chen2021lightweight,
  title={A lightweight neural model for biomedical entity linking},
  author={Chen, Lihu and Varoquaux, Ga{\"e}l and Suchanek, Fabian M},
  booktitle={Proceedings of the AAAI conference on artificial intelligence},
  volume={35},
  number={14},
  pages={12657--12665},
  year={2021}
}

@article{grinsztajn2022tree,
    author = {Grinsztajn, L{\'e}o and Oyallon, Edouard and Varoquaux, Ga{\"e}l},
    journal = {Advances in neural information processing systems},
    title = {Why do tree-based models still outperform deep learning on typical tabular data?},
    year = {2022}
}

@inproceedings{gilpin2018explaining,
    author = {Gilpin, Leilani H and Bau, David and Yuan, Ben Z and Bajwa, Ayesha and Specter, Michael and Kagal, Lalana},
    booktitle = {2018 IEEE 5th International Conference on data science and advanced analytics (DSAA)},
    organization = {IEEE},
    title = {Explaining explanations: An overview of interpretability of machine learning},
    year = {2018}
}

@article{lipton2018mythos,
    author = {Lipton, Zachary C},
    journal = {Queue},
    number = {3},
    title = {The mythos of model interpretability: In machine learning, the concept of interpretability is both important and slippery.},
    year = {2018}
}

@inproceedings{barcelo2020model,
    author = {Pablo Barcel{\'{o}} and
Mika{\"{e}}l Monet and
Jorge P{\'{e}}rez and
Bernardo Subercaseaux},
    booktitle = {Advances in Neural Information Processing Systems 33: Annual Conference
on Neural Information Processing Systems 2020, NeurIPS 2020, December
6-12, 2020, virtual},
    editor = {Hugo Larochelle and
Marc'Aurelio Ranzato and
Raia Hadsell and
Maria{-}Florina Balcan and
Hsuan{-}Tien Lin},
    title = {Model Interpretability through the lens of Computational Complexity},
    url = {https://proceedings.neurips.cc/paper/2020/hash/b1adda14824f50ef24ff1c05bb66faf3-Abstract.html},
    year = {2020}
}

@article{gosiewska2021simpler,
    author = {Gosiewska, Alicja and Kozak, Anna and Biecek, Przemys{\l}aw},
    journal = {Decision Support Systems},
    title = {Simpler is better: Lifting interpretability-performance trade-off via automated feature engineering},
    year = {2021}
}

@inproceedings{caruana2015intelligible,
    author = {Rich Caruana and
Yin Lou and
Johannes Gehrke and
Paul Koch and
Marc Sturm and
Noemie Elhadad},
    booktitle = {Proceedings of the 21th {ACM} {SIGKDD} International Conference on
Knowledge Discovery and Data Mining, Sydney, NSW, Australia, August
10-13, 2015},
    editor = {Longbing Cao and
Chengqi Zhang and
Thorsten Joachims and
Geoffrey I. Webb and
Dragos D. Margineantu and
Graham Williams},
    title = {Intelligible Models for HealthCare: Predicting Pneumonia Risk and
Hospital 30-day Readmission},
    url = {https://doi.org/10.1145/2783258.2788613},
    year = {2015}
}

@inproceedings{kurshan2021current,
    author = {Kurshan, Eren and Chen, Jiahao and Storchan, Victor and Shen, Hongda},
    booktitle = {Proceedings of the second ACM international conference on AI in finance},
    title = {On the current and emerging challenges of developing fair and ethical AI solutions in financial services},
    year = {2021}
}

@article{eliot2021need,
    author = {Eliot, Dr Lance B},
    journal = {Available at SSRN 3975778},
    title = {The need for explainable AI (XAI) is especially crucial in the law},
    year = {2021}
}

@inproceedings{devlin2019bert,
    author = {Devlin, Jacob  and
Chang, Ming-Wei  and
Lee, Kenton  and
Toutanova, Kristina},
    booktitle = {Proc. of NAACL-HLT},
    title = {{BERT}: Pre-training of Deep Bidirectional Transformers for Language Understanding},
    url = {https://aclanthology.org/N19-1423},
    year = {2019}
}

@inproceedings{wang2018glue,
    author = {Alex Wang and
Amanpreet Singh and
Julian Michael and
Felix Hill and
Omer Levy and
Samuel R. Bowman},
    booktitle = {Proc. of ICLR},
    title = {{GLUE:} {A} Multi-Task Benchmark and Analysis Platform for Natural
Language Understanding},
    url = {https://openreview.net/forum?id=rJ4km2R5t7},
    year = {2019}
}

@article{he2023survey,
    author = {He, Kai and Mao, Rui and Lin, Qika and Ruan, Yucheng and Lan, Xiang and Feng, Mengling and Cambria, Erik},
    journal = {arXiv preprint arXiv:2310.05694},
    title = {A survey of large language models for healthcare: from data, technology, and applications to accountability and ethics},
    url = {https://arxiv.org/abs/2310.05694},
year = {2023}
}

@article{sun2023short,
    author = {Sun, Zhongxiang},
    journal = {arXiv preprint arXiv:2303.09136},
    title = {A short survey of viewing large language models in legal aspect},
    url = {https://arxiv.org/abs/2303.09136},
year = {2023}
}

@article{jiang2024survey,
  title={A survey on large language models for code generation},
  author={Jiang, Juyong and Wang, Fan and Shen, Jiasi and Kim, Sungju and Kim, Sunghun},
  journal={ACM Transactions on Software Engineering and Methodology},
  volume={35},
  number={2},
  pages={1--72},
  year={2026},
  publisher={ACM New York, NY}
}

@article{dubey2024llama,
    author = {Dubey, Abhimanyu and Jauhri, Abhinav and Pandey, Abhinav and Kadian, Abhishek and Al-Dahle, Ahmad and Letman, Aiesha and Mathur, Akhil and Schelten, Alan and Yang, Amy and Fan, Angela and others},
    journal = {arXiv preprint arXiv:2407.21783},
    title = {The llama 3 herd of models},
    url = {https://arxiv.org/abs/2407.21783},
year = {2024}
}

@inproceedings{varoquaux2024hype,
  title={Hype, Sustainability, and the Price of the Bigger-is-Better Paradigm in AI},
  author={Varoquaux, Ga{\"e}l and Luccioni, Alexandra Sasha and Whittaker, Meredith},
  booktitle={FAccT 2025-ACM Conference on Fairness, Accountability, and Transparency},
  year={2025}
}

@inproceedings{warstadt2023findings,
  title={Findings of the BabyLM Challenge: Sample-Efficient Pretraining on Developmentally Plausible Corpora},
  author={Warstadt, Alex and Mueller, Aaron and Choshen, Leshem and Wilcox, Ethan and Zhuang, Chengxu and Ciro, Juan and Mosquera, Rafael and Paranjabe, Bhargavi and Williams, Adina and Linzen, Tal and others},
  booktitle={Proceedings of the BabyLM Challenge at the 27th Conference on Computational Natural Language Learning},
  pages={1--34},
  year={2023}
}

@article{wang2024comprehensive,
  title={A comprehensive survey of small language models in the era of large language models: Techniques, enhancements, applications, collaboration with llms, and trustworthiness},
  author={Wang, Fali and Zhang, Zhiwei and Zhang, Xianren and Wu, Zongyu and Mo, Tzuhao and Lu, Qiuhao and Wang, Wanjing and Li, Rui and Xu, Junjie and Tang, Xianfeng and others},
  journal = {arXiv preprint arXiv:2411.03350},
  year={2024},
}

@article{van2024survey,
  title={A survey of small language models},
  author={Van Nguyen, Chien and Shen, Xuan and Aponte, Ryan and Xia, Yu and Basu, Samyadeep and Hu, Zhengmian and Chen, Jian and Parmar, Mihir and Kunapuli, Sasidhar and Barrow, Joe and others},
  journal = {arXiv preprint arXiv:2410.20011},
  year={2024},
}

@article{subramanian2025small,
  title={Small language models (slms) can still pack a punch: A survey},
  author={Subramanian, Shreyas and Elango, Vikram and Gungor, Mecit},
  journal = {arXiv preprint arXiv:2501.05465},
  year={2025},
}

@article{openai2023gpt4,
  title        = {GPT-4 Technical Report},
  author       = {{OpenAI}},
  journal = {arXiv preprint arXiv:2303.08774},
  year         = {2023},
  archivePrefix = {arXiv},
doi          = {10.48550/ARXIV.2303.08774},
  url          = {https://arxiv.org/abs/2303.08774}
}

@article{deepseek2025r1,
  title={DeepSeek-R1 incentivizes reasoning in LLMs through reinforcement learning},
  author={DeepSeek-AI and Daya Guo and Dejian Yang and Haowei Zhang and Jun-Mei Song and Ruoyu Zhang and Runxin Xu and Qihao Zhu and Shirong Ma and Peiyi Wang and Xiaoling Bi and Xiaokang Zhang and Xingkai Yu and Yu Wu and Z. F. Wu and Zhibin Gou and Zhihong Shao and Zhuoshu Li and Ziyi Gao and Aixin Liu and Bing Xue and Bing-Li Wang and Bochao Wu and Bei Feng and Chengda Lu and Chenggang Zhao and Chengqi Deng and Chenyu Zhang and Chong Ruan and Damai Dai and Deli Chen and Dong-Li Ji and Erhang Li and Fangyun Lin and Fucong Dai and Fuli Luo and Guangbo Hao and Guanting Chen and Guowei Li and H. Zhang and Han Bao and Hanwei Xu and Haocheng Wang and Honghui Ding and Huajian Xin and Huazuo Gao and Hui Qu and Hui Li and Jianzhong Guo and Jiashi Li and Jiawei Wang and JingChang Chen and Jingyang Yuan and Junjie Qiu and Jun-Long Li and J. L. Cai and Jiaqi Ni and Jian Liang and Jin Chen and Kai Dong and Kai Hu and Kaige Gao and Kang Guan and Kexin Huang and Kuai Yu and Lean Wang and Lecong Zhang and Liang Zhao and Litong Wang and Liyue Zhang and Lei Xu and Leyi Xia and Mingchuan Zhang and Minghua Zhang and M. Tang and Meng Li and Miaojun Wang and Mingming Li and Ning Tian and Panpan Huang and Peng Zhang and Qiancheng Wang and Qinyu Chen and Qiushi Du and Ruiqi Ge and Ruisong Zhang and Ruizhe Pan and Runji Wang and R. J. Chen and Rui-Qi Jin and Ruyi Chen and Shanghao Lu and Shangyan Zhou and Shanhuang Chen and Shengfeng Ye and Shiyu Wang and Shuiping Yu and Shunfeng Zhou and Shuting Pan and S. S. Li and Shuang Zhou and Shao-Kang Wu and Tao Yun and Tian Pei and Tianyu Sun and T. Wang and Wangding Zeng and Wanjia Zhao and Wen Liu and Wenfeng Liang and Wenjun Gao and Wen-Xia Yu and Wentao Zhang and Wangding Xiao and Wei An and Xiaodong Liu and Xiaohan Wang and Xiaokang Chen and Xiaotao Nie and Xin Cheng and Xin Liu and Xin Xie and Xingchao Liu and Xin-Yu Yang and Xinyuan Li and Xuecheng Su and Xuheng Lin and X. Q. Li and Xiangyu Jin and Xi-Cheng Shen and Xiaosha Chen and Xiaowen Sun and Xiaoxiang Wang and Xinnan Song and Xinyi Zhou and Xianzu Wang and Xinxia Shan and Y. K. Li and Y. Q. Wang and Y. X. Wei and Yang Zhang and Yanhong Xu and Yao Li and Yao Zhao and Yaofeng Sun and Yaohui Wang and Yi Yu and Yichao Zhang and Yifan Shi and Yi Xiong and Ying He and Yishi Piao and Yisong Wang and Yixuan Tan and Yiyang Ma and Yiyuan Liu and Yongqiang Guo and Yuan Ou and Yuduan Wang and Yue Gong and Yu-Jing Zou and Yujia He and Yunfan Xiong and Yu-Wei Luo and Yu-mei You and Yuxuan Liu and Yuyang Zhou and Y. X. Zhu and Yanping Huang and Yao Li and Yi Zheng and Yuchen Zhu and Yunxiang Ma and Ying Tang and Yukun Zha and Yuting Yan and Zehui Ren and Zehui Ren and Zhangli Sha and Zhe Fu and Zhean Xu and Zhenda Xie and Zhen-guo Zhang and Zhewen Hao and Zhicheng Ma and Zhigang Yan and Zhiyu Wu and Zihui Gu and Zijia Zhu and Zijun Liu and Zi-Long Li and Ziwei Xie and Ziyang Song and Zizheng Pan and Zhen Huang and Zhipeng Xu and Zhongyu Zhang and Zhen Zhang},
  journal={Nature},
  year={2025},
  volume={645},
  pages={633 - 638},
  url={https://api.semanticscholar.org/CorpusID:275789950}
}

@inproceedings{subramani-etal-2023-detecting,
  title = {Detecting Personal Information in Training Corpora: an Analysis},
  author = {Subramani, Nishant and Luccioni, Sasha and Dodge, Jesse and Mitchell, Margaret},
  booktitle = {Proceedings of the 3rd Workshop on Trustworthy Natural Language Processing (TrustNLP 2023)},
  editor = {Ovalle, Anaelia and Chang, Kai-Wei and Mehrabi, Ninareh and Pruksachatkun, Yada and Galystan, Aram and Dhamala, Jwala and Verma, Apurv and Cao, Trista and Kumar, Anoop and Gupta, Rahul},
  month = jul,
  year = {2023},
  address = {Toronto, Canada},
  publisher = {Association for Computational Linguistics},
  pages = {208--220},
  url = {https://aclanthology.org/2023.trustnlp-1.18/},
  doi = {10.18653/v1/2023.trustnlp-1.18}
}

@article{
yu2023selective,
title={Selective Pre-training for Private Fine-tuning},
author={Da Yu and Sivakanth Gopi and Janardhan Kulkarni and Zinan Lin and Saurabh Naik and Tomasz Lukasz Religa and Jian Yin and Huishuai Zhang},
journal={Transactions on Machine Learning Research},
issn={2835-8856},
year={2024},
url={https://openreview.net/forum?id=y3u8OpPHxz},
}

@article{Arnett2024ToxicityOfTheCommons,
  title        = {Toxicity of the Commons: Curating Open-Source Pre-Training Data},
  author       = {Arnett, Catherine and Jones, Eliot and Yamshchikov, Ivan P. and Langlais, Pierre-Carl},
  journal = {arXiv preprint arXiv:2410.22587},
  year         = {2024},
  url          = {https://arxiv.org/abs/2410.22587},
}

@article{Kamphuis2024TinyToxicDetector,
  title        = {Tiny-Toxic-Detector: A compact transformer-based model for toxic content detection},
  author       = {Kamphuis, Michiel},
  journal = {arXiv preprint arXiv:2409.02114},
  year         = {2024},
  url          = {https://arxiv.org/abs/2409.02114},
}

@article{Kang2024AutoScale,
  title        = {AutoScale: Automatic Prediction of Compute-optimal Data Composition for Training LLMs},
  author       = {Kang, Feiyang and Sun, Yifan and Wen, Bingbing and Chen, Si and Song, Dawn and Mahmood, Rafid and Jia, Ruoxi},
  journal = {arXiv preprint arXiv:2407.20177},
  year         = {2024},
  url          = {https://arxiv.org/abs/2407.20177}
}

@article{Yang2025DataMixingAgent,
  title        = {Data Mixing Agent: Learning to Re-weight Domains for Continual Pre-training},
  author       = {Yang, Kailai and Liu, Xiao and Ji, Lei and Li, Hao and Gong, Yeyun and Cheng, Peng and Yang, Mao},
  journal = {arXiv preprint arXiv:2507.15640},
  year         = {2025},
  url          = {https://arxiv.org/abs/2507.15640},
}

@inproceedings{thakkar2023self,
    title = "Self-Influence Guided Data Reweighting for Language Model Pre-training",
    author = "Thakkar, Megh  and
      Bolukbasi, Tolga  and
      Ganapathy, Sriram  and
      Vashishth, Shikhar  and
      Chandar, Sarath  and
      Talukdar, Partha",
    editor = "Bouamor, Houda  and
      Pino, Juan  and
      Bali, Kalika",
    booktitle = "Proceedings of the 2023 Conference on Empirical Methods in Natural Language Processing",
    month = dec,
    year = "2023",
    address = "Singapore",
    publisher = "Association for Computational Linguistics",
    url = "https://aclanthology.org/2023.emnlp-main.125/",
    doi = "10.18653/v1/2023.emnlp-main.125",
    pages = "2033--2045"
}

@inproceedings{liu2025rethinking,
  title        = {Take the Essence and Discard the Dross: A Rethinking on Data Selection for Fine-Tuning Large Language Models},
  author       = {Ziche Liu and Rui Ke and Yajiao Liu and Feng Jiang and Haizhou Li},
  booktitle    = {Proceedings of the 2025 Conference of the Nations of the Americas Chapter of the Association for Computational Linguistics: Human Language Technologies (Volume 1: Long Papers)},
  pages        = {6595--6611},
  year         = {2025},
  address      = {Albuquerque, New Mexico},
  publisher    = {Association for Computational Linguistics},
  doi          = {10.18653/v1/2025.naacl-long.336}
}

@inproceedings{wang2025datawhisperer,
  title        = {Data Whisperer: Efficient Data Selection for Task-Specific LLM Fine-Tuning via Few-Shot In-Context Learning},
  author       = {Shaobo Wang and Xiangqi Jin and Ziming Wang and Jize Wang and Jiajun Zhang and Kaixin Li and Zichen Wen and Zhong Li and Conghui He and Xuming Hu and Linfeng Zhang},
  booktitle    = {Proceedings of the 63rd Annual Meeting of the Association for Computational Linguistics (Volume 1: Long Papers)},
  pages        = {23287--23305},
  year         = {2025},
  address      = {Vienna, Austria},
  publisher    = {Association for Computational Linguistics},
  doi          = {10.18653/v1/2025.acl-long.1135},
  url          = {https://aclanthology.org/2025.acl-long.1135/}
}

@inproceedings{gao2025principled,
  title={Principled Data Selection for Alignment: The Hidden Risks of Difficult Examples},
  author={Gao, Chengqian and Li, Haonan and Liu, Liu and Xie, Zeke and Zhao, Peilin and Xu, Zhiqiang},
  booktitle={International Conference on Machine Learning},
  pages={18386--18409},
  year={2025},
  organization={PMLR}
}

@inproceedings{
Chen2025QueryLevelUncertainty,
title={Query-Level Uncertainty in Large Language Models},
author={Lihu Chen and Gerard de Melo and Fabian M. Suchanek and Ga{\"e}l Varoquaux},
booktitle={The Fourteenth International Conference on Learning Representations},
year={2026},
url={https://openreview.net/forum?id=11QZITAMUO}
}

@inproceedings{EnomotoEda2021LearningToCascade,
  author       = {Enomoto, Shohei and Eda, Takeharu},
  title        = {Learning to Cascade: Confidence Calibration for Improving the Accuracy and Computational Cost of Cascade Inference Systems},
  booktitle    = {Proceedings of the Thirty-Fifth AAAI Conference on Artificial Intelligence (AAAI-21)},
  year         = {2021},
  url          = {https://arxiv.org/abs/2104.09286},
}

@inproceedings{wang2017idk,
  author       = {Xin Wang and
                  Yujia Luo and
                  Daniel Crankshaw and
                  Alexey Tumanov and
                  Fisher Yu and
                  Joseph E. Gonzalez},
  editor       = {Amir Globerson and
                  Ricardo Silva},
  title        = {{IDK} Cascades: Fast Deep Learning by Learning not to Overthink},
  booktitle    = {Proceedings of the Thirty-Fourth Conference on Uncertainty in Artificial
                  Intelligence, {UAI} 2018, Monterey, California, USA, August 6-10,
                  2018},
  pages        = {580--590},
  publisher    = {{AUAI} Press},
  year         = {2018},
  url          = {http://auai.org/uai2018/proceedings/papers/212.pdf},
  bibsource    = {dblp computer science bibliography, https://dblp.org}
}

@article{Wang2024CascadeAwareTraining,
  author       = {Wang, Congchao and Augenstein, Sean and Rush, Keith and Jitkrittum, Wittawat and Narasimhan, Harikrishna and Rawat, Ankit Singh and Menon, Aditya Krishna and Go, Alec},
  title        = {Cascade-Aware Training of Language Models},
  journal = {arXiv preprint arXiv:2406.00060},
  year         = {2024},
  url          = {https://arxiv.org/abs/2406.00060},
}

@inproceedings{Zhang2024EfficientContextualLLMCascades,
  author       = {Zhang, Xuechen and Huang, Zijian and Taga, Ege Onur and Joe-Wong, Carlee and Oymak, Samet and Chen, Jiasi},
  title        = {Efficient Contextual LLM Cascades through Budget-Constrained Policy Learning},
  booktitle    = {Advances in Neural Information Processing Systems (NeurIPS 2024)},
  year         = {2024},
  url          = {https://arxiv.org/abs/2404.13082},
}

@inproceedings{Su2025CPRouter,
  title={Cp-router: An uncertainty-aware router between llm and lrm},
  author={Su, Jiayuan and Lin, Fulin and Feng, Zhaopeng and Zheng, Han and Wang, Teng and Xiao, Zhenyu and Zhao, Xinlong and Liu, Zuozhu and Cheng, Lu and Wang, Hongwei},
  booktitle={Proceedings of the AAAI Conference on Artificial Intelligence},
  volume={40},
  number={39},
  pages={33065--33073},
  year={2026}
}

@inproceedings{
guha2024smoothie,
title={Smoothie: Label Free Language Model Routing},
author={Neel Guha and Mayee F Chen and Trevor Chow and Ishan S. Khare and Christopher Re},
booktitle={The Thirty-eighth Annual Conference on Neural Information Processing Systems},
year={2024},
url={https://openreview.net/forum?id=pPSWHsgqRp}
}

@inproceedings{Chen2025TagRouter,
  title        = {TagRouter: Learning Route to LLMs through Tags for Open-Domain Text Generation Tasks},
  author       = {Chen, Zhou and Wei, Zhiqiang and Bai, Yuqi and Xiong, Xue and Wu, Jianmin},
  booktitle    = {Findings of ACL 2025},
  year         = {2025}
}

@misc{Remy2024TransTokenization,
  title        = {Trans-Tokenization and Cross-lingual Vocabulary Transfers: Language Adaptation of LLMs for Low-Resource NLP},
  author       = {Remy, François and Delobelle, Pieter and Avetisyan, Hayastan and Khabibullina, Alfiya and de Lhoneux, Miryam and Demeester, Thomas},
  year         = {2024},
  howpublished = {In “First Conference on Language Modeling (COLM) 2024”, arXiv pre-print arXiv:2408.04303},
archivePrefix= {arXiv},
  primaryClass = {cs.CL},
  url          = {https://arxiv.org/abs/2408.04303}
}

@article{huang2025survey,
  title={A survey on hallucination in large language models: Principles, taxonomy, challenges, and open questions},
  author={Huang, Lei and Yu, Weijiang and Ma, Weitao and Zhong, Weihong and Feng, Zhangyin and Wang, Haotian and Chen, Qianglong and Peng, Weihua and Feng, Xiaocheng and Qin, Bing and others},
  journal={ACM Transactions on Information Systems},
  volume={43},
  number={2},
  pages={1--55},
  year={2025},
  publisher={ACM New York, NY}
}

@article{Kasai2022RealTimeQA,
  title={Realtime qa: What's the answer right now?},
  author={Kasai, Jungo and Sakaguchi, Keisuke and Le Bras, Ronan and Asai, Akari and Yu, Xinyan and Radev, Dragomir and Smith, Noah A and Choi, Yejin and Inui, Kentaro and others},
  journal={Advances in neural information processing systems},
  volume={36},
  pages={49025--49043},
  year={2023}
}

@misc{Feng2023Trends,
  title        = {Trends in Integration of Knowledge and Large Language Models: A Survey and Taxonomy of Methods, Benchmarks, and Applications},
  author       = {Feng, Zhangyin and Ma, Weitao and Yu, Weijiang and Huang, Lei and Wang, Haotian and Chen, Qianglong and Peng, Weihua and Feng, Xiaocheng and Qin, Bing and Liu, Ting},
  year         = {2023},
archivePrefix= {arXiv},
  primaryClass = {cs.CL},
  url          = {https://arxiv.org/abs/2311.05876}
}

@inproceedings{
Mondorf2024BeyondAccuracy,
title={Beyond Accuracy: Evaluating the Reasoning Behavior of Large Language Models - A Survey},
author={Philipp Mondorf and Barbara Plank},
booktitle={First Conference on Language Modeling},
year={2024},
url={https://openreview.net/forum?id=Lmjgl2n11u}
}

@inproceedings{li-etal-2025-knowledge-boundary,
    title = "Knowledge Boundary of Large Language Models: A Survey",
    author = "Li, Moxin  and
      Zhao, Yong  and
      Zhang, Wenxuan  and
      Li, Shuaiyi  and
      Xie, Wenya  and
      Ng, See-Kiong  and
      Chua, Tat-Seng  and
      Deng, Yang",
    editor = "Che, Wanxiang  and
      Nabende, Joyce  and
      Shutova, Ekaterina  and
      Pilehvar, Mohammad Taher",
    booktitle = "Proceedings of the 63rd Annual Meeting of the Association for Computational Linguistics (Volume 1: Long Papers)",
    month = jul,
    year = "2025",
    address = "Vienna, Austria",
    publisher = "Association for Computational Linguistics",
    url = "https://aclanthology.org/2025.acl-long.256/",
    doi = "10.18653/v1/2025.acl-long.256",
    pages = "5131--5157",
    ISBN = "979-8-89176-251-0",
    
}

@article{Ling2023DomainSpecializationLLM,
author = {Ling, Chen and Zhao, Xujiang and Lu, Jiaying and Deng, Chengyuan and Zheng, Can and Wang, Junxiang and Chowdhury, Tanmoy and Li, Yun and Cui, Hejie and Zhang, Xuchao and Zhao, Tianjiao and Panalkar, Amit and Mehta, Dhagash and Pasquali, Stefano and Cheng, Wei and Wang, Haoyu and Liu, Yanchi and Chen, Zhengzhang and Chen, Haifeng and White, Chris and Gu, Quanquan and Pei, Jian and Yang, Carl and Zhao, Liang},
title = {Domain Specialization as the Key to Make Large Language Models Disruptive: A Comprehensive Survey},
year = {2025},
issue_date = {February 2026},
publisher = {Association for Computing Machinery},
address = {New York, NY, USA},
volume = {58},
number = {3},
issn = {0360-0300},
url = {https://doi.org/10.1145/3764579},
doi = {10.1145/3764579},
journal = {ACM Comput. Surv.},
month = oct,
articleno = {79},
numpages = {39}
}

@inproceedings{Shi2024MedAdapter,
  author       = {Shi, Wenqi and Xu, Ran and Zhuang, Yuchen and Yu, Yue and Sun, Haotian and Wu, Hang and Yang, Carl and Wang, May D.},
  title        = {MedAdapter: Efficient Test‐Time Adaptation of Large Language Models Towards Medical Reasoning},
  booktitle    = {Proceedings of the 2024 Conference on Empirical Methods in Natural Language Processing (EMNLP 2024)},
  pages        = {22294--22314},
  year         = {2024},
  address      = {Miami, Florida, USA},
  publisher    = {Association for Computational Linguistics},
  doi          = {10.18653/v1/2024.emnlp-main.1244},
  url          = {https://aclanthology.org/2024.emnlp-main.1244/}
}

@inproceedings{SealsShalin2024Deductive,
  author       = {Seals, Spencer M. and Shalin, Valerie L.},
  title        = {Evaluating the Deductive Competence of Large Language Models},
  booktitle    = {Proceedings of the 2024 Conference of the North American Chapter of the Association for Computational Linguistics: Human Language Technologies (Volume 1: Long Papers)},
  editor       = {Duh, Kevin and Gomez, Helena and Bethard, Steven},
  month        = jun,
  year         = {2024},
  address      = {Mexico City, Mexico},
  publisher    = {Association for Computational Linguistics},
  pages        = {8614--8630},
  doi          = {10.18653/v1/2024.naacl-long.476},
  url          = {https://aclanthology.org/2024.naacl-long.476/}
}

@article{
Sinha2024Compositional,
title={A Survey on Compositional Learning of {AI} Models: Theoretical and Experimental Practices},
author={Sania Sinha and Tanawan Premsri and Parisa Kordjamshidi},
journal={Transactions on Machine Learning Research},
issn={2835-8856},
year={2024},
url={https://openreview.net/forum?id=BXDxwItNqQ}
}

@article{Weller2025Theoretical,
  author       = {Weller, Orion and Boratko, Michael and Naim, Iftekhar and Lee, Jinhyuk},
  title        = {On the Theoretical Limitations of Embedding‐Based Retrieval},
  journal = {arXiv preprint arXiv:2508.21038},
  year         = {2025},
  url          = {https://arxiv.org/abs/2508.21038},
}

@inproceedings{lang2025debate,
  title={Debate helps weak-to-strong generalization},
  author={Lang, Hao and Huang, Fei and Li, Yongbin},
  booktitle={Proceedings of the AAAI Conference on Artificial Intelligence},
  volume={39},
  number={26},
  pages={27410--27418},
  year={2025}
}

@article{gu2024survey,
  title={A survey on llm-as-a-judge},
  author={Gu, Jiawei and Jiang, Xuhui and Shi, Zhichao and Tan, Hexiang and Zhai, Xuehao and Xu, Chengjin and Li, Wei and Shen, Yinghan and Ma, Shengjie and Liu, Honghao and others},
  journal = {arXiv preprint arXiv:2411.15594},
  year={2024},
}

@inproceedings{thakur2024judging,
  title={Judging the judges: Evaluating alignment and vulnerabilities in llms-as-judges},
  author={Thakur, Aman Singh and Choudhary, Kartik and Ramayapally, Venkat Srinik and Vaidyanathan, Sankaran and Hupkes, Dieuwke},
  booktitle={Proceedings of the Fourth Workshop on Generation, Evaluation and Metrics (GEM$^2$)},
  pages={404--430},
  year={2025}
}

@article{li2024llms,
  title={Llms-as-judges: a comprehensive survey on llm-based evaluation methods},
  author={Li, Haitao and Dong, Qian and Chen, Junjie and Su, Huixue and Zhou, Yujia and Ai, Qingyao and Ye, Ziyi and Liu, Yiqun},
  journal = {arXiv preprint arXiv:2412.05579},
  year={2024},
}

@inproceedings{
ye2024justice,
title={Justice or Prejudice? Quantifying Biases in {LLM}-as-a-Judge},
author={Jiayi Ye and Yanbo Wang and Yue Huang and Dongping Chen and Qihui Zhang and Nuno Moniz and Tian Gao and Werner Geyer and Chao Huang and Pin-Yu Chen and Nitesh V Chawla and Xiangliang Zhang},
booktitle={The Thirteenth International Conference on Learning Representations},
year={2025},
url={https://openreview.net/forum?id=3GTtZFiajM}
}

@inproceedings{yang2023knowledge,
  title={From knowledge distillation to self-knowledge distillation: A unified approach with normalized loss and customized soft labels},
  author={Yang, Zhendong and Zeng, Ailing and Li, Zhe and Zhang, Tianke and Yuan, Chun and Li, Yu},
  booktitle={Proceedings of the IEEE/CVF International Conference on Computer Vision},
  pages={17185--17194},
  year={2023}
}

@article{naduaș2025synthetic,
  title={Synthetic data generation using large language models: Advances in text and code},
  author={Nadăș, Mihai and Dioșan, Laura and Tomescu, Andreea},
  journal={IEEE Access},
  year={2025},
  publisher={IEEE}
}

@inproceedings{lu2025demystifying,
  title={Demystifying small language models for edge deployment},
  author={Lu, Zhenyan and Li, Xiang and Cai, Dongqi and Yi, Rongjie and Liu, Fangming and Liu, Wei and Luan, Jian and Zhang, Xiwen and Lane, Nicholas D and Xu, Mengwei},
  booktitle={Proceedings of the 63rd Annual Meeting of the Association for Computational Linguistics (Volume 1: Long Papers)},
  pages={14747--14764},
  year={2025}
}

@inproceedings{jang2025edge,
  title={Edge-first language model inference: Models, metrics, and tradeoffs},
  author={Jang, SiYoung and Morabito, Roberto},
  booktitle={2025 IEEE 45th International Conference on Distributed Computing Systems Workshops (ICDCSW)},
  pages={309--314},
  year={2025},
  organization={IEEE}
}

@inproceedings{samarinas2025distillation,
  title={Distillation and refinement of reasoning in small language models for document re-ranking},
  author={Samarinas, Chris and Zamani, Hamed},
  booktitle={Proceedings of the 2025 International ACM SIGIR Conference on Innovative Concepts and Theories in Information Retrieval (ICTIR)},
  pages={430--435},
  year={2025}
}

@inproceedings{yu2025integrating,
  title={Integrating small language models with retrieval-augmented generation in computing education: Key takeaways, setup, and practical insights},
  author={Yu, Zezhu and Liu, Suqing and Denny, Paul and Bergen, Andreas and Liut, Michael},
  booktitle={Proceedings of the 56th ACM Technical Symposium on Computer Science Education V. 1},
  pages={1302--1308},
  year={2025}
}

@misc{cisco2025agentic,
  author       = {Cisco Systems, Inc.},
  title        = {Agentic AI Poised to Handle 68\% of Customer Service and Support Interactions by 2028},
  howpublished = {Press Release, Cisco Newsroom},
  month        = may,
  year         = {2025},
  url          = {https://newsroom.cisco.com/c/r/newsroom/en/us/a/y2025/m05/agentic-ai-poised-to-handle-68-of-customer-service-and-support-interactions-by-2028.html}
}

@article{belcak2025small,
  title={Small Language Models are the Future of Agentic AI},
  author={Belcak, Peter and Heinrich, Greg and Diao, Shizhe and Fu, Yonggan and Dong, Xin and Muralidharan, Saurav and Lin, Yingyan Celine and Molchanov, Pavlo},
  journal = {arXiv preprint arXiv:2506.02153},
  year={2025},
}

@inproceedings{
penedo2024fineweb,
title={The FineWeb Datasets: Decanting the Web for the Finest Text Data at Scale},
author={Guilherme Penedo and Hynek Kydl{\'\i}{\v{c}}ek and Loubna Ben allal and Anton Lozhkov and Margaret Mitchell and Colin Raffel and Leandro Von Werra and Thomas Wolf},
booktitle={The Thirty-eight Conference on Neural Information Processing Systems Datasets and Benchmarks Track},
year={2024},
url={https://openreview.net/forum?id=n6SCkn2QaG}
}

@inproceedings{kargaran2023glotlid,
  title={GlotLID: Language Identification for Low-Resource Languages},
  author={Kargaran, Amir Hossein and Imani, Ayyoob and Yvon, Fran{\c{c}}ois and Sch{\"u}tze, Hinrich},
  booktitle={Findings of the Association for Computational Linguistics: EMNLP 2023},
  pages={6155--6218},
  year={2023}
}

@article{penedo2025fineweb2,
  title={FineWeb2: One Pipeline to Scale Them All--Adapting Pre-Training Data Processing to Every Language},
  author={Penedo, Guilherme and Kydl{\'\i}{\v{c}}ek, Hynek and Sabol{\v{c}}ec, Vinko and Messmer, Bettina and Foroutan, Negar and Kargaran, Amir Hossein and Raffel, Colin and Jaggi, Martin and Von Werra, Leandro and Wolf, Thomas},
  journal = {arXiv preprint arXiv:2506.20920},
  year={2025},
}

@inproceedings{burchell2023open,
  title={An Open Dataset and Model for Language Identification},
  author={Burchell, Laurie and Birch, Alexandra and Bogoychev, Nikolay and Heafield, Kenneth},
  booktitle={Proceedings of the 61st Annual Meeting of the Association for Computational Linguistics (Volume 2: Short Papers)},
  pages={865--879},
  year={2023}
}

@article{nllb2024scaling,
  author       = {{NLLB Team}},
  title        = {Scaling neural machine translation to 200 languages},
  journal      = {Nat.},
  volume       = {630},
  number       = {8018},
  pages        = {841--846},
  year         = {2024},
  url          = {https://doi.org/10.1038/s41586-024-07335-x},
  doi          = {10.1038/S41586-024-07335-X},
  bibsource    = {dblp computer science bibliography, https://dblp.org}
}

@article{oepen2025hplt,
  title={HPLT 3.0: Very Large-Scale Multilingual Resources for LLM and MT. Mono-and Bi-lingual Data, Multilingual Evaluation, and Pre-Trained Models},
  author={Oepen, Stephan and Arefev, Nikolay and Aulamo, Mikko and Ba{\~n}{\'o}n, Marta and Buljan, Maja and Burchell, Laurie and Charpentier, Lucas and Chen, Pinzhen and Fedorova, Mariya and de Gibert, Ona and others},
  journal = {arXiv preprint arXiv:2511.01066},
  year={2025},
}

@inproceedings{wang2024retrieval,
    title = "Retrieval-Augmented Machine Translation with Unstructured Knowledge",
    author = "Wang, Jiaan  and
      Meng, Fandong  and
      Zhang, Yingxue  and
      Zhou, Jie",
    editor = "Christodoulopoulos, Christos  and
      Chakraborty, Tanmoy  and
      Rose, Carolyn  and
      Peng, Violet",
    booktitle = "Findings of the Association for Computational Linguistics: EMNLP 2025",
    month = nov,
    year = "2025",
    address = "Suzhou, China",
    publisher = "Association for Computational Linguistics",
    url = "https://aclanthology.org/2025.findings-emnlp.313/",
    doi = "10.18653/v1/2025.findings-emnlp.313",
    pages = "5858--5871",
    ISBN = "979-8-89176-335-7"
}

@inproceedings{doshi2024pretraining,
  title={Pretraining language models using translationese},
  author={Doshi, Meet and Dabre, Raj and Bhattacharyya, Pushpak},
  booktitle={Proceedings of the 2024 Conference on Empirical Methods in Natural Language Processing},
  pages={5843--5862},
  year={2024}
}

@inproceedings{sennrich2016improving,
  title={Improving neural machine translation models with monolingual data},
  author={Sennrich, Rico and Haddow, Barry and Birch, Alexandra},
  booktitle={Proceedings of the 54th annual meeting of the association for computational linguistics (volume 1: long papers)},
  pages={86--96},
  year={2016}
}

@inproceedings{wang2025multilingual,
  title={Multilingual language model pretraining using machine-translated data},
  author={Wang, Jiayi and Lu, Yao and Weber, Maurice and Ryabinin, Max and Adelani, David Ifeoluwa and Chen, Yihong and Tang, Raphael and Stenetorp, Pontus},
  booktitle={Proceedings of the 2025 Conference on Empirical Methods in Natural Language Processing},
  pages={28075--28095},
  year={2025}
}

@article{gurgurov2025small,
  title={Small models, big impact: Efficient corpus and graph-based adaptation of small multilingual language models for low-resource languages},
  author={Gurgurov, Daniil and Vykopal, Ivan and van Genabith, Josef and Ostermann, Simon},
  journal = {arXiv preprint arXiv:2502.10140},
  year={2025},
}

@inproceedings{feng2024teaching,
  title={Teaching small language models reasoning through counterfactual distillation},
  author={Feng, Tao and Li, Yicheng and Chenglin, Li and Chen, Hao and Yu, Fei and Zhang, Yin},
  booktitle={Proceedings of the 2024 Conference on Empirical Methods in Natural Language Processing},
  pages={5831--5842},
  year={2024}
}

@inproceedings{zhu2024pad,
  title={PaD: Program-aided distillation can teach small models reasoning better than chain-of-thought fine-tuning},
  author={Zhu, Xuekai and Qi, Biqing and Zhang, Kaiyan and Long, Xinwei and Lin, Zhouhan and Zhou, Bowen},
  booktitle={Proceedings of the 2024 Conference of the North American Chapter of the Association for Computational Linguistics: Human Language Technologies (Volume 1: Long Papers)},
  pages={2571--2597},
  year={2024}
}

@article{wen2025reasoning,
  title={Reasoning Scaffolding: Distilling the Flow of Thought from LLMs},
  author={Wen, Xiangyu and Huang, Junhua and Li, Zeju and Li, Min and Zhong, Jianyuan and Xu, Zhijian and Yuan, Mingxuan and Huang, Yongxiang and Xu, Qiang},
  journal = {arXiv preprint arXiv:2509.23619},
  year={2025},
}

@inproceedings{wang2024codeclm,
  title={Codeclm: Aligning language models with tailored synthetic data},
  author={Wang, Zifeng and Li, Chun-Liang and Perot, Vincent and Le, Long and Miao, Jin and Zhang, Zizhao and Lee, Chen-Yu and Pfister, Tomas},
  booktitle={Findings of the Association for Computational Linguistics: NAACL 2024},
  pages={3712--3729},
  year={2024}
}

@inproceedings{nayak2024learning,
  title={Learning to generate instruction tuning datasets for zero-shot task adaptation},
  author={Nayak, Nihal and Nan, Yiyang and Trost, Avi and Bach, Stephen},
  booktitle={Findings of the Association for Computational Linguistics: ACL 2024},
  pages={12585--12611},
  year={2024}
}

@inproceedings{
ramasesh2021effect,
title={Effect of scale on catastrophic forgetting in neural networks},
author={Vinay Venkatesh Ramasesh and Aitor Lewkowycz and Ethan Dyer},
booktitle={International Conference on Learning Representations},
year={2022},
url={https://openreview.net/forum?id=GhVS8_yPeEa}
}

@inproceedings{chen-etal-2024-unveiling-flaws,
    title = "Unveiling the Flaws: Exploring Imperfections in Synthetic Data and Mitigation Strategies for Large Language Models",
    author = "Chen, Jie  and
      Zhang, Yupeng  and
      Wang, Bingning  and
      Zhao, Wayne Xin  and
      Wen, Ji-Rong  and
      Chen, Weipeng",
    editor = "Al-Onaizan, Yaser  and
      Bansal, Mohit  and
      Chen, Yun-Nung",
    booktitle = "Findings of the Association for Computational Linguistics: EMNLP 2024",
    month = nov,
    year = "2024",
    address = "Miami, Florida, USA",
    publisher = "Association for Computational Linguistics",
    url = "https://aclanthology.org/2024.findings-emnlp.873/",
    doi = "10.18653/v1/2024.findings-emnlp.873",
    pages = "14855--14865"
    
}

@inproceedings{
liu2024best,
title={Best Practices and Lessons Learned on Synthetic Data},
author={Ruibo Liu and Jerry Wei and Fangyu Liu and Chenglei Si and Yanzhe Zhang and Jinmeng Rao and Steven Zheng and Daiyi Peng and Diyi Yang and Denny Zhou and Andrew M. Dai},
booktitle={First Conference on Language Modeling},
year={2024},
url={https://openreview.net/forum?id=OJaWBhh61C}
}

@inproceedings{xie2024doremi,
 author = {Xie, Sang Michael and Pham, Hieu and Dong, Xuanyi and Du, Nan and Liu, Hanxiao and Lu, Yifeng and Liang, Percy S and Le, Quoc V and Ma, Tengyu and Yu, Adams Wei},
 booktitle = {Advances in Neural Information Processing Systems},
 doi = {10.52202/075280-3059},
 editor = {A. Oh and T. Naumann and A. Globerson and K. Saenko and M. Hardt and S. Levine},
 pages = {69798--69818},
 publisher = {Curran Associates, Inc.},
 title = {DoReMi: Optimizing Data Mixtures Speeds Up Language Model Pretraining},
 url = {https://proceedings.neurips.cc/paper_files/paper/2023/file/dcba6be91359358c2355cd920da3fcbd-Paper-Conference.pdf},
 volume = {36},
 year = {2023}
}

\end{document}